\documentclass[11pt,letterpaper]{article}
\input{style/arxiv_preamble}
\title{Stein Guidance without Gradients:\\
Discretization Uniform Diffusion Inference Time Scaling beyond Particle SMC}
\author{
  \textbf{Youheng Zhu}$^1$, 
  \textbf{Chenyang Wang}$^1$, 
  \textbf{Minshuo Chen}$^1$, 
  \textbf{Yiping Lu}$^2$\thanks{The work was done when Yiping was at Northwestern IEMS.} \\
  $^1$Department of Industrial Engineering and Management Sciences, Northwestern University \\
  $^2$Beijing International Center for Mathematical Research, Peking University \\
  \texttt{\{youhengzhu, chenyangwang\}@u.northwestern.edu}, 
  \texttt{minshuo.chen@northwestern.edu}, \\\texttt{yipinglu@bicmr.pku.edu.cn}
}
\begin{document}
\maketitle
\begin{abstract}
\color{black}
Reward-guided diffusion requires effective steering and correction of the bias introduced by approximate value functions. A Brownian example shows that SMC with full multinomial resampling at every step can require particle counts to grow linearly with guidance depth to maintain accuracy. We propose Stein Guidance, which combines derivative-free guidance with whole-path independence Metropolis–Hastings correction. Value-weighted selection from centered Gaussian candidates approximates gradient guidance in expectation, using as few as two queries per step. Centering stabilizes the local normalizer estimates, and the same queries yield an explicit path weight for correction. Under uniform value and spatial regularity bounds, we prove superpolynomial decay of finite-grid total variation error in the number of MH updates, with constants that do not grow with guidance depth. The error prefactor tends to zero as the value functions approach exact Doob values and the time grid is refined.

\end{abstract}

\section{Introduction}

{Diffusion models \citep{sohl2015deep, song2021score} learn and implement a progressive denoising process driven by a score function to generate high-fidelity samples. Such a paradigm empowers and transforms broad engineering and scientific domains. However, in these applications, the generated samples need to satisfy complex conditions. For example, in text-to-image generation, the generated images need to comply with the prompt. In robotics and control, the generated action sequence needs to achieve large task reward or high success rates. Inference time scaling---directly applying training-free modifications to the denoising process---is widely adopted to steer the generation toward desired downstream reward \citep{uehara2025reward}, explore more candidates \citep{ma2025inference,skreta2025fk,chen2025afdps}, or correct sampling bias
\citep{singhal2025framework,zhu2026power,he2026rne,wang2026remasking,ren2026driftlite,qu2026tftf}, and meanwhile to avoid computational overhead from fine-tuning.

\begin{figure}[t]
\centering
\begin{tikzpicture}[tq/every picture, x=.82cm,y=.86cm]
\path[tq/panel] (0,0) rectangle (5.25,6.60);
\path[tq/panel] (5.48,0) rectangle (10.62,4.50);
\path[tq/panel] (10.85,0) rectangle (16.00,4.50);
\path[tq/panel,fill=twolightgray] (5.48,4.80) rectangle (16.00,6.60);

\node[tq/title,anchor=west] at (0.92,6.31) {(a) SMC resampling};

\node[tq/boxred,inner sep=3.1pt] at (2.63,5.38) {%
  \begin{minipage}{3.92cm}\centering
  {\sffamily\tiny Brownian proposals; fixed terminal target}\\[1pt]
  {\small $\displaystyle
    \|\nu_{n,N}^{\rm SMC}-\eta\|_{\TV}\asymp\frac nN$}\\[1pt]
  {\sffamily\tiny $2\le n\le N$; local Bellman error $O(1/n)$}
  \end{minipage}%
};

\node[tq/small,font=\sffamily\fontsize{5.6}{6.7}\selectfont]
  at (2.63,4.46) {shared ancestry $\Longrightarrow$ shared noise};

\input{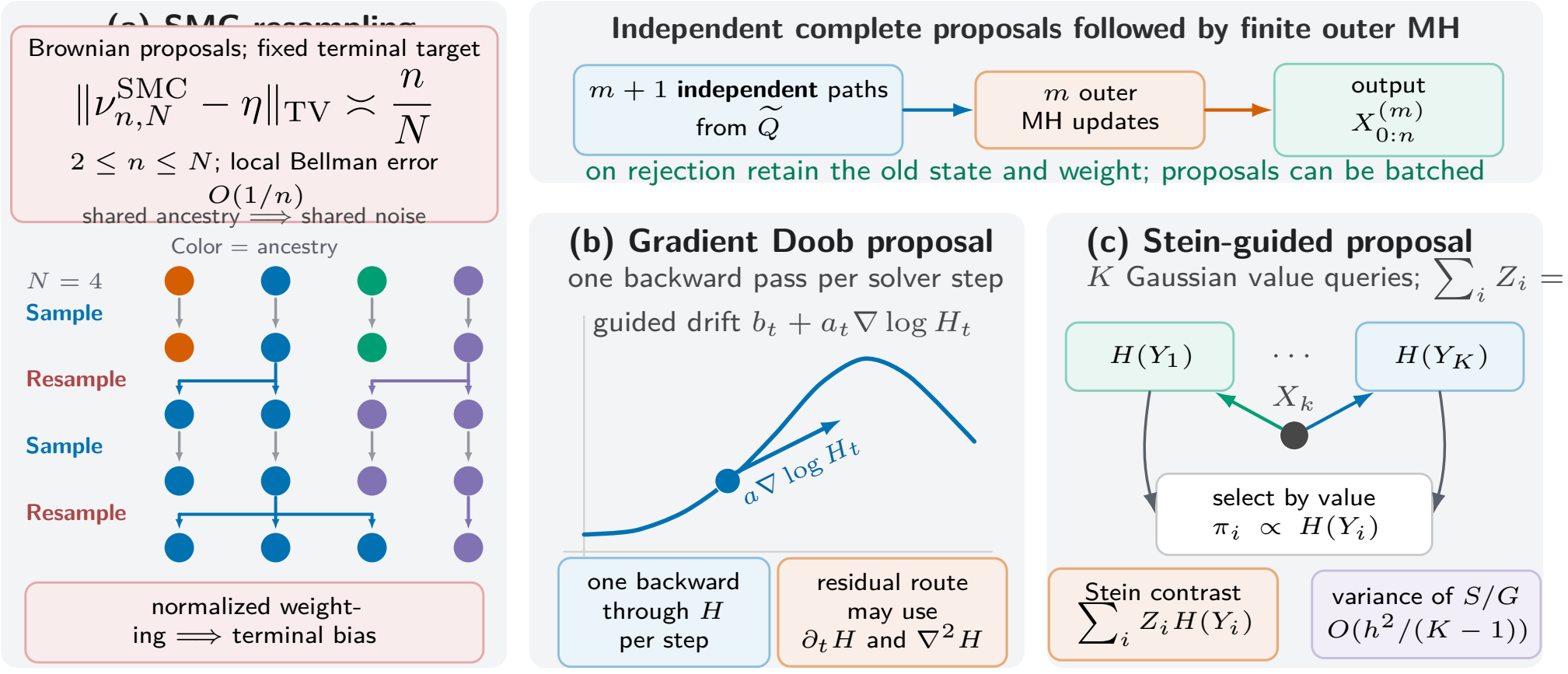}
\node[tq/boxred,text width=3.65cm,minimum height=.45cm,
      font=\sffamily\tiny] at (2.63,0.45)
{normalized weighting $\Longrightarrow$ terminal bias};

\node[tq/subtitle] at (10.74,6.31)
{Independent complete proposals followed by finite outer MH};

\node[tq/boxblue,text width=2.55cm,minimum height=.62cm,
      font=\sffamily\tiny] (onepath) at (7.65,5.52)
{$m+1$ \textbf{independent} paths\\from $\widetilde Q$};
\node[tq/boxorange,text width=1.70cm,minimum height=.62cm,
      font=\sffamily\tiny] (correct) at (11.30,5.52)
{$m$ outer\\MH updates};
\node[tq/boxteal,text width=1.70cm,minimum height=.62cm,
      font=\sffamily\tiny] (accepted) at (14.40,5.52)
{output\\$X_{0:n}^{(m)}$};
\draw[tq/flowblue] (onepath)--(correct);
\draw[tq/floworange] (correct)--(accepted);
\node[tq/small,text=twoteal!78!black] at (10.74,4.90)
{on rejection retain the old state and weight; proposals can be batched};

\node[tq/title,anchor=west] at (5.72,4.20) {(b) Gradient Doob proposal};
\node[tq/small,anchor=west] at (5.72,3.84)
{one backward pass per solver step};

\draw[black!18,line width=.42pt] (5.84,1.15)--(10.28,1.15);
\draw[black!18,line width=.42pt] (6.05,1.00)--(6.05,3.48);
\draw[twoblue,line width=1.05pt,smooth]
  plot coordinates {(6.05,1.32) (6.58,1.36) (7.08,1.54) (7.56,1.86)
                    (8.03,2.31) (8.50,2.81) (8.96,3.06) (9.43,2.88) (10.10,2.24)};
\node[tq/node,minimum size=6.2pt,fill=twoblue] (gx) at (7.54,1.85) {};
\draw[tq/flowblue,line width=1.0pt] (7.62,1.93)--(8.73,2.46)
  node[midway,below,sloped,font=\sffamily\tiny,text=twoblue!88!black]
  {$a\nabla\log H_t$};
\node[tq/small] at (8.12,3.40) {guided drift $b_t+a_t\nabla\log H_t$};

\node[tq/boxblue,text width=1.55cm,minimum height=.70cm,
      font=\sffamily\tiny] at (6.88,0.55)
{one backward\\through $H$ per step};
\node[tq/boxorange,text width=1.70cm,minimum height=.70cm,
      font=\sffamily\tiny] at (9.25,0.55)
{residual route may use\\$\partial_tH$ and $\nabla^2H$};

\node[tq/title,anchor=west] at (11.09,4.20) {(c) Stein-guided proposal};
\node[tq/small,anchor=west] at (11.09,3.84)
{$K$ Gaussian value queries; $\sum_i Z_i=0$};
\node[tq/node,minimum size=7pt,fill=black!72] (x) at (13.42,2.30) {};
\node[tq/small,anchor=south] at (13.42,2.39) {$X_k$};
\node[tq/boxteal,text width=1.18cm,minimum height=.58cm,
      font=\sffamily\tiny] (hm) at (11.92,3.08) {$H(Y_1)$};
\node[tq/boxblue,text width=1.18cm,minimum height=.58cm,
      font=\sffamily\tiny] (hp) at (14.93,3.08) {$H(Y_K)$};
\node[tq/small] at (13.42,3.08) {$\cdots$};
\draw[tq/flowteal] (x)--(hm);
\draw[tq/flowblue] (x)--(hp);
\node[tq/boxgray,text width=2.10cm,minimum height=.60cm,
      font=\sffamily\tiny] (sel) at (13.42,1.52)
{select by value\\$\pi_i\propto H(Y_i)$};
\draw[tq/flow] (hm.south) to[bend right=12] (sel.west);
\draw[tq/flow] (hp.south) to[bend left=12] (sel.east);
\node[tq/boxorange,text width=1.70cm,minimum height=.70cm,
      font=\sffamily\tiny] at (12.06,0.53)
{Stein contrast\\$\sum_iZ_iH(Y_i)$};
\node[tq/boxpurple,text width=1.70cm,minimum height=.70cm,
      font=\sffamily\tiny] at (14.79,0.53)
{variance of $S/G$\\$O(h^2/(K-1))$};
\path (0,-0.08) (16.00,6.67);
\end{tikzpicture}
\caption{\textbf{Guidance depth and efficient correction.}
(a) Repeated resampling accumulates population fluctuations; normalized weighting converts them into terminal bias.
(b) Direct Doob guidance requires value gradients.
(c) Centered Stein pools provide derivative-free guidance and low-variance estimates of local normalizers.
The same queries give path weights for outer {\color{black}Metropolis-Hastings} (MH) correction with independent complete proposals (top).}
\label{fig:overview-tikz}
\end{figure}

A growing body of inference time methods revolves around adding a {\it guidance} term to the original score function. The well-known classifier guidance computes an external gradient of a classifier logit to satisfy categorical conditions. For broader complex conditions, we assess the quality of generated samples by a {\color{black} non-negative} reward function $\phi$. Thus, steering can be formulated as sampling from a reward tilted law
\begin{align}\label{eq:1}
\Pi(\dd X_{0:T}) =Z^{-1}\phi(X_T)P(\dd X_{0:T}),\qquad Z=\E_P[\phi(X_T)],
\end{align}
where $P$ is the path measure of the original diffusion denoising process $X_{0:T}$. In this case, stochastic optimal control provides principled derivations for guidance. One notable example is the Doob's $h$-transform, which defines the guidance term as $\nabla \log h_t(x)$ with $h_t(x) = \E_P[\phi(X_T)\mid X_t=x]$ \citep{tang2025stochastic,nguyen2025h,denker2024deft,zhu2026doit,chen2026guidedbridge,li2026bridgetwist}. Beyond direct gradient, implicit guidance includes other approaches like {\it importance resampling} and {\it proposal and rejection}.

Despite the mathematical cleanness, efficiently and accurately computing the guidance term is challenging. The performance of classifier guidance heavily relies on the accuracy of the external classifier. Doob's $h$-function involves the intractable expectation over the path measure $P$. Therefore, practical implementations of guidance rely on stochastic approximation, and inevitably lead to sampling biases. These biases distort the generated distribution and are empirically tied to the well-recognized failure mode---reward hacking \citep{dandapanthula2026we}.

In this regard, {\it correction} is an essential alternative to stay faithful to the reward tilted law. Correction needs extra computation through maintaining a pool of particles or fresh rollouts, of which the size we term as the {\it statistical width}. Similarly, we term the number of guidance steps we erect on a base diffusion denoising process as {\it guidance depth}. Roughly speaking, the depth distributes the guidance effort, while the width controls correction accuracy. A popular mechanism for guidance and correction is Sequential Monte Carlo (SMC) \citep{wunips2023,cardoso2024montecarlo,skreta2025fk,wang2026urge}, which simulates a batch of particles, reweights and resamples them by the reward during evolution. There are two critical design choices: 1) the correction batch size $N$ (i.e. statistical width), and 2) the number of reweighting steps $n$ (i.e. guidance depth). A large depth makes guidance easier per step, yet repeated resampling creates shared ancestry and accumulates population fluctuations. Normalized weighting can turn these fluctuations into terminal bias. We exhibit a Brownian example with single-output TV error $\Theta(n/N)$ for $2\le n\le N$ (Theorem~\ref{thm:smc-marginal}). Thus increasing depth can require greater width to maintain accuracy. This motivates our central question:
\begin{center}
\emph{Can guidance depth be increased without forcing statistical width to grow?}
\end{center}

To achieve this goal, increasing the number of guidance steps should not lead to stronger correlations between trajectories. This motivates rejection based sampling: generate
proposal paths independently and correct them by acceptance or rejection. 

The challenge remains is to retain the benefits of guidance while correcting the bias introduced by an imperfect value function at low computational cost. Explicit guidance requires value gradients, and its correction either need gradients or time derivatives and Hessians.
{\color{black} Therefore, our main contribution is a derivative-free guidance algorithm \method. Its key idea is simple: \textbf{categorically sampling from centered candidates by their values recovers the first order behavior of a gradient guidance, requiring zero evaluation of derivatives, and as low as two candidate queries to the value stepwise.} Larger pool size of the parallelizable candidates or smaller step size improves the approximation accuracy to the gradient guidance, allowing scaling the compute for better performance. The same queries also allow an explicit path weight that can be used to correct any biases. Specifically, we run $m$ outer Metropolis-Hastings (MH) to accept or reject the proposed path as the correction method. Our main theorem then show that}
{under uniform value and regularity bounds, the finite-grid TV error decays superpolynomially in $m$, with constants independent of guidance depth (Theorem~\ref{omh:thm:main-MH}). The bound separates accumulated Bellman error from finite-pool variance. This is consistent with the intuition behind the Doob $h$-transform: exact Doob transitions need no correction. In our finite-pool implementation, exact Doob values eliminate the Bellman error term, while finer grids and larger pools reduce the remaining pool variance.}
}

{\color{black}
Related approaches to reward-guided diffusion include value-based candidate
selection in SVDD~\citep{li2024svdd}, Monte Carlo estimation of Doob guidance
in DOIT~\citep{zhu2026doit}, and importance weighting with particle resampling
in SMC~\citep{skreta2025fk}.
Table~\ref{tab:guidance-comparison} compares whether these algorithms include
correction, how their output-TV guarantees depend on guidance depth at fixed
statistical width, and whether they support parallel computation.
Our method combines centered value-based selection with whole-path MH
correction, using the same queries for guidance and correction weights.
Section~\ref{app:guidance_comparison} details the assumptions and
implementation choices underlying the comparison.
}

We also demonstrate our theory through experiments. The first synthetic experiment shows that centered pools with just two queries per step match the sampling efficiency of much larger independent pools.
In offline RL experiment, our algorithm reduces critic-$Q$ distribution error most effectively under the same statistical width.
Further text-to-image examples illustrate improved alignment with multi-object prompts.

\providecommand{\yesmark}{\textcolor{teal}{\ding{51}}}
\providecommand{\nomark}{\textcolor{orange}{\ding{55}}}
\begin{table}[t]
\centering
\caption{Guidance and correction methods (conditions in Section~\ref{app:guidance_comparison}).}
\label{tab:guidance-comparison}
\small
\setlength{\tabcolsep}{3pt}
\renewcommand{\arraystretch}{1.08}
\begin{tabularx}{\linewidth}{@{}>{\raggedright\arraybackslash}p{0.35\linewidth}*{3}{>{\centering\arraybackslash}X}@{}}
\toprule
Method & Correction with approximate values & Stable with guidance depth? & Parallelizable?\\
\midrule
SVDD \citep{li2024svdd} & \nomark & \yesmark & \yesmark\\
SMC & \yesmark & \nomark & \yesmark\\
DOIT, MC \citep{zhu2026doit} & \nomark & \nomark\textsuperscript{\textcolor{purple}{$\dagger$}} & \yesmark\ ($M$)\\
Stein ($K=2$) with outer MH & \yesmark & \yesmark & fixed at 2\\
\textbf{Stein with outer MH} & \yesmark & \yesmark & \yesmark\ ($K$)\\
\bottomrule
\end{tabularx}
\par\vspace{2pt}
\begin{minipage}{\linewidth}\footnotesize
At fixed statistical width, we compare how output-TV bounds depend on guidance depth.
For Stein, see Theorem~\ref{omh:thm:main-MH}; for the SMC counterexample, see Theorem~\ref{thm:smc-marginal}.
\textsuperscript{\color{purple}$\dagger$}At fixed $M$, DOIT's TV bound can worsen as smaller noise scales amplify the rollout error bound \citep[Lemma~5.2 and Theorem~5.4]{zhu2026doit}.
\end{minipage}
\end{table}

\FloatBarrier

\section{The Obstruction: Resampling Bias Grows with Depth}
\label{sec:setup-smc}

{\color{black} For a fixed sampling error tolerance, a scalable inference-time sampler should allow the guidance depth $n$ to increase without forcing the statistical width $N$ to grow. In this section, we exhibit a fixed Brownian example in which standard SMC with full multinomial resampling has terminal marginal TV error $\Theta(n/N)$ for $2\le n\le N$.}

{\color{black} To illustrate this, we consider an SMC sampler with statistical width (number of particles) $N$ and guidance depth $n=1/h$ in Theorem~\ref{thm:smc-marginal}. In the theory of SMC~\citep{zhao2024probabilistic, zhu2026power}, the exact intermediate value function satisfies the Bellman equation $H_t^\star(x_t)=\int P_h(d x_{t+h}|x_t)H_{t+h}^\star(x_{t+h})$; the Bellman error measures the deviation of an approximate value function from this identity. %
We work in the $O(1/n)$ local Bellman error regime of~\citet{zhu2026power}, which removes the exponential-in-$n$ factor from the SMC complexity bounds in that analysis.
Nevertheless, Theorem~\ref{thm:smc-marginal} shows that a linear-in-$n$ dependence can remain: in our example, the terminal marginal TV error is $\Theta(n/N)$ for $2\le n\le N$.
Repeated resampling propagates ancestral fluctuations and adds fresh sampling noise, allowing the variance of particle averages to accumulate with depth.
Subsequent normalized weighting converts these fluctuations into terminal marginal bias; in our construction, the resulting Jensen gaps accumulate to order $n/N$.}

\begin{theorem}[Brownian SMC marginal bias]\label{thm:smc-marginal}
Let $P_h(x,\cdot)=\mathcal N(x,h)$, where $h=1/n$.
There exist a smooth reward $\phi>0$, a value function $H:[0,1]\times\R\to(0,\infty)$ bounded above and away from zero, and constants $C,c_-,c_+>0$, all independent of $n,N$, such that for every $n\ge2$, $H_k=H(k/n,\cdot)$ satisfies $H_0\equiv1$, $H_n=\phi$, and
\begin{equation}\label{eq:smc-bellman-bound}
\max_{0\le k<n}\sup_x
\max\left\{\frac{P_hH_{k+1}(x)}{H_k(x)},
           \frac{H_k(x)}{P_hH_{k+1}(x)}\right\}
\le 1+\frac Cn.
\end{equation}
Starting at zero, run $N$-particle SMC with proposal $P_h(x,\dd y)$, incremental weight $H_k(y)/H_{k-1}(x)$, and multinomial resampling after every step.
The law $\nu_{n,N}^{\rm SMC}$ of one terminal particle satisfies
\begin{equation}\label{eq:smc-marginal-bounds}
c_-\frac nN
\le \bigl\|\nu_{n,N}^{\rm SMC}-\eta\bigr\|_{\TV}
\le c_+\frac nN,
\qquad 2\le n\le N,
\end{equation}
where $\eta(\dd x)\propto\phi(x)\mathcal N(0,1)(\dd x)$ is fixed independently of $n,N$.
\end{theorem}
Appendix~\ref{app:proofs} gives the construction and proof, relating the accumulated fluctuations to a sum of Jensen gaps. {\color{black} Consequently, in this example the terminal marginal TV error is $\Theta(n/N)$ throughout $2\le n\le N$, so maintaining a fixed accuracy as depth increases can require particle count to grow proportionally to depth.
}

\section{\color{black}Stein Guidance with Terminal Correction}
\label{m:method}

{\color{black} In this section, we construct an derivative-free algorithm whose guidance depth can be increased without forcing statistical width to grow, resolving the main issue of SMC. This is achieved by independently proposing complete guided trajectories, followed by acceptance and rejection at the end, thereby eliminating the dependency induced by shared ancestry and its contribution to the sampling error.}

{\color{black} Unlike SMC, independent trajectories are not interactive and cannot redistribute weight among ancestral paths, therefore correction is done at the end of each trajectory, specifically by a Metropolis-Hastings (MH) step. MH uses the importance weight between the proposal and the target, and calculates the acceptance rate to determine whether or not to update the current accepted trajectory. The resulting Markov chain then mixes towards the target law, which is the invariant law of the Markov chain, correcting any biases of the guided proposal. These biases are non-negligible because the effective guidance is not exactly the Doob's $h$-transform. Overall, the complete algorithm has two parts: \emph{proposal generation} (Algorithm~\ref{alg:stein}) and \emph{correction} (Algorithm~\ref{alg:outer-mh}).}

{\color{black} Consider a discretized diffusion backward process with uniform $n$ time step $h$, the totally time $T=nh$ and we write $t_k=kh$ to indicate the $k$th intermediate time. Corresponding to the continuous path law $P, \Pi$ in \eqref{eq:1}, we write the discrete path law $P_n=\delta_{x_\star}\prod_{k=0}^{n-1}P_k$, and the tilted path law $\Pi_n=Z_n^{-1}\phi(X_n)P_n$ we aim to sample from. Here $P_k(x,\cdot)$ is the Gaussian Euler discretized transition kernel, which is the law of $x+hb_k(x)+\sqrt h\,\sigma_k(x)Z$, with $Z\sim N(0,I_d)$. The values $H_0,\ldots,H_n$ are finite, strictly positive, and measurable, with $H_n=\phi$. We also assume the terminal reward $\phi$ to be strictly positive. We assume integrability so that $Z_n=\mathbb E_{P_n}\phi(X_n)$ is the finite normalizer, and each $G_k=P_kH_{k+1}$ is finite on reachable states. During a denoising diffusion process, one starts from $x_\star$, generate $X_1\sim P_0(x_\star,\cdot)$, and sequentially $X_k\sim P_{k-1}(X_{k-1},\cdot)$.
}

{\color{black} We define the local tilt as} $K_k^H(x,\dd y)=P_k(x,\dd y)H_{k+1}(y)/G_k(x)$, and $\Pi_n^H=\delta_{x_\star}\prod_kK_k^H$. It {\color{black}is a tilted kernel of the base kernel $P_k$ at each discretization step, and} exactly equals the target $\Pi_n$ when $H_k=G_k$ {\color{black}for every $k$, i.e. in coincidence with the Doob's $h$-transform.} The value functions $H_0,\ldots,H_n$ stay fixed throughout the execution of the algorithm. We reserve $m$ for the number of outer MH updates. Random initialization requires the additional factor in Proposition~\ref{omh:prop:initial}.

\paragraph{Derivative-free guidance for efficient correction.}
{\color{black} The derivatives of the value functions may be unavailable or computationally expensive to evaluate.
We therefore seek guidance and correction using only value function queries.
Existing derivative-free approaches target the local tilt $K_k^H$ through rejection sampling~\citep{golowich2026reject} or local MH, or select a candidate from a finite pool with probabilities proportional to its value~\citep{li2024svdd,zhu2026power,golowich2026reject,jiang2026smcevolve}.
With approximate value functions, however, even exact sampling from $K_k^H$ generally leaves path-level bias.
Direct evaluation of the corresponding path importance weights involves the local normalizers $G_k=P_kH_{k+1}$, which are generally intractable. Alternative approaches would require access to time derivatives and Laplacians, or a cumulative sum of the gradient, which are not practical in high-dimensional generation, and violates our derivative-free requirement.}

\textbf{Our design principle is to improve the guided transition and its correction weight together.}
{\color{black} At each step, we evaluate $H_{k+1}$ on a pool of candidate next states.
These values serve two purposes: they determine which candidate is selected, and their average estimates the local normalizer $G_k(x)$.
Our construction also uses these pool averages in a computable correction weight for the complete trajectory~\citep{zhu2026power}.
Since the outer MH acceptance probability depends on this weight, controlling fluctuations in the pool averages is important for efficient correction.}

This motivates our Stein guidance.
Stein's identity gives a derivative-free construction based on $K$ parallel value queries (Section~\ref{m:stein_guidance}).
The centered Gaussian noises sum to zero, so the candidates' linear value changes cancel in the pool average.
The remaining fluctuation is governed by curvature.
Under Assumption~\ref{ass:theory-regularity}, its normalized variance is $O(h^2/(K-1))$, compared with $O(h/K)$ for independent candidates.
Section~\ref{m:theory} shows how this reduces the finite-pool term in the outer MH output-TV bound.
For $K=2$, the centered pool contains two opposite Gaussian perturbations.
Their value difference gives a central finite difference along the sampled direction.
\textbf{With centering, two value queries already recover the leading guidance drift in expectation} (Section~\ref{m:two_value_guidance}).

\subsection{Derivative-Free Guidance via Stein's Identity}
{\color{black} For the following one-step calculations, fix $k$ and $X_k=x$, and write $\mu=x+hb_k(x)$, $\sigma=\sigma_k(x)\in\R^{d_x\times d}$, $H=H_{k+1}$, and $G=G_k(x)$.}

\subsubsection{Warm up: Derivative-Free Guidance via Two Query Finite Difference}
\label{m:two_value_guidance}
To avoid derivative evaluations while still steering the dynamics toward regions of higher reward, a natural approach is to estimate the guidance using finite differences. We show that this guidance can be implemented by selecting between two symmetric queries based on their value function evaluations.

Instead of using a single diffusion proposal $Y=x+hb(x)+\sqrt h\,\sigma Z$ with $Z\sim N(0,I_d)$, we construct two symmetric queries $Y^\pm=x+hb(x)\pm\sqrt h\,\sigma Z$. We select between them with probabilities $\Pr(J=\pm1\mid Z)=H^\pm/(H^++H^-)$, where $H^\pm=H(Y^\pm)$. For positive $C^2$ value functions, Taylor expansion and averaging over $Z$ give, as $h\to0$,
\begin{equation}\label{m:twoq_difference}
\E[Y^J]
=x+hb(x)+\sqrt h\,\sigma\E\!\left[Z\frac{H^+-H^-}{H^++H^-}\right]
=x+h\bigl[b(x)+\sigma\sigma^\top\nabla\log H(x)\bigr]+o(h).
\end{equation}
Thus value-weighted selection recovers the Doob guidance drift, {\color{black}i.e. $\sigma\sigma^\top\nabla\log H(x)$ up to leading factor} in expectation. Selecting between two symmetric queries implements derivative-free guidance based on finite differences.

\subsubsection{Stein Guidance}
\label{m:stein_guidance}
{\color{black}
We extend the two-query rule to $K\ge2$ candidates.
Draw $\xi_i\stackrel{\rm iid}{\sim}N(0,I_d)$ and set
\begin{equation}\label{m:pool}\textstyle
Z_i=\sqrt{\frac K{K-1}}\big(\xi_i-\frac1K\sum_{j=1}^K\xi_j\big),
\qquad Y_i=\mu+\sqrt h\,\sigma Z_i.
\end{equation}
Each $Z_i$ remains standard Gaussian, and $\sum_iZ_i=0$.
Query all $H(Y_i)$ in parallel, then set $X_{k+1}=Y_J$ with $\Pr(J=i\mid Y_1,\ldots,Y_K)=H(Y_i)/\sum_jH(Y_j)$.
For $K=2$, this recovers the symmetric rule above.

The selected state's conditional mean is
\begin{equation}\label{m:stein}\textstyle
\E[Y_J\mid Z_1,\ldots,Z_K]
=\mu+\sqrt h\,\sigma\frac{\sum_iZ_iH(Y_i)}{\sum_iH(Y_i)}.
\end{equation}
Stein's identity relates the value-weighted noises to gradients: for $Z\sim N(0,I_d)$, whenever Gaussian integration by parts is valid,
\begin{equation}\label{m:stein_derivation}\textstyle
\frac1{\sqrt h}\E\big[ZH(\mu+\sqrt h\,\sigma Z)\big]
=\sigma^\top\E\big[\nabla H(\mu+\sqrt h\,\sigma Z)\big].
\end{equation}
This applies to each candidate because its Gaussian marginal is preserved.
Since $\sum_iZ_i=0$, for each fixed $K$ the denominator in~\eqref{m:stein} is $KH(\mu)+O_{L^2}(h)$.
Under Assumption~\ref{ass:theory-regularity}, $H$ is bounded away from zero, so replacing the denominator by its leading term and applying~\eqref{m:stein_derivation} termwise gives $\E\big[\frac{\sum_iZ_iH(Y_i)}{\sum_iH(Y_i)}\big]=\sqrt h\,\sigma^\top\nabla\log H(\mu)+o(\sqrt h)$.
Substituting into~\eqref{m:stein}, with $\mu=x+hb_k(x)$, recovers~\eqref{m:twoq_difference}.
Finite-pool selection is generally biased relative to $K_k^H$; Appendix~\ref{sec:score} quantifies the mean error.

The algorithm also records $S=K^{-1}\sum_iH(Y_i)$, an unbiased estimate of $G$ used in the whole-path correction weight.
Centering cancels the $O(\sqrt h)$ linear term in its Taylor expansion, leaving fluctuations of order $h$.
Under the same assumptions, $\Var(S/G)=O(h^2/(K-1))$, compared with $O(h/K)$ for independent candidates.
Algorithm~\ref{alg:stein} repeats the update and accumulates the correction weight.
}

\begin{sqalgorithm}{One complete \method proposal}\label{alg:stein}
\setlength{\topsep}{0pt}\setlength{\partopsep}{0pt}
\begin{tabbing}
\hspace{1.2em}\=\hspace{1.2em}\=\kill
\textbf{Input:} $x_\star,h,n$, $K\ge2$, $\{b_k,\sigma_k,H_{k+1}\}_{k=0}^{n-1}$.\quad\textbf{Init:} $X_0\gets x_\star$, $\ell\gets0$.\\
\textbf{for} $k=0,\ldots,n-1$ \textbf{do}\\
\>Draw a fresh $U_k=(Y_1,\ldots,Y_K)$ at $X_k$ using~\eqref{m:pool}.\\
\>$v_i\gets H_{k+1}(Y_i)$ ($i=1,\ldots,K$, in parallel); $S_k\gets K^{-1}\sum_{i=1}^K v_i$.\\
\>$\ell\gets\ell+\log S_k$; \textbf{if} $k\ge1$ \textbf{then} $\ell\gets\ell-\log c$.\\
\>Draw $J\sim\mathrm{Categorical}\bigl((v_i/(KS_k))_{i=1}^K\bigr)$; $X_{k+1}\gets Y_J$; $c\gets v_J$.\\
\textbf{end for}\\
\textbf{return} $(X_{0:n},R=e^\ell)$.
\end{tabbing}
\end{sqalgorithm}

\subsection{Terminal Metropolis--Hastings Correction}

{\color{black}
Algorithm~\ref{alg:stein} returns a path together with the weight $R=H_0(x_\star)\prod_{k=0}^{n-1}\frac{S_k}{H_k(X_k)}$, computed from its value queries.
Let $\cQ$ denote the proposal law on the augmented space of candidate pools, selected indices, and paths.
Lemma~\ref{lem:pool} in Appendix~\ref{sec:setup} gives a one-step weighting identity.
Iterating it cancels the intermediate value factors; using $H_n=\phi$ yields $\E_{\cQ}[R F(X_{0:n})]=Z_n\E_{\Pi_n}[F]$ for every bounded measurable path function $F$ (Theorem~\ref{omh:thm:proper-global}, Appendix~\ref{sec:global}).
For each measurable event $A$ in the augmented space, define $\cP(A):=Z_n^{-1}\E_{\cQ}[R\,\mathbf1_A]$; equivalently, $\cP=(R/Z_n)\cQ$.
The weighting identity ensures that $\cP$ is a probability measure with path marginal $\Pi_n$.

\paragraph{Whole-path Metropolis correction.}
Independence MH with proposal $\cQ$ and target $\cP$ accepts a fresh augmented proposal with probability $1\wedge R'/R$; the unknown normalizer $Z_n$ cancels.
}

\begin{sqalgorithm}{Outer independence MH with $m$ updates}\label{alg:outer-mh}
\setlength{\topsep}{0pt}\setlength{\partopsep}{0pt}
\begin{tabbing}
\hspace{1.2em}\=\hspace{1.2em}\=\kill
\textbf{Input:} $m\ge0$; fixed inputs to Algorithm~\ref{alg:stein}.\quad\textbf{Init:} $(X_{0:n},R)\gets\text{Algorithm~\ref{alg:stein}}$.\\
\textbf{for} $j=1,\ldots,m$ \textbf{do}\\
\>$(X'_{0:n},R')\gets\text{Algorithm~\ref{alg:stein}}$, $u\sim\mathrm{Uniform}(0,1)$, independently.\\
\>\textbf{if} $u\le\min\{1,R'/R\}$ \textbf{then} $(X_{0:n},R)\gets(X'_{0:n},R')$.\\
\textbf{end for}\\
\textbf{return} $X_{0:n}$ (or $X_n$).
\end{tabbing}
\end{sqalgorithm}
{\color{black}
Algorithm~\ref{alg:outer-mh} implements this MH update using the returned pair $(X_{0:n},R)$, retaining both on rejection.
The augmented target $\cP$ is invariant, while Section~\ref{m:theory} quantifies the finite-run initialization error after $m$ updates.
Only the retained path or endpoint and its weight need be stored.
All $m+1$ proposals can be generated in parallel before the ordered accept/retain decisions.
Independent outputs require independent repetitions of the entire procedure.
}

Appendix~\ref{app:local_refinement} extends the proposal with $r$ local pool MH updates and proves its proper weighting identity.

\section{Sampling Error and Guidance Depth}
\label{m:theory}

We bound the sampling error after $m$ outer independence MH updates.
Theorem~\ref{omh:thm:main-MH} controls the total variation distance between the output law $\nu_m$ and the finite-grid target $\Pi_n$ defined in Section~\ref{m:method}.
The bound separates accumulated Bellman error from local Stein variance.

We use the following regularity conditions to bound the sampling error.
\begin{assumption}[Uniform value and spatial bounds]\label{ass:theory-regularity}
Fix the horizon $T$.
There are constants $0<c_0\le C_0<\infty$ and $L_2,\Sigma<\infty$ that do not depend on the grid or pool size.
For every grid with this horizon, every $k=0,\ldots,n-1$, and every $x\in\R^{d_x}$,
\begin{equation}\label{eq:theory-regularity}
\begin{gathered}
H_{k+1}\in C^2(\R^{d_x}),\quad
c_0\le H_{k+1}(x)\le C_0,\quad
\|\nabla^2H_{k+1}(x)\|_{\rm op}\le L_2,\quad
\|\sigma_k(x)\|_{\rm op}\le\Sigma.
\end{gathered}
\end{equation}
\end{assumption}

\paragraph{Bellman error and the Doob transform.}
Ideal intermediate value functions $H_k^*$ satisfy the Bellman equation $H_k^*=P_kH_{k+1}^*$ with $H_n^*=\phi$ \citep{zhao2024probabilistic, zhu2026power}. Their tilted transitions give the discrete Doob transform and sample $\Pi_n$.
For approximate values, the ratio $G_k/H_k$ measures the failure of this equation, as in the Bellman error of~\cite{zhu2026power}.
We use the logarithmic form $\log(P_kH_{k+1})-\log H_k$ as the local Bellman error.
This is also the Bellman residual for stochastic control with terminal reward $\log\phi$ and a KL cost for changing the reference transitions~\citep{todorov2006linearly}.

{\color{black}We measure the accumulated Bellman error under the path law $\Pi_n^H$ defined in Section~\ref{m:method}.}
Let $B$ be the accumulated Bellman error along a path, defined below.
We measure its variation under $\Pi_n^H$ by the centered norm $\mathcal A_\ell$:
\begin{equation}\label{m:action}
 \mathcal A_\ell=\inf_{{\color{black}a}\in\R}\|B-{\color{black}a}\|_{L^\ell(\Pi_n^H)},\qquad
 B(X_{0:n})=\sum_{k<n}\log\frac{G_k(X_k)}{H_k(X_k)},\quad 1\le\ell<\infty.
\end{equation}
For each fixed $\ell$, Assumption~\ref{ass:theory-regularity}
bounds $\mathcal A_\ell$ uniformly in guidance depth
(Lemma~\ref{omh:lem:ideal-density-moments}).
Subtracting a constant removes the part of $B$ that cancels when the path law is normalized.
For a fixed time grid and fixed value functions $H_0,\ldots,H_n$, $\mathcal A_\ell$ is independent of $K$.
It vanishes for exact discrete Doob values, or whenever $B$ is constant.
The following theorem bounds the sampling error.
Its proof appears in Section~\ref{omh:sec:main-proof}.

\begin{theorem}[Sampling error bound with approximate values]\label{omh:thm:main-MH}
Use Algorithm~\ref{alg:stein} with the fixed start, terminal value $H_n=\phi$, and fixed value functions of Section~\ref{m:method}.
Suppose Assumption~\ref{ass:theory-regularity} holds.
For every real $p\ge1$, there is a constant $C_p$ depending only on $p,T,d,c_0,C_0,L_2,\Sigma$.
For all $n\ge1$, $K\ge2$, and integers $m\ge p$, the physical path law $\nu_m$ returned by Algorithm~\ref{alg:outer-mh} satisfies
\begin{equation}\label{omh:eq:main-MH-high}
{\|\nu_m-\Pi_n\|_{\TV}
\le\min\Bigl\{1,\frac{C_p}{m^p}
\min\Bigl\{1,\Bigl(\underbrace{\mathcal A_{2p+2}^{\,2}\vphantom{\frac{Th}{K-1}}}_{\text{\normalfont\scriptsize Bellman error}}+\underbrace{\frac{Th}{K-1}}_{\text{\normalfont\scriptsize local Stein variance}}\Bigr)^{(p+1)/2}\Bigr\}\Bigr\}.}
\end{equation}
The same bound holds for the endpoint marginal law.
One output uses $Kn(m+1)$ candidate values.
\end{theorem}

The theorem indicates a superpolynomial decaying rate for the sampling error with regard to the number of MH correction $m$, this is faster than the $O(1/N)$ rate of SMC based sampling~\citep{zhu2026power}{\color{black}\footnote{\color{black}SMC-based MCMC methods, including iterated conditional SMC and particle Gibbs, can also converge superpolynomially in iteration count under suitable assumptions~\citep{andrieu2018uniform}.}}, and the $O(1/\sqrt M)$ rate of DOIT~\citep{zhu2026doit}, where $N,M$ are the statistical widths for the two types of algorithms respectively.

The coefficient of Theorem~\ref{omh:thm:main-MH} separates the Bellman error term from candidate-pool noise.
The local Stein variance term accumulates the $O(h^2/(K-1))$ variance of each normalized pool average over $n=T/h$ steps.
The gain from $K$ diminishes once the Bellman error term dominates. 
When $B$ is constant, as for exact discrete Doob values, the Bellman error term vanishes. At fixed $m$, finer grids and larger pools reduce the remaining term in the error bound.

\paragraph{Correction cost and guidance depth.}\label{sec:parallel_complexity}
For any fixed TV accuracy, Theorem~\ref{omh:thm:main-MH} bounds the required number of outer MH updates independently of guidance depth.
Therefore, this algorithm provides a positive answer to the question we raise in the beginning.

\section{Experiments: Centering, Pool Size, and Tilted Distributions}
\label{m:experiments}
We first isolate the effect of centering and pool size in a controlled
SIS benchmark, then compare critic-tilted action distributions under a frozen
learned policy. These experiments measure normalizer efficiency and
distribution accuracy, respectively.

\subsection{Centering and pool size}
On a scalar Brownian benchmark with target $\mathcal N(1.5,1)$, we compare
centered and IID pools under exact and perturbed value functions.
Centered $K=2$ uses opposite Gaussian innovations; IID pools use independent ones.
Adjusted NESS measures normalizer-estimation efficiency;
its definition and the protocol appear in Appendix~\ref{app:centering-essential-settings}.

\begin{figure}[t]
\centering\includegraphics[width=0.86\linewidth]{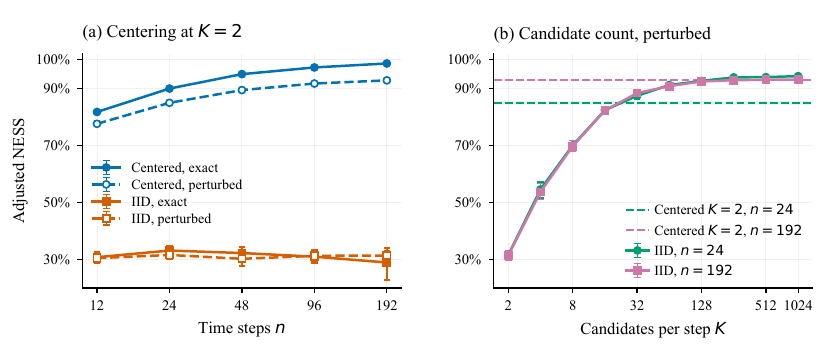}
\caption{Effect of centering and pool size. (a) Mesh refinement with $K=2$ for both centered pools and uncentered IID pools, under exact and perturbed guides. (b) Uncentered IID pools require much larger $K$ to reach the adjusted NESS of centered $K=2$ under the perturbed guide; horizontal bands show the centered references.}
\label{fig:synthetic-centering}
\end{figure}

With $K=2$, finer grids improve centered pools, while IID pools remain near
$30\%$ adjusted NESS under both guides (Figure~\ref{fig:synthetic-centering}(a)).
Under the perturbed guide at $n=192$,
centered $K=2$ reaches $92.76\%$ adjusted NESS, versus $31.40\%$ for IID $K=2$.
IID $K=128$ reaches $92.38\%$ and $K=256$ reaches $92.81\%$, using 64 and 128 times
as many candidate-value evaluations per path as centered $K=2$.
Thus a two-point centered pool matches
the normalizer efficiency of much larger independent pools in this benchmark.

\subsection{Critic-tilted actions in offline RL}
\label{sec:rl-pool}
On Hopper-medium-v2, we compare sampling from
$\pi_{\beta_s}(a|s)\propto p_0(a|s)e^{\beta_sQ(s,a)}$ with a fixed diffusion
policy $p_0$ and critic $Q$.
The three targets assign $30\%$, $60\%$, or $80\%$ of their mass to the
original policy's highest-$Q$ $10\%$ of actions.
We report the Kolmogorov--Smirnov (KS) distance between each method's critic-$Q$
distribution and an independent numerical target reference, averaged over eight
fixed states; lower is better.

Stein uses $K=2,8$ candidates per step and $L=m+1=32$ complete proposals per
output, with Tweedie value estimates and no backpropagation.
We compare against FK~\citep{singhal2025framework}, DAS~\citep{kim2025das},
DOIT~\citep{zhu2026doit} (native and fixed-temperature), uncorrected
SVDD~\citep{li2024svdd}, TFG~\citep{ye2024tfg}, and
TTS-local~\citep{zhang2025search}.
Settings, calibration, and uncertainty estimates appear in
Appendix~\ref{app:rl-experiment-details}.
\begin{table}[t]
\centering\small
\setlength{\tabcolsep}{3pt}
\begin{tabular*}{\linewidth}{@{\extracolsep{\fill}}lrrr@{\hspace{1.5em}}lrrr@{}}
\toprule
Method & $30\%$ & $60\%$ & $80\%$ & Method & $30\%$ & $60\%$ & $80\%$ \\
\midrule
Stein $K2/L32$ & \textbf{0.0306} & \textbf{0.0697} & \textbf{0.1128} & FK $N32$ & 0.0504 & 0.1093 & 0.1816 \\
Stein $K8/L32$ & 0.0426 & 0.0942 & 0.1511 & DAS $N32$ & 0.0307 & 0.0769 & 0.1355 \\
DOIT native & 0.2418 & 0.4778 & 0.6179 & TFG & 0.1788 & 0.2808 & 0.3586 \\
DOIT fixed $\beta$ & 0.2780 & 0.4774 & 0.5903 & TTS-local & 0.1781 & 0.2767 & 0.3752 \\
SVDD $M32$ & 0.1218 & 0.2276 & 0.2979 & & & & \\
\bottomrule
\end{tabular*}
\caption{Mean conditional critic-$Q$ KS over eight fixed Hopper observations (lower is better).}
\label{tab:rl-pool-main}
\end{table}

\begin{figure}[t]
\centering\includegraphics[width=0.86\linewidth]{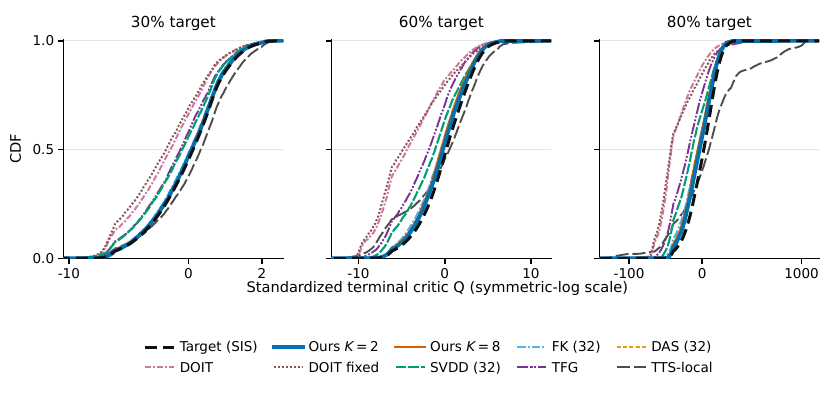}
\caption{Critic-$Q$ distributions at the three target strengths. Curves equally mix the eight states after standardization by the target reference. The horizontal axis is linear near zero and logarithmic in the tails; black is the numerical SIS target reference.}
\label{fig:rl-distribution-curves}
\end{figure}

The $K=2$ sampler has the lowest measured KS at all three targets
(Table~\ref{tab:rl-pool-main}).
It closely matches DAS at $30\%$ and has lower KS at $60\%$ and $80\%$.
Its KS is also $36$--$39\%$ lower than FK's.
Bootstrap intervals that account for multiple comparisons support its gains
over both DOIT variants, SVDD, TFG, and TTS-local at all three targets
(15 comparisons); $K=8$ improves over these baselines in 14 of 15 comparisons.
Figure~\ref{fig:rl-distribution-curves} shows how closely Stein's corrected
$Q$ distributions follow the target reference.

Whole-path MH correction provides a large part of the improvement.
For $K=2$, KS falls from $0.2174$, $0.2804$, and $0.4041$ before correction
to $0.0306$, $0.0697$, and $0.1128$ at $L=32$.
Larger candidate pools do not consistently improve accuracy at this proposal count.
This is consistent with Theorem~\ref{omh:thm:main-MH}: increasing $K$ reduces
pool noise but does not remove the Bellman error of fixed approximate values.

\subsection{Text-to-Image Generation}

We apply our algorithm to text-to-image generation with Stable Diffusion v1.5~\citep{rombach2022high}. The pretrained model provides the conditional image distribution $p_{\mathrm{data}}(x \mid c)$, where \(x\) is the generated image and \(c\) is the text prompt. The pool sizes are \(K=2\) and \(K=4\), with \(L=11\) complete proposals per image: one initial draw followed by 10 whole-path MH updates. FK-Steering~\citep{singhal2025framework}, SVDD~\citep{li2024svdd}, and URGE~\citep{wang2026urge} use \(N=4\) particles on the same backbone. Table~\ref{visualization_test} shows four two-object prompts.

Table~\ref{tab:t2i-main} reports the mean CLIP score~\citep{hessel2021clipscore}, HPSv2~\citep{wu2023hpsv2}, and ImageReward~\citep{xu2023imagereward} over 40 two-object prompts. Stein with \(K=2\) achieves the highest mean on all three metrics.

\begin{table}[t]
    \small
    \centering
    \renewcommand{\arraystretch}{1.15}
    \begin{tabular}{l c c c}
    \toprule
    \textbf{Sampler} & \textbf{CLIP score $\uparrow$} & \textbf{HPSv2 $\uparrow$} & \textbf{ImageReward $\uparrow$} \\
    \midrule
    \texttt{FK-Steering} $(N = 4)$ & 0.2883 & 0.2584 & 0.3489 \\
    \texttt{SVDD} $(K = 4)$        & 0.2761 & 0.2451 & 0.0098 \\
    \texttt{URGE} $(N = 4)$        & 0.2864 & 0.2498 & 0.5723 \\
    \texttt{\bfseries Stein} $(K = 2)$ & \textbf{0.2961} & \textbf{0.2622} & \textbf{0.8290} \\
    \texttt{Stein} $(K = 4)$       & 0.2925 & 0.2599 & 0.3445 \\
    \bottomrule
    \end{tabular}
    \caption{Text-to-image results on 40 prompts from the two-object subset of GenEval~\citep{ghosh2023geneval}, with Stable Diffusion v1.5 and 100 DDIM steps. Each entry is the mean over prompts.}
    \label{tab:t2i-main}
\end{table}

\begingroup
    \small
    \setlength{\tabcolsep}{3pt}
    \ifdefined\promptcol\else\newlength{\promptcol}\fi
    \ifdefined\imgcol\else\newlength{\imgcol}\fi
    \setlength{\promptcol}{0.135\linewidth}
    \setlength{\imgcol}{\dimexpr(0.90\linewidth-\promptcol-8\tabcolsep)/4\relax}
    \setlength{\LTcapwidth}{0.90\linewidth}
    \renewcommand{\arraystretch}{1.1}
    \begin{longtable}{@{}>{\centering\arraybackslash}m{\promptcol}*{4}{>{\centering\arraybackslash}m{\imgcol}}@{}}
    \caption{Comparison of four methods on the same four two-object prompts.}
    \label{visualization_test}\\
    \toprule
    \textbf{Prompt}\par\textit{a photo of ...} &
    \textit{a book and a laptop.} &
    \textit{a knife and a zebra.} &
    \textit{a laptop and a carrot.} &
    \textit{a dining table and a bear.} \\
    \midrule
    \endfirsthead
    \multicolumn{5}{c}{\tablename~\thetable\ (continued)}\\
    \toprule
    \textbf{Prompt}\par\textit{a photo of ...} &
    \textit{a book and a laptop.} &
    \textit{a knife and a zebra.} &
    \textit{a laptop and a carrot.} &
    \textit{a dining table and a bear.} \\
    \midrule
    \endhead
    \midrule
    \multicolumn{5}{r}{\textit{Continued on next page}}\\
    \endfoot
    \bottomrule
    \endlastfoot
    \texttt{FK-\allowbreak{}Steering}\par SDv1.5,\par $N=4$ &
    \includegraphics[width=\linewidth,keepaspectratio]{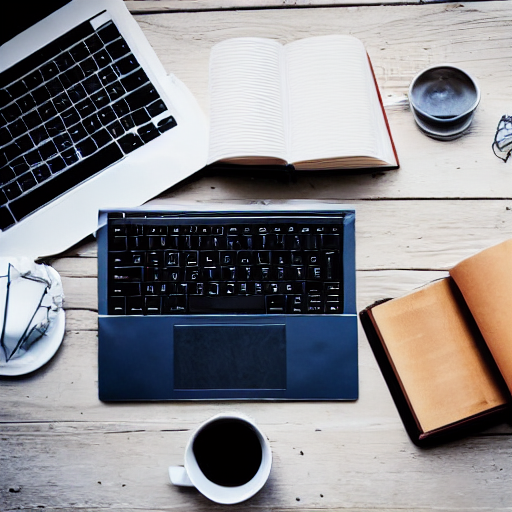} &
    \includegraphics[width=\linewidth,keepaspectratio]{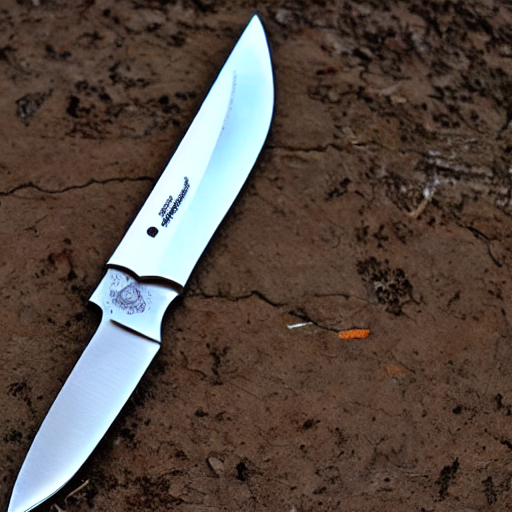} &
    \includegraphics[width=\linewidth,keepaspectratio]{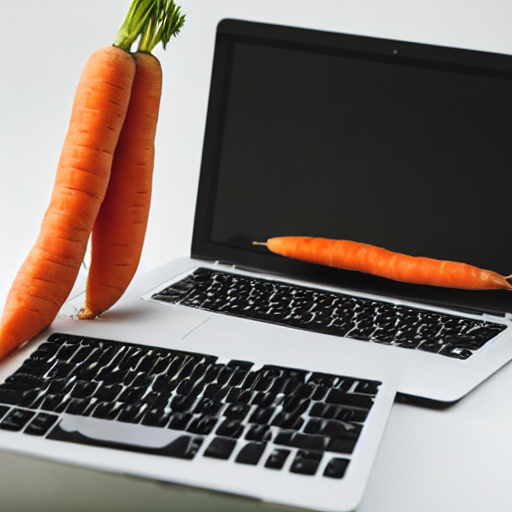} &
    \includegraphics[width=\linewidth,keepaspectratio]{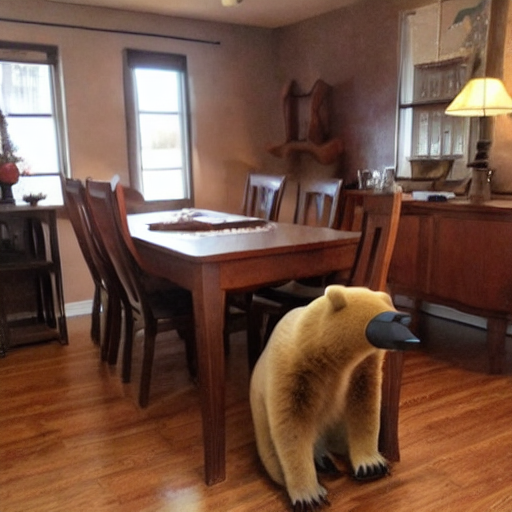} \\
    \texttt{SVDD}\par SDv1.5\par $N=4$ &
    \includegraphics[width=\linewidth,keepaspectratio]{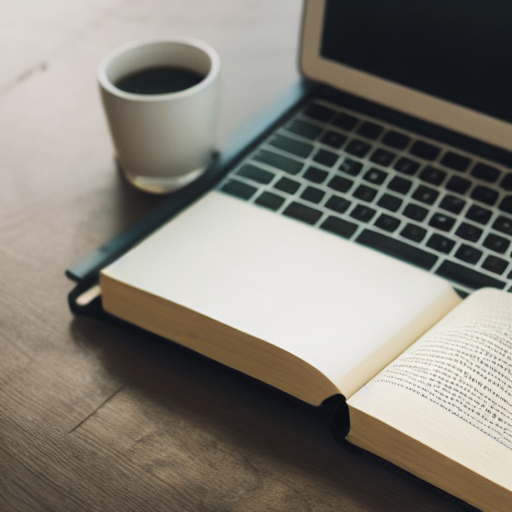} &
    \includegraphics[width=\linewidth,keepaspectratio]{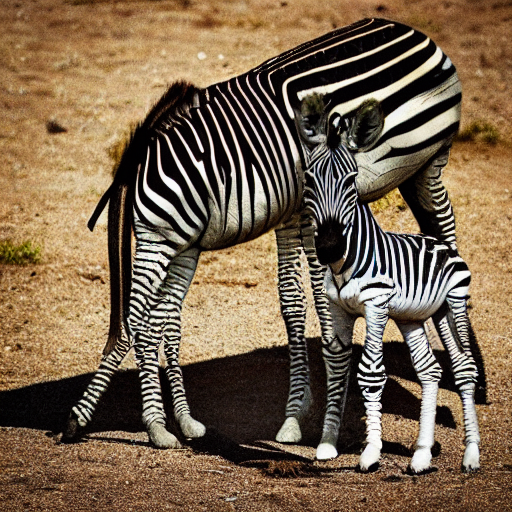} &
    \includegraphics[width=\linewidth,keepaspectratio]{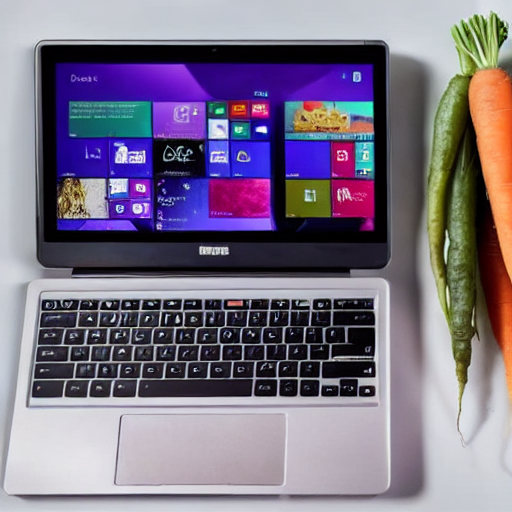} &
    \includegraphics[width=\linewidth,keepaspectratio]{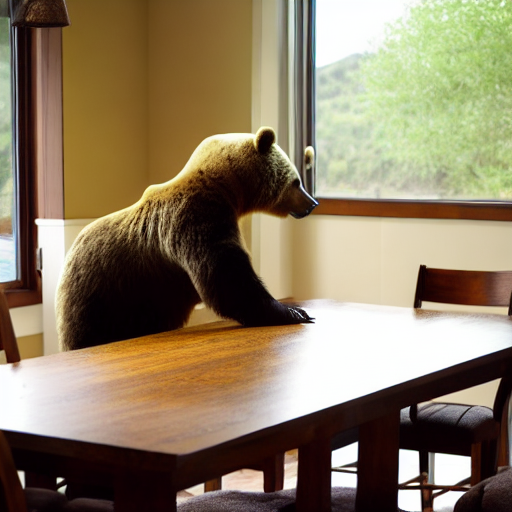} \\
    \texttt{URGE}\par SDv1.5\par $N=4$ &
    \includegraphics[width=\linewidth,keepaspectratio]{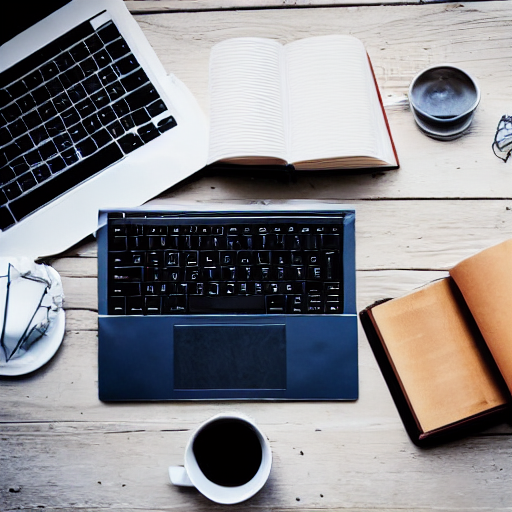} &
    \includegraphics[width=\linewidth,keepaspectratio]{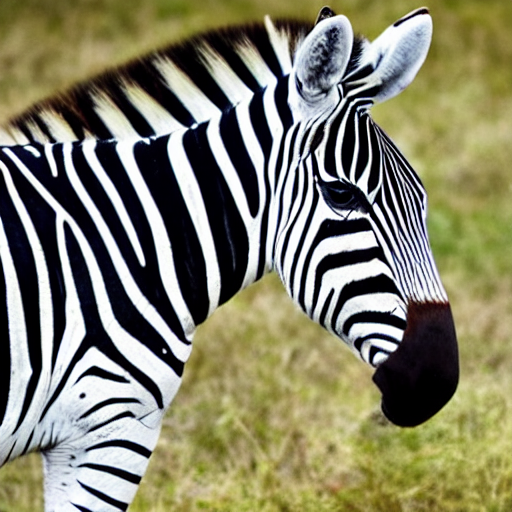} &
    \includegraphics[width=\linewidth,keepaspectratio]{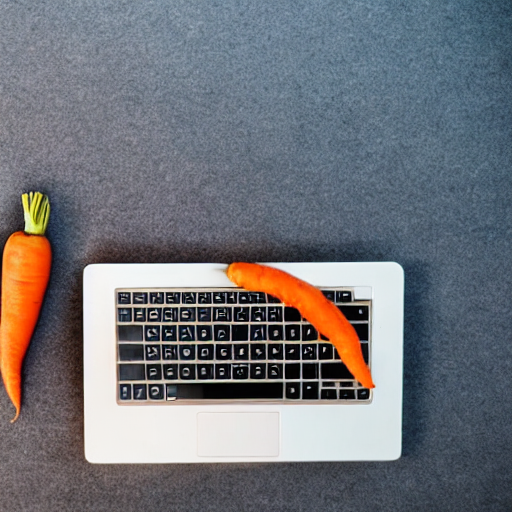} &
    \includegraphics[width=\linewidth,keepaspectratio]{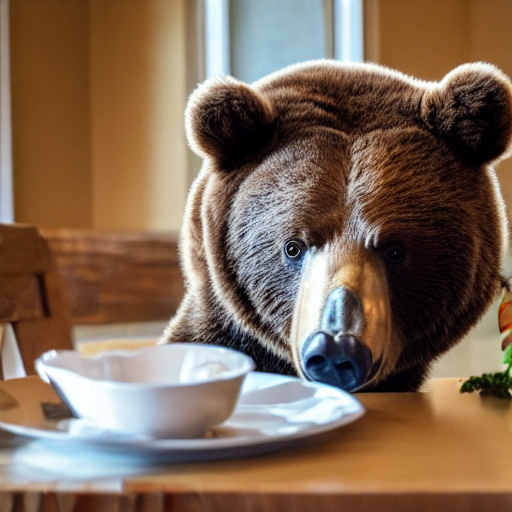} \\
    \texttt{\bfseries Stein}\par SDv1.5\par $K=2$ &
    \includegraphics[width=\linewidth,keepaspectratio]{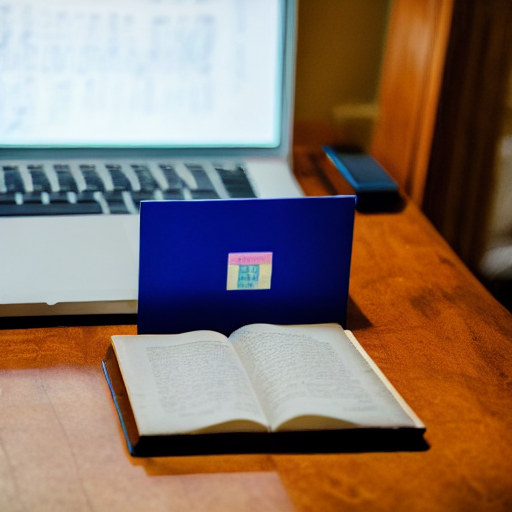} &
    \includegraphics[width=\linewidth,keepaspectratio]{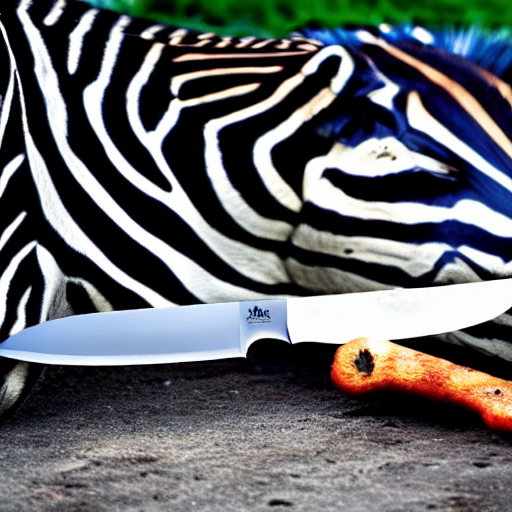} &
    \includegraphics[width=\linewidth,keepaspectratio]{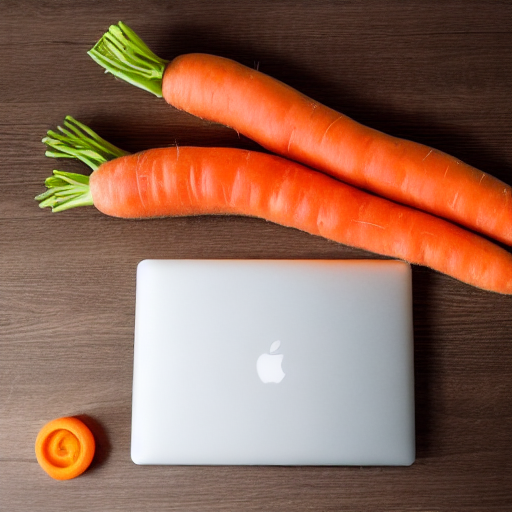} &
    \includegraphics[width=\linewidth,keepaspectratio]{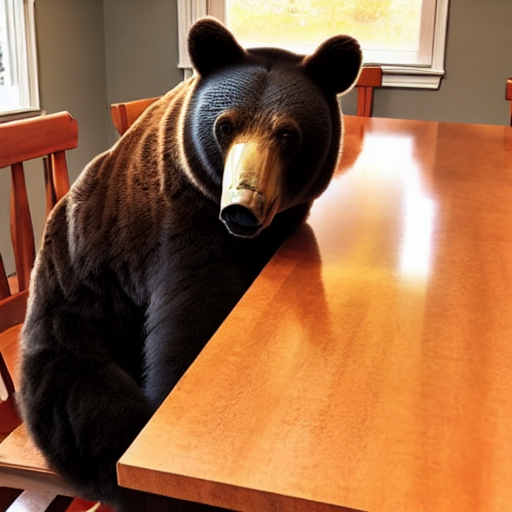} \\
    \texttt{\bfseries Stein}\par SDv1.5\par $K=4$ &
    \includegraphics[width=\linewidth,keepaspectratio]{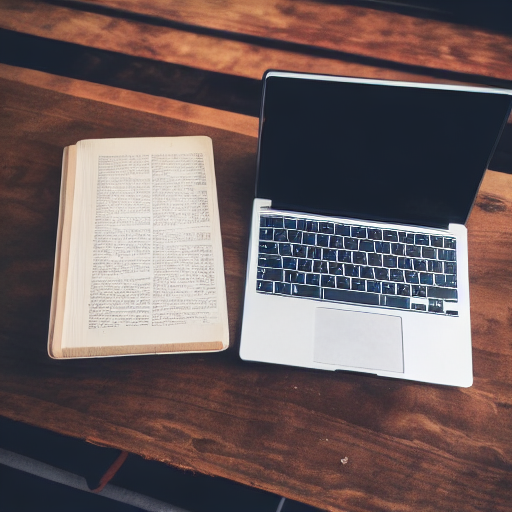} &
    \includegraphics[width=\linewidth,keepaspectratio]{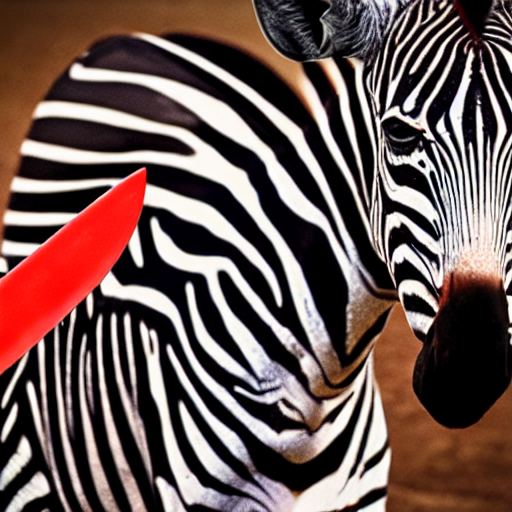} &
    \includegraphics[width=\linewidth,keepaspectratio]{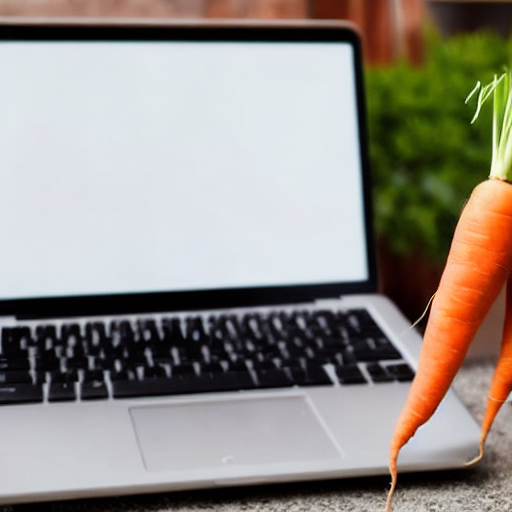} &
    \includegraphics[width=\linewidth,keepaspectratio]{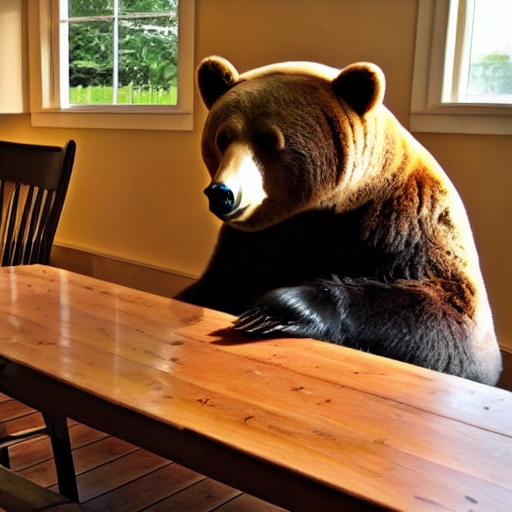} \\
    \end{longtable}
\endgroup

\FloatBarrier
\section{Conclusion}
{\color{black}
We propose Stein Guidance with whole-path MH correction.
The same value queries steer proposals without gradients and provide
path weights for correction, while centering stabilizes these weights.
Under the stated regularity assumptions, a fixed finite-grid sampling
accuracy can be achieved with a number of MH updates that does not grow
with guidance depth, and the error bound improves as intermediate values
approach exact Doob values.
This avoids the depth--width coupling exhibited by our SMC counterexample.
}

\section*{Reproducibility Statement}
 Complete statements, proofs, essential experimental settings, and selected results follow in the appendices. %
\label{main:end}
\nocite{lew2022recursive}
\bibliographystyle{style/iclr2027_conference}
\bibliography{references}
\section*{AI Use Disclosure}
Generative AI assisted with manuscript restructuring, mathematical exposition, some of the proofs, experiment design and analysis, implementation, and figure preparation. The authors take responsibility for the final content, including all proofs, citations, and numerical results.
\clearpage\appendix
\noindent\textbf{Reading the appendices.}
Appendix~\ref{app:related-work} discusses related work and scope.
Appendix~\ref{app:stein_pools} provides the Stein identities and local pool moments.
Appendices~\ref{app:outer_moments} and~\ref{app:outer_mh} contain the complete-weight analysis and the proof of Theorem~\ref{omh:thm:main-MH}.
Appendix~\ref{app:proofs} proves the Brownian resampling obstruction.
Appendices~\ref{app:outer_sir_complexity} and~\ref{app:local_refinement} give terminal importance resampling and the proper weighting of local refinement.
Essential experimental settings and selected supporting results appear in Appendix~\ref{app:supporting-results}.
\section{Related Work and Scope}
\label{app:related-work}
This section relates centered candidate pools and whole-path correction to existing guidance methods.
Value-based decoding uses future values for derivative-free selection \citep{li2024svdd}; twisted and controlled particle methods use approximate backward information \citep{whiteley2014twisted,heng2020controlled}, and URGE uses path correction \citep{wang2026urge}.
Gaussian Stein estimators \citep{stein,nesterov}, multiple try Metropolis \citep{liu,yangliu}, and independence-MH coupling bounds \citep{deligiannidis} supply established ingredients.
Uniform regularity supplies the moments needed for depth-uniform correction. Variation in the accumulated Bellman error and centered pool moments quantify the two contributions to the sampling error bound.
Section~\ref{app:guidance_comparison} records the comparison conditions for Monte Carlo Doob guidance \citep{zhu2026doit}.

The centered norm $\mathcal A_r$ measures variation of the accumulated Bellman error under the ideal value-tilted path law. It is independent of pool size and measures near-Doob accuracy at the path level. Multiplying each exact value by a positive time-dependent factor leaves this variation zero. Uniform regularity guarantees finite moments. Constants can depend on dimension, and a continuous target requires a separate discretization bound.
Centered pools need not improve every antithetic construction: pairs cancel all odd Hermite components, whereas larger centered pools can retain them.
The Brownian SMC obstruction uses full resampling at every step, consistent with the role of resampling intensity for weakly informative potentials \citep{chopin2022resampling}.
The Gaussian experiments measure local-refinement and estimator effects.
The learned-policy RL comparison shows that a small, forward-only candidate pool followed by whole-path MH can substantially improve on uncorrected guidance.
At $L=32$, $K=2$ achieves the lowest observed $Q$-KS at all three targets, with supported gains over DOIT, SVDD, TFG, and TTS-local and competitive or superior accuracy relative to FK and DAS in observed $Q$-KS.

\begin{figure}[htbp]
\centering
\begin{tikzpicture}[
  x=1cm,y=1cm,
  font=\sffamily\fontsize{7.4}{8.7}\selectfont,
  line cap=round,line join=round,
  box/.style={draw=black!45,fill=white,rounded corners=1.5pt,
    line width=.45pt,align=center,inner sep=3pt,minimum height=.85cm},
  flow/.style={-{Latex[length=1.5mm,width=1.0mm]},draw=black!65,line width=.6pt},
  note/.style={align=center,text=black!75,font=\sffamily\fontsize{7.1}{8.3}\selectfont},
  title/.style={anchor=west,font=\sffamily\bfseries\fontsize{8.5}{10}\selectfont},
  steinbox/.style={box,draw=blue!45!black,fill=blue!4,line width=.7pt}
]
\path[use as bounding box] (0,0) rectangle (13.8,8.35);

\draw[draw=black!23,rounded corners=2pt,line width=.45pt]
  (.02,4.95) rectangle (13.78,8.30);
\node[title] at (.25,7.99) {(a) SDG: Stein regularization of clean predictions};
\node[note,anchor=east] at (13.5,7.99) {At each diffusion step};

\node[box,text width=1.30cm] (s0) at (1.03,6.94)
  {Noisy particles\\$\{x_t^i\}$};
\node[box,text width=1.82cm] (s1) at (3.35,6.94)
  {Tweedie\\clean predictions};
\node[steinbox,text width=2.40cm] (s2) at (6.15,6.94)
  {\textbf{Kernelized Stein\\variational inference}\\One SVGD update};
\node[box,text width=1.42cm] (s3) at (8.88,6.94)
  {Re-noise to\\the current level};
\node[box,text width=2.55cm] (s4) at (11.86,6.94)
  {Reward-gradient guidance\\via a new clean prediction};
\draw[flow] (s0)--(s1);
\draw[flow] (s1)--(s2);
\draw[flow] (s2)--(s3);
\draw[flow] (s3)--(s4);

\node[note,text width=10.4cm] at (6.15,5.92)
  {\textbf{Posterior regularization:} move clean predictions toward reference\\
   denoising posteriors using approximate scores and kernel repulsion.};
\draw[flow,dashed,draw=black!40] (s4.south) -- (11.86,5.40)
  -- (1.03,5.40) -- (s0.south);
\node[note,fill=white,inner xsep=4pt] at (6.45,5.40)
  {Continue with the updated noisy particles};

\draw[draw=black!23,rounded corners=2pt,line width=.45pt]
  (.02,.05) rectangle (13.78,4.65);
\node[title] at (.25,4.34) {(b) Ours: value-weighted selection provides Stein guidance};
\node[note,anchor=east] at (13.5,4.34) {$K=2$ shown};

\path[draw=blue!45!black,fill=blue!4,rounded corners=2pt,line width=.6pt]
  (2.05,2.40) rectangle (10.65,4.01);
\node[note,text width=8.10cm,text=blue!45!black] at (6.35,3.73)
  {\textbf{Gaussian Stein identity} links value-weighted selection\\
   to the leading guidance drift in expectation.};

\node[box,text width=1.30cm] (o0) at (1.03,2.99)
  {Current state\\$X_k$};
\node[box,text width=1.91cm] (o1) at (3.39,2.99)
  {Centered candidates\\$Y^+,Y^-$};
\node[box,text width=1.91cm] (o2) at (6.23,2.99)
  {Value queries\\$H^+,H^-$};
\node[box,text width=1.91cm] (o3) at (9.13,2.99)
  {Select one candidate\\$p_\pm\propto H^\pm$};
\node[box,text width=1.91cm] (o4) at (12.30,2.99)
  {Next state\\$X_{k+1}=Y_J$};
\draw[flow] (o0)--(o1);
\draw[flow] (o1)--(o2);
\draw[flow] (o2)--(o3);
\draw[flow] (o3)--(o4);

\node[note,anchor=west,text width=2.85cm,align=left] at (.35,1.35)
  {Reuse these values\\for the correction weight.};
\node[box,text width=2.23cm,minimum height=.77cm] (w0) at (5.03,1.35)
  {Pool average\\$S_k=(H^++H^-)/2$};
\node[box,text width=2.15cm,minimum height=.77cm] (w1) at (8.36,1.35)
  {Accumulate\\path weight $R$};
\node[box,text width=2.15cm,minimum height=.77cm] (w2) at (11.93,1.35)
  {Whole-path\\MH correction};
\draw[flow] (o2.south) -- (6.23,2.08) -| (w0.north);
\draw[flow] (w0)--(w1);
\draw[flow] (w1)--(w2)
  node[midway,above=2pt,note,text width=1.1cm] {Complete\\path};
\node[note,text width=12.8cm] at (6.9,.45)
  {Repeat candidate selection along each independently generated proposal path;
   apply MH after the path is complete.};
\end{tikzpicture}
\caption{{\color{black}\textbf{Where Stein's method enters the two algorithms.}
SDG uses kernelized Stein variational inference to regularize clean predictions toward reference denoising posteriors before reward-gradient guidance.
Our method uses the classical Gaussian Stein identity to relate value-weighted candidate selection to gradient guidance.
The same candidate values provide a path weight for subsequent whole-path MH correction.
The two-candidate case is shown for clarity.}}
\label{fig:sdg-comparison}
\end{figure}
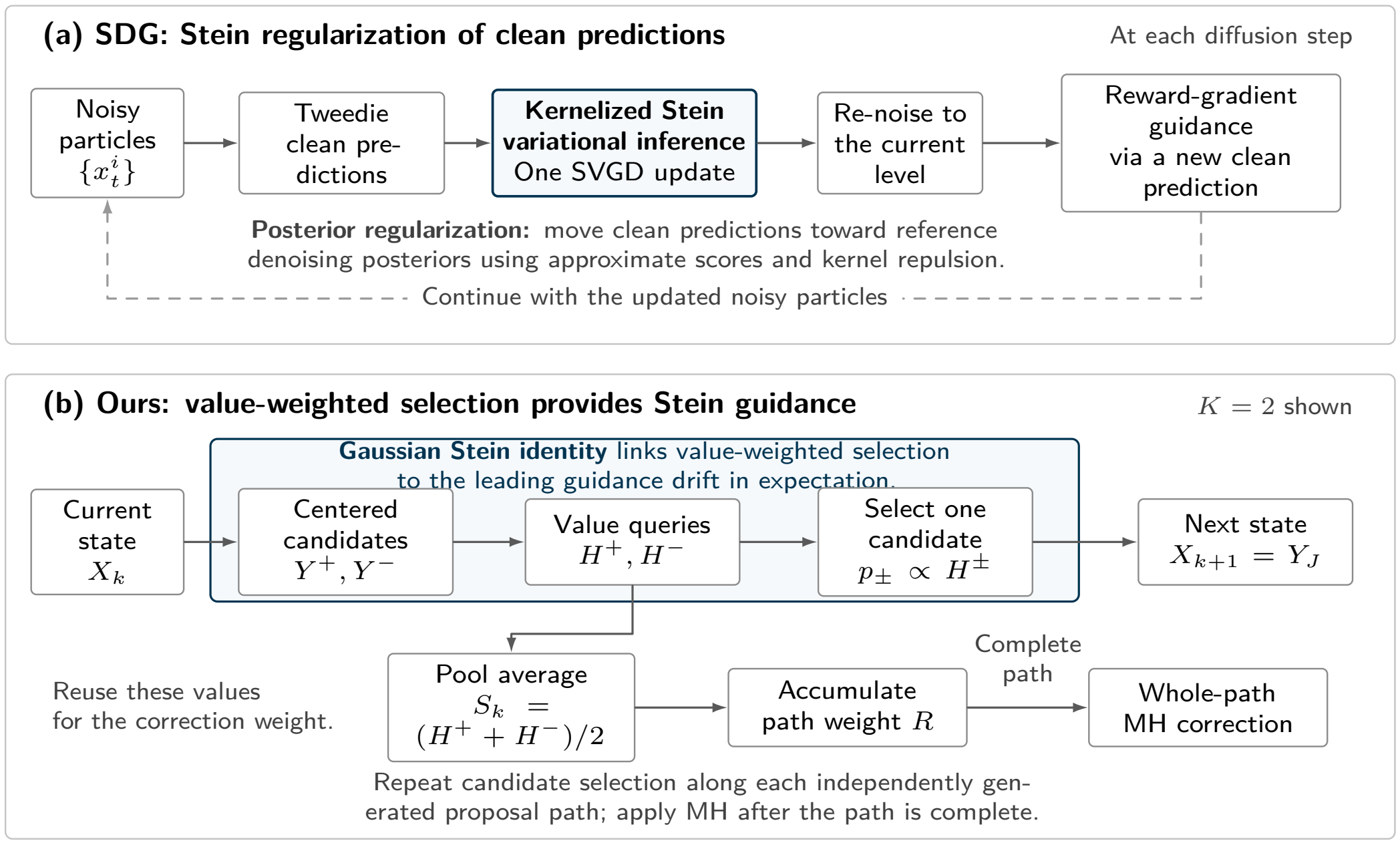

{\color{black}
\paragraph{Stein Diffusion Guidance.}
Stein Diffusion Guidance (SDG)~\citep{nguyen2025sdg} aims to make reward guidance more reliable by regularizing clean predictions (Figure~\ref{fig:sdg-comparison}).
At each diffusion step, it uses Tweedie's formula to obtain clean predictions from a population of noisy particles, applies one Stein variational gradient descent (SVGD) update, and re-noises the updated predictions before continuing reward-gradient guidance.
The SVGD step uses approximate posterior scores and kernel repulsion to move these predictions toward the reference diffusion's denoising posteriors, which describe plausible clean endpoints given the current noisy states.
This regularization is intended to reduce reward overestimation in regions where the reward model is less reliable.

SDG uses a kernelized Stein variational method for posterior regularization, whereas our method uses the classical Gaussian Stein identity (Gaussian integration by parts) to relate value-weighted candidate selection to gradient guidance.
Our guidance selects a next-state candidate from a centered Gaussian pool with probability proportional to its value, without evaluating value gradients.
Centering stabilizes the pool average, and the same value queries supply both the selection probabilities and a computable path weight for subsequent whole-path MH correction.
}

\FloatBarrier

\section{Gaussian Stein Pools: Construction, Moments, and Guidance}
\label{app:stein_pools}\label{app:steinproofs}\label{app:details}
The local cancellation in Algorithm~\ref{alg:stein} supplies the sharper pool
moments used by the finite-moment path analysis. We establish those
bounds here, then establish the estimator and mean-guidance guarantees
for the same candidate construction.
\subsection{Targets and properly weighted candidate pools}\label{sec:setup}
The pool identity below is the common starting point for local and
complete-path correction: weighting a selected candidate by its pool
average recovers the unnormalized value-tilted transition.

Fix a time grid $t_k=kh$, $k=0,\ldots,n$, with $T=nh$ and initial state
$X_0=x_\star$. Let $P_k$ be the transition kernels of a reference simulator.
The reference path law and its positive terminal reward tilt are
\begin{equation}\label{eq:target}
 P_n=\delta_{x_\star}\prod_{k=0}^{n-1}P_k,
 \qquad
 \Pi_n(\dd x_{0:n})=\frac{\phi(x_n)}{Z_n}P_n(\dd x_{0:n}),
 \qquad Z_n=\E_{P_n}\phi(X_n)\in(0,\infty).
\end{equation}
All statements of exactness below concern this fixed grid target. A
discretization error between $P_n$ and a continuous process is a separate
quantity.

Let $H_0,\ldots,H_n$ be positive value functions with $H_n=\phi$. Define
\begin{equation}\label{eq:GH}
 G_k(x)=P_kH_{k+1}(x),\qquad
 \Gamma_k(x)=\frac{G_k(x)}{H_k(x)},\qquad
 K_k^H(x,\dd y)=\frac{P_k(x,\dd y)H_{k+1}(y)}{G_k(x)}.
\end{equation}
We assume $0<G_k(x)<\infty$ on the states under consideration, and write
$\Pi_n^H=\delta_{x_\star}\prod_kK_k^H$. These locally normalized kernels
define the intermediate target of local correction. When the discrete
Bellman equations $H_k=P_kH_{k+1}$ hold, $\Pi_n^H=\Pi_n$.
We use $\norm{\mu-\nu}_{\TV}=\sup_A|\mu(A)-\nu(A)|$.

At a fixed parent $x$, abbreviate $H=H_{k+1}$ and $G=G_k(x)$. A candidate
pool $U=(Y_1,\ldots,Y_K)$ has distribution $q_{\mathrm{pool}}=q_{k,x,K}$, and each of its
coordinates has marginal law $P_k(x,\cdot)$. Dependence within the pool
is allowed. Set
\begin{equation}\label{eq:pool}
 S(U)=\frac1K\sum_{i=1}^K H(Y_i),\qquad
 w(U)=\frac{S(U)}G,\qquad
 \pi_i(U)=\frac{H(Y_i)}{\sum_{j=1}^K H(Y_j)}.
\end{equation}
Let $C_x$ select $Y_J$ with conditional probabilities
$\mathbb P(J=i\mid U)=\pi_i(U)$.
Thus $C_xf(U)=\sum_iH(Y_i)f(Y_i)/(KS(U))$ is the conditional mean of $f(Y_J)$.

\begin{lemma}[Proper weighting of a pool]\label{lem:pool}
For every bounded measurable $f$,
\begin{equation}\label{m:proper}
 \E_{q_{\mathrm{pool}}}[S(U)C_xf(U)]=P_k(Hf)(x).
\end{equation}
Consequently $\E_{q_{\mathrm{pool}}}S=G$, $\E_{q_{\mathrm{pool}}}w=1$, and the probability measure $p_{\mathrm{pool}}=wq_{\mathrm{pool}}$
satisfies $p_{\mathrm{pool}}C_x=K_k^H(x,\cdot)$.
The weight $S(U)$ and selection probabilities require neither derivatives nor an evaluation of $G$.
\end{lemma}
\begin{proof}[Proof of Lemma~\ref{lem:pool}]
By the conditional selection probabilities and the correct marginal
law of every candidate,
\[
 \E_{q_{\mathrm{pool}}}[S C_xf]
 =\frac1K\sum_i\E_{q_{\mathrm{pool}}}[H(Y_i)f(Y_i)]
 =P_k(Hf)(x).
\]
Taking $f=1$ gives $\E_{q_{\mathrm{pool}}}S=G$. Dividing the identity by $G$ proves
$p_{\mathrm{pool}}C_x=K_k^H(x,\cdot)$.
\end{proof}

This identity requires only correct candidate marginals. The Gaussian Stein
identity provides additional structure: it explains the guidance
implicit in weighted selection and identifies dependence patterns that
reduce fluctuations of $S$.

\subsection{Centered Gaussian candidates and implicit guidance}\label{sec:steinconstruction}
Centering couples the candidates without changing their Gaussian marginals.
It removes constant noise from the Stein estimator and linear noise from
the pool average; the identities below also identify the mean guidance
implemented by weighted selection.

For the remainder of the local analysis, take an Euler reference kernel
with mean $\mu=x+hb_k(x)$ and covariance $h\sigma\sigma^\top$, where
$\sigma=\sigma_k(x)\in\R^{d_x\times d}$. Here $d_x$ and $d$ are
the state and noise dimensions, and $\gauss=N(0,I_d)$. Draw independent
$G_1,\ldots,G_K\sim N(0,I_d)$ and construct
\begin{equation}\label{eq:construction}
 \bar G=\frac1K\sum_{i=1}^K G_i,\qquad
 Z_i=\sqrt{\frac K{K-1}}(G_i-\bar G),\qquad
 Y_i=\mu+\sqrt h\,\sigma Z_i,\qquad K\ge2.
\end{equation}
The resulting vectors satisfy
\begin{equation}\label{eq:cov}
 Z_i\sim N(0,I_d),\qquad \sum_i Z_i=0,\qquad
 \Cov(Z_i,Z_j)=-\frac{I_d}{K-1}\quad(i\ne j).
\end{equation}
Thus centering preserves every required marginal after rescaling.
The case $K=2$ is an antithetic pair. The case $K=3$, for example, can
equivalently be generated from independent $G,F\sim N(0,I_d)$ by
\begin{equation}\label{eq:three}
 Z_1=G,\qquad Z_2=-G/2+\sqrt3F/2,\qquad
 Z_3=-G/2-\sqrt3F/2.
\end{equation}
The construction therefore works without a decomposition into pairs.

\begin{proposition}[Stein estimation and mean guidance]\label{prop:stein}
Suppose $H\in C^1$ is positive and, for
$F(z)=H(\mu+\sqrt h\sigma z)$, Gaussian integration by parts is valid.
A sufficient condition is $F\in W^{1,2}(\gauss)$. Define
\begin{equation}\label{eq:vhat}
 \vhat=\frac1{K\sqrt h}\sum_i Z_iH(Y_i),\qquad
 v_h=\sigma^\top\E\nabla H(\mu+\sqrt h\sigma Z),\qquad
 s_h=\frac{v_h}{G},\qquad \shat=\frac{\vhat}{S},
\end{equation}
where $Z\sim N(0,I_d)$ and the displayed expectations exist. Then
$\E\vhat=v_h$. Moreover, for any finite common baseline $B$, even one
depending on the entire pool,
\begin{equation}\label{eq:baseline}
 \vhat=\frac1{K\sqrt h}\sum_i Z_i\{H(Y_i)-B\}.
\end{equation}
Weighted selection from an uncorrected pool satisfies
\begin{equation}\label{eq:meanconditional}
 \E[Y_J\mid U]=\mu+h\sigma\shat,
\end{equation}
whereas the exact local tilt has mean
\begin{equation}\label{eq:targetmean}
 \E_{K_k^H(x,\cdot)}Y
 =\mu+h\sigma s_h
 =\mu+h\sigma\sigma^\top\nabla_\mu
       \log\E H(\mu+\sqrt h\sigma Z),
\end{equation}
whenever the last differentiation under the expectation is justified.
\end{proposition}
\begin{proof}[Proof of Proposition~\ref{prop:stein}]
Direct covariance calculation in~\eqref{eq:construction} gives
\eqref{eq:cov}. Gaussian integration by parts for each standard
Gaussian marginal yields
\[
 \E[Z_iH(Y_i)]
 =\sqrt h\,\sigma^\top\E\nabla H(Y_i).
\]
All these expectations are the same. Averaging proves
$\E\vhat=v_h$. The identity $\sum_iZ_i=0$ proves
\eqref{eq:baseline}, including for a random common baseline.
Furthermore,
\[
 \E[Y_J\mid U]
 =\mu+\sqrt h\sigma\sum_i\pi_iZ_i
 =\mu+h\sigma\frac{\vhat}{S}.
\]
Under the exact tilt the mean is
$\mu+\sqrt h\sigma\E[ZH(Y)]/G$, which equals
$\mu+h\sigma v_h/G$ by the same integration by parts identity.
Differentiation with respect to $\mu$ gives the final expression in
\eqref{eq:targetmean} when the asserted interchange is justified.
\end{proof}

Equation~\eqref{eq:targetmean} differentiates with respect to $\mu$ while
holding $\sigma$ fixed. It does not take the total derivative of
$P_kH_{k+1}(x)$ when the drift and diffusion depend on $x$.
The numerator $\vhat$ is unbiased, while the finite pool ratio $\shat$
is generally biased. The zero sum identity removes a constant baseline
exactly, without querying $H(\mu)$.

The same structure can be read directly from the selection probabilities.
Under Assumption~\ref{ass:theory-regularity},
or more generally whenever the following integration by parts is valid,
\begin{equation}\label{eq:selectionstein}
 \E\sum_i\pi_iZ_i
 =\sqrt h\frac K{K-1}\E\sum_i
       \pi_i(1-\pi_i)\sigma^\top\nabla\log H(Y_i).
\end{equation}
For fixed $K$, as $h\to0$ the factor $K/(K-1)$ cancels the limit
$\sum_i\pi_i(1-\pi_i)\to1-1/K$. With independent Gaussian candidates,
the factor $K/(K-1)$ is absent, and uncorrected weighted selection
attenuates the leading guidance by $1-1/K$. This comparison concerns
selection from a raw pool.

\begin{proof}[Proof of the selection identity~\eqref{eq:selectionstein}]
Let $\rho_{ii}=1$ and $\rho_{ij}=-1/(K-1)$ for $i\ne j$.
Gaussian integration by parts in the independent variables $G_i$
implies, even though the covariance of the $Z_i$ is singular,
\[
 \E[Z_i\pi_i]=\sum_j\rho_{ij}\E[\nabla_{Z_j}\pi_i].
\]
Put $a_j=\sigma^\top\nabla\log H(Y_j)$. Differentiating the quotient
that defines $\pi_i$ gives
\[
 \nabla_{Z_j}\pi_i
 =\sqrt h\,\pi_i(\mathbf1_{i=j}a_i-\pi_j a_j).
\]
After summing over $i$, the coefficient of $\pi_j a_j$ is
$1-\sum_i\rho_{ij}\pi_i$. Since
\[
 \sum_i\rho_{ij}\pi_i
 =\frac{K\pi_j-1}{K-1},
\]
this coefficient is $K(1-\pi_j)/(K-1)$, proving the identity.
Under~Assumption~\ref{ass:theory-regularity}, gradients are bounded as shown below, so
the differentiation and integration by parts are justified by Gaussian
integrability. The same bounds justify dominated convergence as
$h\to0$. For independent candidates, replace $\rho$ by the identity
matrix; the resulting coefficient is $1-\pi_j$, which proves the
stated attenuation of the leading guidance.
\end{proof}

\subsubsection{The $K=2$ case}
Both the proposal and correction weight have a simple form for Stein ($K=2$).
For $K=2$, the centered Gaussian construction is exactly
\[
Z_1=(G_1-G_2)/\sqrt2=:Z,\qquad Z_2=-Z,\qquad Z\sim N(0,I_d).
\]
Its candidate law, categorical selection, and query estimator are therefore identical to the antithetic finite difference in~\eqref{m:twoq_difference}. The two forms of its path weight also agree:
\begin{align*}
\phi(X_n)\prod_{k=0}^{n-1}\frac{S_{2,k}}{H_{k+1}(X_{k+1})}
&=\frac{\prod_{k=0}^{n-1}S_{2,k}}{\prod_{k=1}^{n-1}H_k(X_k)}\\
&=H_0(x_\star)\prod_{k=0}^{n-1}\frac{S_{2,k}}{H_k(X_k)}.
\end{align*}
These identities give an exact antithetic finite-difference representation of Algorithm~\ref{alg:stein} at $K=2$, including its correction weight.

With a bounded third derivative, the familiar central-difference expansion is
\[
\frac{H(\mu+\sqrt h\sigma Z)-H(\mu-\sqrt h\sigma Z)}{2\sqrt h}
=(\sigma Z)^\top\nabla H(\mu)+O(h\|\sigma Z\|^3).
\]
Its leading ratio is $ZZ^\top\sigma^\top\nabla\log H(\mu)$,
whose expectation is $\sigma^\top\nabla\log H(\mu)$.
Proposition~\ref{prop:stein} gives the exact form of Stein's identity behind this
Taylor interpretation. Under Assumption~\ref{ass:theory-regularity},
Theorem~\ref{thm:score} directly gives an
$O(h^2)$ error in the one-step mean displacement, without requiring a
third derivative. A mean displacement bound is distinct from a
second-order weak expansion for every test function.

\subsection{Exact spectrum and sharp normalizer variance}\label{sec:moments}

The sum constraint cancels the first order term of any local value
expansion. A Gaussian Hermite calculation gives a stronger conclusion
with no Taylor remainder. Expand
$F\in L^2(\gauss)$ in the orthonormal multivariate Hermite basis as
$F=\sum_\alpha c_\alpha\psi_\alpha$. Define its energy at degree $\ell$
by $a_\ell=\sum_{|\alpha|=\ell}c_\alpha^2$.

\begin{lemma}[Exact Hermite spectrum of the pool average]\label{lem:hermite}
For the construction~\eqref{eq:construction},
\begin{equation}\label{eq:spectrum}
 \Var\left(\frac1K\sum_iF(Z_i)\right)
 =\sum_{\ell\ge1}\lambda_{K,\ell}a_\ell,\qquad
 \lambda_{K,\ell}=
 \frac{1+(K-1)(-1/(K-1))^\ell}{K}.
\end{equation}
In particular $\lambda_{K,1}=0$, and
$0\le\lambda_{K,\ell}\le1/(K-1)$ for every $\ell\ge2$. Hence
\begin{equation}\label{eq:spectralbound}
 \Var\left(\frac1K\sum_iF(Z_i)\right)
 \le\frac1{K-1}\sum_{\ell\ge2}a_\ell
 \le\frac{\Var_{\gauss}(F)}{K-1}.
\end{equation}
The last inequality also holds componentwise for vector valued $F$.
\end{lemma}
\begin{proof}[Proof of Lemma~\ref{lem:hermite}]
If $Z,Z'$ are jointly standard Gaussian and
$\E[ZZ'^\top]=\rho I_d$, their exponential generating function obeys
\[
 \E\left[e^{t^\top Z-\norm t^2/2}
             e^{s^\top Z'-\norm s^2/2}\right]
 =e^{\rho t^\top s}.
\]
Comparing coefficients gives
$\E[\psi_\alpha(Z)\psi_\beta(Z')]
=\mathbf1_{\alpha=\beta}\rho^{|\alpha|}$. Thus, for a finite
Hermite expansion,
\begin{align*}
 \Var\left(\frac1K\sum_iF(Z_i)\right)
 &=\frac1{K^2}\sum_{i,j}\sum_{\ell\ge1}
      \rho_{ij}^{\ell}a_\ell\\
 &=\sum_{\ell\ge1}
      \frac{1+(K-1)(-1/(K-1))^\ell}{K}a_\ell.
\end{align*}
Finite expansions are dense in $L^2(\gauss)$. The averaging map is
continuous into the joint $L^2$ space, since
$\norm{K^{-1}\sum_iF(Z_i)}_2\le\norm F_{L^2(\gauss)}$.
The formula therefore extends to every $F\in L^2(\gauss)$.
For $K=2$, the eigenvalues are zero at odd degrees and one at even
degrees. For $K>2$, the largest even degree coefficient with
$\ell\ge2$ occurs at $\ell=2$ and is $1/(K-1)$; the odd degree
coefficients are between zero and $1/K$. Degree one vanishes in every
case. These observations prove~\eqref{eq:spectralbound}.
\end{proof}

\begin{theorem}[Sharp curvature bound]\label{thm:curvature}\label{m:curvature}
Let $H\in C^2$ be positive, let
$F(z)=H(\mu+\sqrt h\sigma z)\in W^{2,2}(\gauss)$, and let
$G=\E_{\gauss} F>0$. For the normalized pool weight $w=S/G$,
\begin{equation}\label{eq:sharpvariance}
 {
 \Var(w)\le
 \frac{h^2}{2(K-1)G^2}
 \E\norm{\sigma^\top\nabla^2H(\mu+\sqrt h\sigma Z)\sigma}_F^2.}
\end{equation}
There is no assumption that $H$ is bounded or that $w$ has a uniform
positive lower bound. Equality holds whenever $F$ is a polynomial of
degree at most two. In particular, positive quadratic value functions
show that the constant $1/2$ cannot be reduced uniformly.
\end{theorem}
\begin{proof}[Proof of Theorem~\ref{thm:curvature}]
The normalized Hermite differentiation identity
$\partial_j\psi_\alpha=\sqrt{\alpha_j}\psi_{\alpha-e_j}$
and orthogonality imply
\[
 \E_{\gauss}\norm{\nabla^2F}_F^2
 =\sum_\alpha |\alpha|(|\alpha|-1)c_\alpha^2
 =\sum_{\ell\ge2}\ell(\ell-1)a_\ell.
\]
The identity holds first for polynomials and then for
$W^{2,2}(\gauss)$ by approximation in that Sobolev norm.
Because $\ell(\ell-1)\ge2$ for $\ell\ge2$,
Lemma~\ref{lem:hermite} yields
\[
 \Var(S)\le\frac1{2(K-1)}\E_{\gauss}\norm{\nabla^2F}_F^2.
\]
The chain rule gives
$\nabla^2F=h\sigma^\top\nabla^2H(\mu+\sqrt h\sigma Z)\sigma$.
Division by $G^2$ proves the result. If $F$ has degree at most two,
the only nonconstant degree surviving the pool average is degree two,
and $\lambda_{K,2}=1/(K-1)$, so both inequalities are equalities.
A positive nonconstant quadratic with nonzero curvature gives a
nonzero equality case.
\end{proof}

For later path bounds, the useful uniform assumption is a relative
curvature bound:
\begin{equation}\label{eq:relativecurvature}
 \sup_{k,x,h}
 \frac{\E\norm{\sigma_k(x)^\top\nabla^2H_{k+1}
 (\mu_k(x)+\sqrt h\sigma_k(x)Z)\sigma_k(x)}_F^2}{G_k(x)^2}
 \le\mathcal C_2^2.
\end{equation}
Together with the regularity in Theorem~\ref{thm:curvature}, it yields
$\sup_{k,x}\Var(w)\le\mathcal C_2^2h^2/[2(K-1)]$.
The Hessian appears only in the analysis; the proposal queries values of $H$.

\subsection{All fixed moments from the same regularity conditions}
For each output-TV order $p$, the path analysis uses several fixed pool-moment orders
to separate Bellman error from candidate noise. We derive every fixed real order $\ell\ge2$ from
the bounded-value and bounded-Hessian conditions; higher derivatives
are unnecessary. The main theorem combines these local moments with the direct complete-weight
bound in Theorem~\ref{omh:thm:regularity-moments}. The lower bound obtained along the way is useful for
the local-refinement construction.

We use Assumption~\ref{ass:theory-regularity} throughout this subsection. Write
$m_\ell =(\E|N(0,1)|^\ell )^{1/\ell}$ and
$M_{\ell,d}=(\E\norm{N(0,I_d)}^\ell )^{1/\ell}$.
\begin{lemma}[Gaussian rotation inequality]\label{lem:rotation}
Let $G$ be a standard Gaussian vector in any finite dimension. If
$\ell \ge2$ and $f\in W^{1,\ell}$ with respect to its Gaussian measure, then
\begin{equation}\label{eq:rotation}
 \norm{f(G)-\E f(G)}_{L^\ell}
 \le\frac\pi2 m_\ell \norm{\norm{\nabla f(G)}}_{L^\ell}.
\end{equation}
\end{lemma}
\begin{proof}
For smooth $f$, take an independent copy $G'$ and put
$G_t=G\cos t+G'\sin t$ and $V_t=-G\sin t+G'\cos t$.
For each $t$, the vectors $G_t,V_t$ are independent standard
Gaussian vectors. Conditional Jensen, the fundamental theorem of
calculus and Minkowski give
\begin{align*}
 \norm{f(G)-\E f(G)}_\ell 
 &\le\norm{f(G)-f(G')}_\ell \\
 &\le\int_0^{\pi/2}\norm{\nabla f(G_t)^\top V_t}_\ell \,\dd t\\
 &=\frac\pi2 m_\ell \norm{\norm{\nabla f(G)}}_\ell .
\end{align*}
The last equality conditions on $G_t$: the inner product is a scalar
Gaussian with conditional standard deviation $\norm{\nabla f(G_t)}$.
Truncation and smooth approximation in Gaussian $W^{1,\ell}$ extend the
inequality to its stated domain.
\end{proof}

\begin{theorem}[Uniform bounds for all fixed moments]\label{thm:moments}\label{omh:thm:local-high}
Under~Assumption~\ref{ass:theory-regularity}, for every $K\ge2$ and real $\ell \ge2$,
\begin{equation}\label{eq:poolmoment}
\begin{gathered}
 w\ge c_0/C_0,\qquad
 \norm{w-1}_{L^\ell (q_{\mathrm{pool}})}\le B_\ell \frac h{\sqrt{K-1}},\\
 B_\ell =\frac{\pi m_\ell M_{\ell,d}L_2\Sigma^2}{2c_0}.
\end{gathered}
\end{equation}
In particular,
\begin{equation}\label{omh:eq:Bp-explicit}
 B_\ell \le\frac\pi2\frac{L_2\Sigma^2}{c_0}
             \sqrt{(\ell +2)(d+\ell +2)}.
\end{equation}
The constants are independent of $K$ and $h$.
\end{theorem}
\begin{proof}[Proof of Theorem~\ref{thm:moments}]
Set $c_K=\sqrt{K/(K-1)}$. Regard $S$ as a function of the independent
Gaussian input $\mathbf G=(G_1,\ldots,G_K)$. Then
\[
 \nabla_{G_i}S
 =\frac{c_K\sqrt h}{K}\sigma^\top
    \left(\nabla H(Y_i)-\frac1K\sum_j\nabla H(Y_j)\right).
\]
The average minimizes the sum of squared Euclidean distances.
Using this fact and the Hessian bound gives
\begin{align*}
 \norm{\nabla_{\mathbf G}S}^2
 &\le\frac{c_K^2h\Sigma^2}{K^2}
       \sum_i\norm{\nabla H(Y_i)-\nabla H(\mu)}^2\\
 &\le\frac{c_K^2h^2L_2^2\Sigma^4}{K^2}
       \sum_i\norm{Z_i}^2.
\end{align*}
For $\ell \ge2$, Minkowski in $L^{\ell /2}$ yields
\[
 \norm{\left(\sum_i\norm{Z_i}^2\right)^{1/2}}_\ell 
 \le\sqrt K M_{\ell,d}.
\]
Consequently Lemma~\ref{lem:rotation} and $\E_{q_{\mathrm{pool}}}S=G$ imply
\[
 \norm{S-G}_\ell 
 \le\frac\pi2 m_\ell M_{\ell,d}L_2\Sigma^2
       \frac h{\sqrt{K-1}}.
\]
All Sobolev requirements hold because $S$ is bounded and the
displayed derivative bound has every finite Gaussian moment.
Finally $G\ge c_0$ gives the relative moment estimate, and
$S\ge c_0$, $G\le C_0$ give $w\ge c_0/C_0$.

For its explicit dependence on $\ell$, Gaussian integration gives
\[
 m_\ell =\left\{\frac{2^{\ell /2}\Gamma((\ell +1)/2)}{\sqrt\pi}\right\}^{1/\ell},
 \qquad
 M_{\ell,d}=\left\{\frac{2^{\ell /2}\Gamma((d+\ell )/2)}{\Gamma(d/2)}\right\}^{1/\ell}.
\]
Let $s=\lceil \ell /2\rceil$. Monotonicity of $L^\ell$ norms and the even
moment identities imply
\[
 m_\ell \le m_{2s}
 =\left\{\prod_{j=1}^s(2j-1)\right\}^{1/(2s)}
 \le\sqrt{\ell +2},
 \quad
 M_{\ell,d}\le
 \left\{\prod_{j=0}^{s-1}(d+2j)\right\}^{1/(2s)}
 \le\sqrt{d+\ell +2}.
\]
Substitution proves~\eqref{omh:eq:Bp-explicit}.
\end{proof}

The proof differentiates the pool average with respect to its independent
Gaussian inputs. Centering leaves differences of gradients, which provide
the extra factor $\sqrt h$. Every fixed finite moment is covered,
including at $K=2$; there is no restriction $\ell <K$.

Assumption~\ref{ass:theory-regularity} is sufficient for these moment bounds.
The generic outer-MH inequalities in Appendix~\ref{app:outer_mh} use moments
of the normalized complete weight. The path moment bound permits
conditional pool moments to be supplied directly, without a uniform positive
lower bound on the normalized pool weight. The weaker relative
$L^2$ curvature bound~\eqref{eq:relativecurvature} alone does not
provide all higher moments.

\subsection{Stein ratio accuracy and mean displacement}\label{sec:score}
The guidance estimator and the corrected output have different error
metrics. Here we quantify the estimator's variance and bias, then show
how multiplication by the step size reduces its effect on the selected
candidate's mean.

The same candidate geometry controls the guidance estimator. Under
Assumption~\ref{ass:theory-regularity}, let $\norm{\nabla H}_\infty\le L_1$. On the full
Euclidean state space one may already take
$L_1=\sqrt{2L_2(C_0-c_0)}$.
The bound is written separately to expose the dependence on gradient
size.

To see the claimed gradient bound, use the full Euclidean state space.
If $L_2>0$, the descent inequality at
$y=x-\nabla H(x)/L_2$ gives
\[
 c_0\le H(y)
 \le H(x)-\frac{\norm{\nabla H(x)}^2}{2L_2}
 \le C_0-\frac{\norm{\nabla H(x)}^2}{2L_2}.
\]
Thus $\norm{\nabla H(x)}^2\le2L_2(C_0-c_0)$. If $L_2=0$, $H$ is
affine and bounded on the full state space, hence constant.

\begin{theorem}[Stein variance, ratio bias and mean error]\label{thm:score}
Let $\nu=K-1$, $V=L_1^2\Sigma^2d(d+2)$ and
$A_s=dL_2^2\Sigma^4/2$. Under~Assumption~\ref{ass:theory-regularity},
\begin{align}
 \E\norm{\vhat-v_h}^2&\le\frac V\nu,\label{eq:vvar}\\
 \norm{\E\shat-s_h}
 &\le\frac{\sqrt{VA_s}}{c_0^2}\frac h\nu
       +\frac{\Sigma L_1A_s}{c_0^3}\frac{h^2}\nu,
       \label{eq:scorebias}\\
 \E\norm{\shat-s_h}^2
 &\le\frac{2V}{c_0^2\nu}
       +\frac{2\Sigma^2L_1^2A_s}{c_0^4}\frac{h^2}\nu.
       \label{eq:scoremse}
\end{align}
Consequently, for $0<h\le h_0$, uncorrected weighted selection obeys
\begin{equation}\label{eq:meanerror}
 \norm{\E_{q_{\mathrm{pool}}}Y_J-\E_{K_k^H(x,\cdot)}Y}
 \le C\frac{h^2}{K-1},
\end{equation}
where $C$ depends only on $c_0,L_1,L_2,\Sigma,d,h_0$.
\end{theorem}
\begin{proof}[Proof of Theorem~\ref{thm:score}]
The baseline identity writes $\vhat=K^{-1}\sum_iF_v(Z_i)$, where
\[
 F_v(z)=\frac z{\sqrt h}\{H(\mu+\sqrt h\sigma z)-H(\mu)\},
 \qquad \norm{F_v(z)}\le L_1\Sigma\norm z^2.
\]
Sum~\eqref{eq:spectralbound} over its coordinates. Since
$\E\norm Z^4=d(d+2)$, this proves
\[
 \E\norm{\vhat-v_h}^2\le\frac V\nu.
\]
Theorem~\ref{thm:curvature} and
$\norm{\sigma^\top\nabla^2H\sigma}_F^2\le dL_2^2\Sigma^4$
give $\Var(S)\le A_sh^2/\nu$. Also $\norm{v_h}\le\Sigma L_1$.

Because $S,G\ge c_0$,
\[
 \norm{1/S-1/G}_2\le c_0^{-2}\sqrt{\Var(S)},
 \qquad
 \E\frac1S-\frac1G=\E\frac{(S-G)^2}{G^2S}.
\]
The second equality follows by expanding $1/S$ around $G$ and using
$\E(S-G)=0$. Hence
\begin{align*}
 \E\shat-s_h
 &=\E\left[(\vhat-v_h)(1/S-1/G)\right]
     +v_h\left(\E(1/S)-1/G\right),\\
 \norm{\E\shat-s_h}
 &\le c_0^{-2}
       \sqrt{\E\norm{\vhat-v_h}^2\Var(S)}
       +c_0^{-3}\norm{v_h}\Var(S).
\end{align*}
Substitution proves~\eqref{eq:scorebias}. The decomposition
\[
 \shat-s_h=\frac{\vhat-v_h}{S}
          +\frac{v_h(G-S)}{SG}
\]
and $\norm{a+b}^2\le2\norm a^2+2\norm b^2$ prove
\eqref{eq:scoremse}. Finally, subtracting
\eqref{eq:targetmean} from the expectation of
\eqref{eq:meanconditional} multiplies the ratio bias by $h\sigma$.
Using $h\le h_0$ proves~\eqref{eq:meanerror}, for example with
$C=\Sigma\sqrt{VA_s}/c_0^2+
       h_0\Sigma^2L_1A_s/c_0^3$.
\end{proof}

\section{Complete-Path Weights and Moment Bounds}
\label{app:outer_moments}
We first prove the complete-path importance-weight identity. Uniform spatial regularity then supplies all fixed moments of a complete proposal weight. A change of measure separates finite-pool noise from the accumulated Bellman error. The resulting bound on centered moments quantifies the improvement when this error varies little across ideal tilted paths and yields the correction bounds in Appendix~\ref{app:outer_mh}.

\subsection{From a weighted pool to the target path}\label{sec:global}
The following theorem justifies using the weight returned by Algorithm~\ref{alg:stein} in whole-path MH.
It shows that reweighting the augmented proposal recovers the target path law.
We use the standing assumptions of Section~\ref{m:method}, including the fixed start and $H_n=\phi$.
The statement also allows nonuniform step sizes $h_k$ and pool sizes $K_k$.
{\color{black} Let $\cQ$ denote the proposal law on the augmented space of candidate pools, selected indices, and paths.}
Each step uses fresh Gaussian inputs and a fresh selection uniform, independent of the preceding history.

\begin{theorem}[Exact augmented correction]\label{omh:thm:proper-global}
Every bounded measurable path function $F$ satisfies
\begin{equation}\label{m:weight}
R=H_0(x_\star)\prod_{k=0}^{n-1}\frac{S_k}{H_k(X_k)},\qquad
\E_{\cQ}[RF]=\E_{P_n}[\phi(X_n)F].
\end{equation}
In particular,
\begin{equation}\label{omh:eq:weight-normalization}
\E_{\cQ}[R]=Z_n,\qquad W=\frac{R}{Z_n},\qquad \E_{\cQ}[W]=1.
\end{equation}
Thus $\cP=W\cQ$ is a probability law. For every measurable set of paths $A$,
\[
\cP(X_{0:n}\in A)=\Pi_n(A).
\]
This is the physical marginal: after the candidate pools and selected indices are integrated out, the selected path $X_{0:n}$ has law $\Pi_n$.
\end{theorem}
\begin{proof}[Proof of Theorem~\ref{omh:thm:proper-global}]
Equation~\eqref{m:proper} makes the weighted child transition $P_k(H_{k+1}f)/H_k$. Iteration for nonnegative product cylinder functions telescopes the intermediate values, leaving $H_n=\phi$ and cancellation of the initial $H_0$. Taking $F=1$ gives finite mass $Z_n$; a monotone-class argument extends the identity to bounded measurable path functions.
In particular, $\E_{\cQ}R=Z_n$ and $\cP=(R/Z_n)\cQ$ has physical marginal $\Pi_n$.
\end{proof}

For the moment bounds below, we factor the same weight into pool and Bellman terms.
At a parent $x$, let $C_x$ select a child from a pool with probabilities proportional to $H_{k+1}(Y_i)$. Correct candidate marginals give
\begin{equation}\label{eq:globalpoolidentity}
 \E_q[S_k(U)C_xf(U)]
 =\frac1{K_k}\sum_{i=1}^{K_k}\E H_{k+1}(Y_i)f(Y_i)
 =P_k(H_{k+1}f)(x).
\end{equation}
Here $q$ denotes the pool law at this parent. Write
\begin{equation}\label{eq:action}
 \Gamma_k=G_k/H_k,\qquad
 B(X_{0:n})=\sum_{k=0}^{n-1}\log\Gamma_k(X_k).
\end{equation}
Here $\log\Gamma_k$ is the local Bellman error and $B$ is the accumulated Bellman error along the path. The centered norm $\mathcal A_r$ in~\eqref{m:action} measures the variation of $B$ around a constant under $\Pi_n^H$. With normalized pool averages $w_k=S_k/G_k(X_k)$, the complete weight factors as
\begin{equation}\label{eq:globalweight}
 R=H_0(x_\star)\prod_{k=0}^{n-1}\frac{S_k}{H_k(X_k)}
   =H_0(x_\star)e^B\prod_{k=0}^{n-1}w_k.
\end{equation}

\subsection{Complete-weight moments from spatial regularity}
\label{omh:sec:regularity-moments}
Telescoping against the selected values gives a second factorization that does not involve the accumulated Bellman error. It proves the depth-uniform moment bounds used in the main theorem, using only the candidate marginals and their common Gaussian scale.

\begin{theorem}[Complete-path moments from spatial regularity]
\label{omh:thm:regularity-moments}
Under~\eqref{eq:theory-regularity}, the fixed start and terminal agreement $H_n=\phi$, let $\sum_k h_k=T$ and allow arbitrary pool sizes $K_k\ge2$. For every real $\ell\ge2$,
\begin{equation}\label{omh:eq:regularity-moments}
 \E_{\cQ}W^\ell\le\mathsf M_\ell,
 \qquad
 \mathsf M_\ell=\left(\frac{C_0}{c_0}\right)^\ell e^{a_\ell T},
\end{equation}
where
\begin{equation}\label{omh:eq:regularity-constants}
 V=\frac{2L_2(C_0-c_0)\Sigma^2d}{c_0^2},
 \qquad
 a_\ell=\frac{\ell(\ell-1)}2
             \left(\frac{C_0}{c_0}\right)^{\ell-2}V.
\end{equation}
These constants are independent of the numerical depth and pool sizes. The result also holds for uncentered pools with the same Gaussian candidate marginals.
\end{theorem}
\begin{proof}
Let $\Fil_k$ contain the history before pool $k$ is drawn, and write
$H_{k,i}=H_{k+1}(Y_{k,i})$, $S_k=K_k^{-1}\sum_iH_{k,i}$.
For the selected index $J_k$, define
\begin{equation}\label{omh:eq:selected-martingale}
 \eta_k=\frac{S_k}{H_{k,J_k}},\qquad
 N_j=\prod_{k<j}\eta_k,\qquad N_0=1.
\end{equation}
Conditional on $\Fil_k$ and the pool $U_k$, selection has probability
$H_{k,i}/(K_kS_k)$, and hence
\[
 \E[\eta_k\mid\Fil_k,U_k]
 =\sum_i\frac{H_{k,i}}{K_kS_k}\frac{S_k}{H_{k,i}}=1.
\]
Thus $N_j$ is a positive mean-one martingale. Cancellation of the selected values in~\eqref{eq:globalweight} gives the pathwise identity
\begin{equation}\label{omh:eq:selected-telescope}
 R=H_n(X_n)N_n=\phi(X_n)N_n.
\end{equation}

The global Hessian and value bounds imply
$\|\nabla H_{k+1}\|^2\le L_g^2:=2L_2(C_0-c_0)$.
Indeed, for $L_2>0$ the descent inequality at
$x-\nabla H_{k+1}(x)/L_2$ gives
$c_0\le H_{k+1}(x)-\|\nabla H_{k+1}(x)\|^2/(2L_2)$;
if $L_2=0$, boundedness makes the affine value constant.
Writing $\mu_k$ for the pool's common frozen mean, we obtain
\begin{align*}
 \E[(\eta_k-1)^2\mid\Fil_k,U_k]
 &=\frac1{K_kS_k}\sum_i\frac{(S_k-H_{k,i})^2}{H_{k,i}}\\
 &\le\frac1{K_kc_0^2}\sum_i
       \{H_{k,i}-H_{k+1}(\mu_k)\}^2.
\end{align*}
The last inequality uses the minimizing property of the sample mean.
Each candidate marginal is $\mu_k+\sqrt{h_k}\sigma_kZ$ with
$Z\sim N(0,I_d)$, so the gradient bound gives
\[
 \E[(\eta_k-1)^2\mid\Fil_k]
 \le\frac{L_g^2h_k}{c_0^2}\E\|\sigma_kZ\|^2
 \le Vh_k.
\]
No independence or centering within the pool is used.
Since $c_0/C_0\le\eta_k\le C_0/c_0$, Taylor's theorem for
$z^\ell$ about $1$, together with the conditional mean-one identity, gives
\[
 \E[\eta_k^\ell\mid\Fil_k]\le1+a_\ell h_k.
\]
Iteration yields
$\E N_n^\ell\le\prod_k(1+a_\ell h_k)\le e^{a_\ell T}$.
Finally, $c_0\le\phi\le C_0$ implies $c_0\le Z_n\le C_0$.
Equation~\eqref{omh:eq:selected-telescope} therefore gives
$W\le(C_0/c_0)N_n$ and proves~\eqref{omh:eq:regularity-moments}.
\end{proof}
This argument controls complete weights directly. It needs neither temporal regularity of the values nor a drift bound: candidates are compared with their common frozen mean. It supplies no mesh-vanishing coefficient by itself. The following change-of-measure argument quantifies such a coefficient through centered moments of the accumulated Bellman error.

\subsection{The accumulated Bellman error under the ideal tilted path}\label{sec:bellman}
The ideal transitions $K_k^H(x,\dd y)=H_{k+1}(y)P_k(x,\dd y)/G_k(x)$ define
$\Pi_n^H=\delta_{x_\star}\prod_kK_k^H$. Telescoping gives
\begin{equation}\label{omh:eq:ideal-density}
 L:=\frac{\dd\Pi_n}{\dd\Pi_n^H}
 =\frac{H_0(x_\star)}{Z_n}e^B
 =\frac{e^B}{\E_{\Pi_n^H}e^B},\qquad
 \Lambda_s:=\|L\|_{L^s(\Pi_n^H)}.
\end{equation}
All centered norms of the accumulated Bellman error in this appendix use this ideal path law. Neither $L$ nor $\mathcal A_r$ depends on the pool size when the reference grid and value functions are fixed.

\begin{lemma}[Ideal density moments and Bellman error bounds]
\label{omh:lem:ideal-density-moments}
Under~\eqref{eq:theory-regularity}, the fixed start and $H_n=\phi$, let
$\rho=C_0/c_0$ and use $V,a_s$ from~\eqref{omh:eq:regularity-constants}.
For every real $s\ge2$ and $1\le r<\infty$,
\begin{align}
 \Lambda_s&\le\overline\Lambda_s:=\rho e^{a_sT/s},
 &\E_{\Pi_n^H}L^{-1}&\le\rho e^{VT},
 \label{omh:eq:ideal-density-moments}\\
 \mathcal A_r&\le\|\log L\|_{L^r(\Pi_n^H)}
 \le\frac re(1+\rho e^{VT})^{1/r}.
 \label{omh:eq:finite-action-existence}
\end{align}
Thus the stated regularity conditions bound every required centered norm of the accumulated Bellman error and every required density moment, with constants independent of $n$ and $K$.
\end{lemma}
\begin{proof}
Under $\Pi_n^H$, put
$\eta_k^H=G_k(X_k)/H_{k+1}(X_{k+1})$ and
$N_j^H=\prod_{k<j}\eta_k^H$. Relative to the physical path history
$\mathcal F_k^H=\sigma(X_0,\ldots,X_k)$, the ideal transition gives
$\E[\eta_k^H\mid\mathcal F_k^H]=1$. Moreover,
$\rho^{-1}\le\eta_k^H\le\rho$, and
\[
 \E[(\eta_k^H-1)^2\mid X_k=x]
 =\frac1{G_k(x)}P_k\!\left(\frac{(G_k(x)-H_{k+1})^2}{H_{k+1}}\right)(x)
 \le c_0^{-2}\Var_{P_k(x,\cdot)}H_{k+1}\le Vh_k.
\]
The last inequality uses the global gradient bound established in the proof of Theorem~\ref{omh:thm:regularity-moments}, comparing the Gaussian candidate with its frozen mean. Taylor expansion of $z^s$ about $1$ gives
$\E[(\eta_k^H)^s\mid\mathcal F_k^H]\le1+a_sh_k$ for $s\ge2$.
Conditional iteration therefore yields $\E(N_n^H)^s\le e^{a_sT}$.
The pathwise identity
\[
 N_n^H=\frac{H_0(x_\star)}{\phi(X_n)}e^B,
 \qquad L=\frac{\phi(X_n)}{Z_n}N_n^H
\]
and $\phi/Z_n\le\rho$ prove the positive-moment bound.

For the inverse factor the same kernel gives
\[
 \E[(\eta_k^H)^{-1}\mid X_k=x]
 =\frac{P_k(H_{k+1}^2)(x)}{G_k(x)^2}
 =1+\frac{\Var_{P_k(x,\cdot)}H_{k+1}}{G_k(x)^2}
 \le1+Vh_k.
\]
Iterating this inequality, without claiming that the inverse product is a mean-one martingale, proves $\E(N_n^H)^{-1}\le e^{VT}$ and hence the inverse-density bound. Taking
${\color{black}a}=\log\E_{\Pi_n^H}e^B$ in the definition of $\mathcal A_r$ gives its first bound. Finally, for $z>0$,
$|\log z|^r\le(r/e)^r(z+z^{-1})$; this follows from
$\sup_{u\ge0}u^re^{-u}=(r/e)^r$ on each side of $z=1$.
Use $\E L=1$ and the inverse-density bound to conclude.
\end{proof}
The upper bound in~\eqref{omh:eq:finite-action-existence} establishes finiteness uniformly in numerical depth. A small $\mathcal A_r$ additionally requires that the accumulated Bellman error vary little across paths under $\Pi_n^H$.

\begin{proposition}[Accumulated Bellman error and path discrepancy]
\label{omh:prop:bellman}\label{prop:bellman}
Whenever $\Lambda_2$ and $\mathcal A_2$ are finite,
\begin{equation}\label{omh:eq:bellman-TV}
 \|\Pi_n^H-\Pi_n\|_{\TV}\le\min\{1,2\Lambda_2\mathcal A_2\}.
\end{equation}
Let $\Pi_{n,k}^H$ be the $k$th marginal of $\Pi_n^H$. For any $r\ge1$, if
\begin{equation}\label{omh:eq:local-action-cert}
 \inf_{a\in\mathbb R}
 \|\log\Gamma_k-a\|_{L^r(\Pi_{n,k}^H)}\le h_k\epsilon_{r,k},
\end{equation}
then $\mathcal A_r\le\sum_kh_k\epsilon_{r,k}$. In particular,
$\epsilon_{r,k}\le\epsilon_r$ gives $\mathcal A_r\le T\epsilon_r$.
\end{proposition}
\begin{proof}
Let $X,X'$ be independent ideal paths, with corresponding $B,B'$ and $L,L'$.
Since $\E L'=1$, conditional Jensen and
$|e^u-e^v|\le(e^u+e^v)|u-v|$ imply
\[
 \tfrac12\E|L-1|
 \le\tfrac12\E|L-L'|
 \le\tfrac12\E[(L+L')|B-B'|]
 \le2\Lambda_2\mathcal A_2.
\]
The last step uses Cauchy--Schwarz, $\|L+L'\|_2\le2\Lambda_2$,
and $\|B-B'\|_2\le2\inf_{{\color{black}a}}\|B-{\color{black}a}\|_2$. TV is also at most one.
For the bound from local residuals, choose deterministic centers $a_k$ arbitrarily close to the displayed infima. Minkowski gives
\[
 \mathcal A_r\le\left\|\sum_k(\log\Gamma_k(X_k)-a_k)\right\|_r
 \le\sum_k\|\log\Gamma_k-a_k\|_{L^r(\Pi_{n,k}^H)}.
\]
Letting the centering errors tend to zero proves the claim; no independence across time is used.
\end{proof}
The distribution in~\eqref{omh:eq:local-action-cert} is the ideal path marginal. A bound under the reference or finite-pool proposal marginal requires a separate comparison.

\subsection{The conditional-moment interface}
To make the dependence on the accumulated Bellman error explicit, we use conditional moments of the normalized pool averages. Uniform regularity supplies these at every fixed order, but more general proposals can use the same bounds. They are distinct from the selected-index moments proved above. We write $\ell\ge2$ for the moment order throughout Appendices~\ref{app:steinproofs}--\ref{app:outer_sir_complexity}; the main decay order $p$ uses $\ell=p+1$.
\begin{assumption}[Conditional moments at a fixed order]\label{omh:ass:moment}
Let $\Fil_k$ contain the proposal history before pool $k$ is drawn. At a specified order $\ell \ge2$, deterministic finite $u_{\ell,k}\ge0$ satisfy
\[
 \E[w_k\mid\Fil_k]=1,\qquad
 \E[|w_k-1|^\ell \mid\Fil_k]\le u_{\ell,k}^\ell .
\]
\end{assumption}
The mean-one identity follows from the candidate marginals. Under~\eqref{eq:theory-regularity}, Theorem~\ref{omh:thm:local-high} gives
$u_{\ell,k}=B_\ell h_k/\sqrt{K_k-1}$ for every fixed real $\ell \ge2$, with no derivatives above order two and no restriction $\ell <K_k$.
The weaker relative-curvature condition~\eqref{eq:relativecurvature} supplies only
$u_{2,k}^2=\mathcal C_2^2h_k^2/[2(K_k-1)]$.
Neither the product argument nor outer MH requires a positive lower bound on $w_k$.
Define the accumulated local fluctuation budget
\begin{equation}\label{omh:eq:Vbudget}
 \Vbudget_\ell=\sum_{k=0}^{n-1}\log\{1+(\ell-1)u_{\ell,k}^2\}.
\end{equation}

\subsection{A centered product-martingale bound}
Conditional centering, rather than independence across time, controls the accumulated pool noise. The next inequality keeps the bound centered at one as the proposal approaches its target.

\begin{lemma}[$L^\ell$ quadratic smoothness]
\label{omh:lem:smoothness}
Let $\ell \ge2$ and $X,Y\in L^\ell$. If $X$ is $\Fil$-measurable and
$\E[Y\mid\Fil]=0$, then
\begin{equation}\label{omh:eq:smoothness}
 \norm{X+Y}_\ell^2\le\norm X_\ell^2+(\ell -1)\norm Y_\ell^2.
\end{equation}
\end{lemma}
\begin{proof}
For $\ell =2$, expand the square and use $\E XY=0$. For $\ell >2$, set
$a(t)=\E|X+tY|^\ell$ and $f_\eta(t)=(a(t)+\eta^\ell )^{2/\ell}$, where
$\eta>0$. H\"older's inequality and domination by powers of
$|X|+|Y|$ justify differentiation. The squared-first-derivative term
in $f_\eta''$ is nonpositive, because $2/\ell -1<0$. Therefore
\begin{align*}
 f_\eta''(t)
 &\le2(\ell -1)(a(t)+\eta^\ell )^{2/\ell -1}
           \E\{|X+tY|^{\ell -2}Y^2\}\\
 &\le2(\ell -1)(a(t)+\eta^\ell )^{-(\ell -2)/\ell}
                a(t)^{(\ell -2)/\ell}\norm Y_\ell^2
 \le2(\ell -1)\norm Y_\ell^2.
\end{align*}
Conditional centering gives
$a'(0)=\ell \E[|X|^{\ell -2}XY]=0$, so $f_\eta'(0)=0$.
Integrating on $[0,1]$ gives
$f_\eta(1)\le f_\eta(0)+(\ell -1)\norm Y_\ell^2$.
Letting $\eta\downarrow0$ proves the result.
\end{proof}

\begin{theorem}[Product moments at one fixed order]
\label{omh:thm:product}
Under Assumption~\ref{omh:ass:moment} at order $\ell \ge2$, the product
$M_j=\prod_{k<j}w_k$, $M_0=1$, is a positive mean-one martingale.
Writing $\Vbudget_\ell =\sum_k\log\{1+(\ell -1)u_{\ell,k}^2\}$,
\begin{equation}\label{omh:eq:product-bounds}
 \norm{M_n}_\ell^2\le e^{\Vbudget_\ell},
 \qquad \norm{M_n-1}_\ell^2\le e^{\Vbudget_\ell}-1.
\end{equation}
\end{theorem}
\begin{proof}
The filtration $\Fil_k$ contains the generated history up to the
current parent, before pool $k$ is drawn. Proper weighting gives
$\E[w_k\mid\Fil_k]=1$. Put $Y_k=M_k(w_k-1)$. Since $M_k$ is
$\Fil_k$-measurable, the conditional moment assumption gives
\[
 \E[Y_k\mid\Fil_k]=0,
 \qquad
 \E|Y_k|^\ell
 =\E\{M_k^\ell \E[|w_k-1|^\ell \mid\Fil_k]\}
 \le u_{\ell,k}^\ell \E M_k^\ell .
\]
Starting with $M_0=1$, induction establishes finiteness of all moments
used below. Let $b_k=(\ell -1)u_{\ell,k}^2$. Applying
Lemma~\ref{omh:lem:smoothness} to $M_{k+1}=M_k+Y_k$ gives
\[
 \norm{M_{k+1}}_\ell^2
 \le(1+b_k)\norm{M_k}_\ell^2
 \le\prod_{j=0}^k(1+b_j).
\]
Also $\E[M_{k+1}\mid\Fil_k]=M_k$, which proves the mean-one
martingale assertion. Applying the same lemma to
$M_{k+1}-1=(M_k-1)+Y_k$ and summing gives
\begin{align*}
 \norm{M_n-1}_\ell^2
 &\le\sum_{k=0}^{n-1}b_k\norm{M_k}_\ell^2
 \le\sum_{k=0}^{n-1}b_k\prod_{j<k}(1+b_j)\\
 &=\prod_{k=0}^{n-1}(1+b_k)-1=e^{\Vbudget_\ell}-1.
\end{align*}
This proves both bounds without independent increments or an increasing
sequence of moment orders along the path.
\end{proof}

\subsection{Change of measure and centered complete-path weights}
The pool martingale and the ideal density depend on the same physical path. The following change of measure separates their moments without assuming independence.

\begin{lemma}[Finite-moment separation]\label{omh:lem:finite-separation}
Put $M=M_n$ and $\widehat{\cQ}=M\cQ$. This is a probability law whose physical path marginal is $\Pi_n^H$, and
\begin{equation}\label{omh:eq:W-decomposition}
 W=ML.
\end{equation}
For every real $\ell\ge2$ for which the right-hand side is finite,
\begin{equation}\label{omh:eq:finite-separation}
 \|W-1\|_{L^\ell(\cQ)}
 \le\|M-1\|_{L^\ell(\cQ)}
 +4\Lambda_{4\ell}
 \|M\|_{L^{4\ell-3}(\cQ)}^{(4\ell-3)/(4\ell)}\mathcal A_{2\ell}.
\end{equation}
\end{lemma}
\begin{proof}
The product martingale gives $\E_{\cQ}M=1$. Dividing the weighted-pool identity~\eqref{eq:globalpoolidentity} by $G_k(x)$ gives
$\E_q[w_kC_xf]=K_k^Hf(x)$. Iterating this identity proves that the physical marginal of $M\cQ$ is $\Pi_n^H$. Equations~\eqref{eq:globalweight} and~\eqref{omh:eq:ideal-density} then give $W=ML$.

Extend $\widehat{\cQ}$ by an independent physical path $X'\sim\Pi_n^H$ and denote its values by $B',L'$. Since $\E L'=1$, Jensen conditional on the augmented state and the exponential difference inequality give
\begin{align*}
 \E_{\cQ}|M(L-1)|^\ell
 &=\E_{\widehat{\cQ}}M^{\ell-1}|L-1|^\ell\\
 &\le\E_{\widehat{\cQ}\otimes\Pi_n^H}
 M^{\ell-1}(L+L')^\ell|B-B'|^\ell.
\end{align*}
Apply H\"older with exponents $4,4,2$ and then take the $\ell$th root. The three factors are bounded respectively by
\[
 \bigl(\E_{\widehat{\cQ}}M^{4\ell-4}\bigr)^{1/(4\ell)}
 =\|M\|_{L^{4\ell-3}(\cQ)}^{(4\ell-3)/(4\ell)},
 \quad2\Lambda_{4\ell},\quad2\mathcal A_{2\ell}.
\]
For the last factor subtract an arbitrary deterministic center from each of $B,B'$ and take the infimum. The marginal identity just proved justifies both ideal-path norms; it does not make $M$ and $L$ independent. Finally, use
$W-1=(M-1)+M(L-1)$ and Minkowski.
\end{proof}

\begin{theorem}[Complete-path moments with Bellman error dependence]
\label{omh:thm:global-moments}\label{thm:globalchi}
Suppose Assumption~\ref{omh:ass:moment} holds at orders $\ell$ and $4\ell-3$, where $\ell\ge2$, and $\Lambda_{4\ell},\mathcal A_{2\ell}<\infty$. Then
\begin{equation}\label{omh:eq:Delta-definition}
 \|W-1\|_{L^\ell(\cQ)}\le\deltab_\ell
 :=\sqrt{e^{\Vbudget_\ell}-1}
 +4\Lambda_{4\ell}
 e^{(4\ell-3)\Vbudget_{4\ell-3}/(8\ell)}\mathcal A_{2\ell}.
\end{equation}
In particular $\E_{\cQ}W^\ell\le(1+\deltab_\ell)^\ell$, and at order two
\begin{equation}\label{omh:eq:Dstar}
 D=\chi^2(\cP\Vert\cQ)=\E_{\cQ}(W-1)^2
 \le\dbound:=\deltab_2^2.
\end{equation}
Independently of the centered moment bound, for any $\ell\ge2$ the local moment at order $2\ell-1$ and $\Lambda_{2\ell}<\infty$ give
\begin{equation}\label{omh:eq:separated-full-moment}
 \E_{\cQ}W^\ell
 \le\Lambda_{2\ell}^\ell
 e^{(2\ell-1)\Vbudget_{2\ell-1}/4}.
\end{equation}
If $B$ is constant, then $L=1$ and $W=M$. This case needs only the local moment at order $\ell$ and gives
$\|W-1\|_\ell\le\sqrt{e^{\Vbudget_\ell}-1}$.
In particular a conditional variance bound $v_k$ gives
$D\le\prod_k(1+v_k)-1$. Under the relative-curvature condition on a uniform grid,
\begin{equation}\label{eq:globalexact}
 D\le\exp\left\{\frac{\mathcal C_2^2Th}{2(K-1)}\right\}-1.
\end{equation}
\end{theorem}
\begin{proof}
Insert the two product bounds from Theorem~\ref{omh:thm:product} into Lemma~\ref{omh:lem:finite-separation}. Minkowski gives the uncentered moment, and $\E W=1$ identifies the variance with the centered second moment. For~\eqref{omh:eq:separated-full-moment}, change measure once and use Cauchy--Schwarz:
\[
 \E_{\cQ}W^\ell
 =\E_{\widehat{\cQ}}M^{\ell-1}L^\ell
 \le(\E_{\cQ}M^{2\ell-1})^{1/2}
      (\E_{\Pi_n^H}L^{2\ell})^{1/2}
 \le e^{(2\ell-1)\Vbudget_{2\ell-1}/4}\Lambda_{2\ell}^\ell.
\]
Constant $B$ makes $L=1$ by normalization, so the sharper assertions follow directly from the product theorem without higher-order moment conditions on the density or accumulated Bellman error. Finally, take
$v_k=\mathcal C_2^2h^2/[2(K-1)]$ and use $\log(1+v)\le v$ and $nh^2=Th$.
\end{proof}

\begin{corollary}[Uniform scaling with Bellman error]
\label{omh:cor:weight-scaling}
Under~\eqref{eq:theory-regularity}, on a uniform grid $h=T/n$ with pool size $K\ge2$, all conditions of Theorem~\ref{omh:thm:global-moments} hold at every fixed real $\ell\ge2$. The local moment bounds give
\begin{equation}\label{omh:eq:Vp-explicit}
 \Vbudget_\ell\le(\ell-1)B_\ell^2\frac{Th}{K-1}
 \le\frac{\pi^2L_2^2\Sigma^4}{4c_0^2}
 (\ell-1)(\ell+2)(d+\ell+2)\frac{Th}{K-1}.
\end{equation}
There is $D_\ell<\infty$, depending only on $\ell,T,d,c_0,C_0,L_2,\Sigma$, such that
\begin{equation}\label{omh:eq:finite-weight-scaling}
 \|W-1\|_\ell\le\deltab_\ell
 \le D_\ell\left(\mathcal A_{2\ell}+\sqrt{\frac{Th}{K-1}}\right).
\end{equation}
Consequently,
\begin{equation}\label{omh:eq:D-near}
 D\le\dbound\le2D_2^2\left(\mathcal A_4^2+\frac{Th}{K-1}\right).
\end{equation}
\end{corollary}
\begin{proof}
Theorem~\ref{omh:thm:local-high} supplies each required local moment, and Lemma~\ref{omh:lem:ideal-density-moments} supplies the ideal density moments and centered norms of the accumulated Bellman error. Set $\delta^2=Th/(K-1)\le T^2$ and $\gamma_s=(s-1)B_s^2$. Summing the local bounds with $\log(1+x)\le x$ proves~\eqref{omh:eq:Vp-explicit}. Since $e^v-1\le ve^v$, one admissible constant is
\begin{equation}\label{omh:eq:finite-scaling-constant}
 D_\ell=\max\left\{
 \sqrt{\gamma_\ell}\,e^{\gamma_\ell T^2/2},\quad
 4\overline\Lambda_{4\ell}
 e^{(4\ell-3)\gamma_{4\ell-3}T^2/(8\ell)}\right\}.
\end{equation}
Substitute into~\eqref{omh:eq:Delta-definition}, then square and use $(a+b)^2\le2(a^2+b^2)$ at order two.
\end{proof}
The constants depend on the fixed moment order and can grow rapidly with it. The local variance scale $h^2/(K-1)$ accumulates to $Th/(K-1)$ despite dependence across time; no conditional centering of $\log w_k$ is needed.

\subsection{Initialization and value function requirements}
The fixed starting point and terminal value determine which factors appear in the weight. This subsection records the adjustments needed when either is changed.

\begin{proposition}[Random initialization]
\label{omh:prop:initial}
Let the reference start from $\mu_0$ and the proposal draw
$X_0\sim\nu_0$, where $\mu_0$ and $\nu_0$ are equivalent and
$\ell_0=\dd\mu_0/\dd\nu_0$ is finite and positive almost surely.
For $P_n^{\mu_0}=\mu_0\prod_kP_k$ and
$\Pi_n^{\mu_0}=Z_n^{-1}\phi(X_n)P_n^{\mu_0}$, the proper weight is
\begin{equation}\label{omh:eq:random-initial-R}
 R=\ell_0(X_0)H_0(X_0)
       \prod_{k=0}^{n-1}\frac{S_k(U_k)}{H_k(X_k)}.
\end{equation}
Moment bounds for this initialization require control of its additional initial density factor.
\end{proposition}
\begin{proof}
Conditioning on $X_0=x$, the fixed-initial-state proper-weighting
identity gives, for bounded physical path functions $F$,
\[
 \E_{\cQ}[RF]
 =\int\ell_0(x)\nu_0(\dd x)
       \E_{P_n^x}[\phi(X_n)F]
 =\E_{P_n^{\mu_0}}[\phi(X_n)F].
\]
In particular, $\E R=Z_n$. Including $X_0$ in the filtration
preserves $\E[w_k\mid\Fil_k]=1$, so $M_n$ remains mean one.
The selected-value telescoping identity becomes
$R=\ell_0(X_0)\phi(X_n)N_n$. Thus any moment claim must also account for $\ell_0(X_0)$; proper weighting alone does not supply its moments. The generic outer-MH results continue to apply whenever their weight-moment hypotheses are verified.
\end{proof}

If $\nu_0=\mu_0$ and the values solve the exact discrete Doob equations,
the incremental Bellman sum is zero but
$W=H_0(X_0)M_n/Z_n$. The initial value need not be constant, so the
fixed-start exact-Doob rate does not automatically extend to this
initialization. Sampling instead from
$\nu_0(\dd x)=H_0(x)\mu_0(\dd x)/\mu_0(H_0)$ makes
$\ell_0H_0$ constant, when $0<\mu_0(H_0)<\infty$; the cost of
sampling that initial tilt must be counted separately.

The terminal condition $H_n=\phi$ is also substantive. If the queried
terminal value differs, proper weighting requires the factor
$\phi(X_n)/H_n(X_n)$. Its moments require separate control. The value functions
and complete-proposal mechanism are fixed across outer proposals.
Value functions fitted on separate data can be handled conditionally on those
data. Adaptation from the proposals being corrected, fresh noisy evaluations of the value functions, or numerical clipping do not inherit the weighting identity
without an additional argument.

\subsection{An unbounded-value diffusion with all fixed moments}
\label{omh:sec:brownian}
Uniform value bounds are sufficient rather than necessary. This example verifies the conditional moments directly.

\begin{proposition}[Brownian exponential tilt]
\label{omh:prop:brownian}
Take $P_k(x,\cdot)=N(x,h)$, $X_0=0$, $T=nh$, and
$\phi(x)=e^{\beta x}$ with $\beta\ne0$. The exact discrete Doob
values are $H_k(x)=\exp\{\beta x+\beta^2(T-t_k)/2\}$. For $K=2$,
\begin{equation}\label{omh:eq:brownian-xi}
 w_k=\xi_h(Z_k)=e^{-\beta^2h/2}\cosh(\beta\sqrt h Z_k),
 \qquad Z_k\stackrel{\rm iid}{\sim}N(0,1),
\end{equation}
and
\begin{equation}\label{omh:eq:brownian-D}
 D=\{\cosh(\beta^2h)\}^n-1
   =\frac{\beta^4T}{2}h+O(h^2).
\end{equation}
For every fixed finite $\ell \ge2$ and $0<h\le h_0$,
$\norm{\xi_h-1}_\ell \le C_{\ell,\beta,h_0}h$, uniformly in the parent.
Thus this model satisfies the conditional-moment assumptions at every
fixed finite order although $H$ has neither a global upper bound nor a
global positive lower bound.
\end{proposition}
\begin{proof}
Gaussian integration gives $P_kH_{k+1}=H_k$. Averaging the two values
$H_{k+1}(x\pm\sqrt h Z_k)$ and dividing by $H_k(x)$
proves~\eqref{omh:eq:brownian-xi}. The parent cancels, so fresh Gaussian
draws make these particular factors independent across time despite
the selections. With $u=\beta^2h$, the Gaussian moment-generating
function gives $\E\xi_h=1$ and $\E\xi_h^2=\cosh u$.
Here $B=0$ and $W=\prod_k\xi_h(Z_k)$, proving the exact second
moment. The expansion follows from
$\log\cosh u=u^2/2+O(u^4)$ and $n=T/h$.

For the fixed-order moment bound, use
$\cosh t-1\le t^2\cosh|t|/2$ and $1-e^{-u/2}\le u/2$.
If $u\le u_0=\beta^2h_0$, then
\[
 |\xi_h(Z)-1|
 \le\frac u2\{1+Z^2\cosh(\sqrt{u_0}Z)\}.
\]
The bracket has every finite Gaussian moment, so one may take
$C_{\ell,\beta,h_0}=(\beta^2/2)
 \norm{1+Z^2\cosh(|\beta|\sqrt{h_0}Z)}_\ell <\infty$.
The bound does not depend on the parent, proving the required
conditional moment estimate directly.
\end{proof}
In this example, exact values eliminate the accumulated Bellman error but leave
nonzero finite-pool noise. Appendix~\ref{app:brownian} gives the exact
variance formulas for general pool size $K$.

\section{Sampling Error Bounds for Outer Metropolis--Hastings}
\label{app:outer_mh}

This appendix proves Theorem~\ref{omh:thm:main-MH}. Generic MH moment bounds give a depth-uniform baseline. A bound on the centered complete weight makes the dependence on the variation of the accumulated Bellman error and on pool size explicit. All required moments follow from the stated spatial regularity.
Coupling and moment bounds for independence MH with unbounded weights are
established tools; see \citet{deligiannidis}. We give the constants used here
because their dependence on the centered weight moments determines the
near-Doob diffusion rates.

\subsection{Generic setting and an ordered coupling}
The coupling isolates the initial excess target mass and follows how long that mass can remain uncorrected.
\label{omh:sec:MH-general}
Let $q$ be a probability measure on a standard Borel space, let
$0<w<\infty$ $q$-almost surely, and suppose $\E_qw=1$.
Set $\pi=wq$, and let $\MH$ be independence MH with proposal $q$ and
acceptance probability $\min\{1,w(y)/w(x)\}$. Throughout this appendix the
initial law is $q$. For an integer $m\ge0$, define
\begin{equation}\label{omh:eq:generic-quantities}
\begin{aligned}
d_m&=\norm{q\MH^m-\pi}_{\TV},\\
s&=\E_q(w-1)_+=\E_q(1-w)_+=\norm{q-\pi}_{\TV},
\qquad c=\E_q\min(w,1)=1-s>0.
\end{aligned}
\end{equation}
When finite, write $D=\E_q(w-1)^2$. For a possible weight value $a>0$,
define its rejection probability by
\begin{equation}\label{omh:eq:rejection-function}
r(a)=\E_q(1-w/a)_+,
\qquad 1-r(a)=\E_q\min(1,w/a)>0.
\end{equation}
The variable $a$ is an argument, not a lower bound on the weights.

\begin{theorem}[Proposal-initialized coupling bound]
\label{omh:thm:coupling}
For every integer $m\ge0$,
\begin{equation}\label{omh:eq:coupling-bound}
d_m\le\E_q[(w-1)_+r(w)^m].
\end{equation}
For every $a>1$,
\begin{equation}\label{omh:eq:reject-ratio}
1-r(a)\ge\frac ca,
\qquad r(a)\le\frac{a-1+s}{a},
\qquad \frac{r(a)}{1-r(a)}\le\frac{a-1+s}{c}.
\end{equation}
\end{theorem}
\begin{proof}
If $s=0$, then $q=\pi$ and every error is zero. Otherwise, couple
$X_0\sim q$ and $Y_0\sim\pi$ by making them equal with probability $c$,
with common law $c^{-1}\min(1,w)q$. On the remaining event draw their
states from the residual laws
\[
\frac{(1-w)_+}{s}q
\quad\hbox{and}\quad
\frac{(w-1)_+}{s}q,
\]
respectively. Their weights then satisfy $w(X_0)<1<w(Y_0)$ almost surely.

Use the same independent proposal and uniform random number for both
chains at every update. Until the chains meet, the higher-weight chain
stays at $Y_0$. Indeed, if it accepts a proposal $Z$, the lower-weight
chain also accepts because its acceptance threshold is no smaller. If
the higher-weight chain rejects, then $w(Z)<w(Y_0)$, so the lower chain's
retained weight remains below $w(Y_0)$ whether it accepts or rejects.
This proves the ordering and freezing by induction. Once the chains
meet, keep them together.

Conditionally on a residual $Y_0=y$, failure to meet in $m$ updates has
probability $r(w(y))^m$. The $Y$-chain is stationary, and the coupling
inequality gives
\[
d_m\le s\int r(w(y))^m\frac{(w(y)-1)_+}{s}\,q(\dd y),
\]
proving~\eqref{omh:eq:coupling-bound}.
For $a>1$, $\min(a,w)\ge\min(1,w)$ gives $1-r(a)\ge c/a$.
Also
\[
a r(a)=\E_q(a-w)_+=a-1+\E_q(w-a)_+\le a-1+s.
\]
Combining these two estimates proves~\eqref{omh:eq:reject-ratio}.
\end{proof}

\subsection{One update and stationary rejection}
One update already converts weight variance into output TV. Stationary rejection is a separate quantity, which the same weights also control.
\begin{proposition}[One-step bias and stationary rejection]
\label{omh:prop:one-step}\label{lem:oneglobal}
If $D<\infty$, then
\begin{equation}\label{omh:eq:one-step}
d_1\le\E_q\frac{(w-1)_+^2}{w}\le D.
\end{equation}
For independent $U,U'\sim q$, the stationary rejection probability is
\begin{equation}\label{omh:eq:stationary-rejection}
\rho_{\rm stat}=\frac12\E|w(U)-w(U')|\le\sqrt{D/2}.
\end{equation}
\end{proposition}
\begin{proof}
At a point of weight $a>0$, the density of $q\MH$ with respect to $q$ is
\begin{equation}\label{eq:onedensity}
f_1(a)=r(a)+\E_q\min(1,a/w)
=1+\E_q\frac{a-w}{\max(a,w)}.
\end{equation}
The function $v\mapsto(a-v)/\max(a,v)$ is convex: it equals $1-v/a$
below $a$ and $a/v-1$ above $a$, with matching one-sided derivatives.
Jensen and $\E_qw=1$ therefore imply
\[
f_1(a)\ge
\begin{cases}
a,&a\le1,\\
2-a^{-1},&a\ge1.
\end{cases}
\]
Thus $(a-f_1(a))_+\le(a-1)_+^2/a$. Integrating this positive density
deficit proves the first inequality in~\eqref{omh:eq:one-step}; the second
uses $1/w\le1$ on $\{w>1\}$.

At stationarity, the average acceptance probability is
\[
\iint\pi(\dd x)q(\dd y)\min(1,w(y)/w(x))
=\E\min(w(U),w(U')).
\]
The identity $\min(a,b)=(a+b-|a-b|)/2$ gives the stated rejection
probability. Independence and Cauchy--Schwarz give
$\E|w(U)-w(U')|\le\sqrt{2D}$.
\end{proof}
The stationary rejection bound concerns a chain already at its invariant
law; it is distinct from the proposal-initialized error after one update.

\subsection{A centered second-moment bound for every update count}
A variance bound alone gives a bound valid at every positive update count, without any lower bound on the weight.
\begin{theorem}[All-$m$ second-moment guarantee]
\label{omh:thm:MH-second}
If $D<\infty$, then for every integer $m\ge1$,
\begin{equation}\label{omh:eq:MH-second}
d_m\le\min\{s,D,3D/m\}.
\end{equation}
If $D\le1$, the last term can be replaced by $2D/m$.
\end{theorem}
\begin{proof}
TV contraction and invariance give $d_m\le d_0=s$ and
$d_m\le d_1\le D$. For $0\le r<1$ and integer $m\ge1$,
\[
mr^m\le\sum_{j=1}^m r^j\le\frac r{1-r}.
\]
Theorem~\ref{omh:thm:coupling} therefore gives
\begin{equation}\label{omh:eq:variance-integrand}
md_m\le\E_q\left[(w-1)_+\frac{r(w)}{1-r(w)}\right].
\end{equation}
Let $D_+=\E_q(w-1)_+^2$ and $D_-=\E_q(1-w)_+^2$.
Then $D=D_++D_-$, $s^2\le D_-$, and $s\le\sqrt D/2$ because
$2s=\E_q|w-1|$. If $D\le1$, then $c\ge1/2$, and
\eqref{omh:eq:reject-ratio} bounds the expectation in
\eqref{omh:eq:variance-integrand} by
\[
\frac{D_++s^2}{c}\le\frac Dc\le2D.
\]
If $D>1$, use $\min(1,w/a)\ge w/(a+w)$ and Jensen under $\pi$ to obtain
\[
1-r(a)\ge\E_q\frac w{a+w}
=\E_\pi\frac1{a+w}
\ge\frac1{a+\E_\pi w}
=\frac1{a+1+D}.
\]
Hence $r(a)/(1-r(a))\le a+D$, and the expectation in
\eqref{omh:eq:variance-integrand} is at most
\[
\E_q[(w-1)_+(w+D)]=D_++(1+D)s
\le D+(1+D)\le3D,
\]
where $s\le1$ and $D>1$ were used.
\end{proof}
The numerator is the centered variance $D$, rather than $1+D$.
No uniform upper or positive lower bound on $w$ is required.

\subsection{Higher centered moments and explicit overlap constants}
Higher moments improve the power of the update budget. We use $\ell$ for moment order, so the corresponding decay power is $\ell-1$.
\begin{theorem}[High-order near-target MH bound]
\label{omh:thm:MH-high}
Let $\ell\ge2$ and suppose $\E_q|w-1|^\ell<\infty$.
For every integer $m\ge \ell-1$,
\begin{equation}\label{omh:eq:MH-high}
d_m\le
\left(\frac{2(\ell-1)}{cm}\right)^{\ell-1}\E_q(w-1)_+^\ell.
\end{equation}
If $D\le1$, the prefactor can be replaced by
$[4(\ell-1)/m]^{\ell-1}$. More generally,
\begin{equation}\label{omh:eq:overlap-lower}
c\ge\frac1{2+D}.
\end{equation}
\end{theorem}
\begin{proof}
Put $\alpha=\ell-1\ge1$. For $m\ge\alpha$ and $0\le r<1$,
\begin{equation}\label{omh:eq:power-rejection}
r^m\le\left(\frac\alpha m\right)^\alpha
\left(\frac r{1-r}\right)^\alpha.
\end{equation}
For $0<r<1$, maximize $r^{m-\alpha}(1-r)^\alpha$; its maximum is
\[
\left(\frac{m-\alpha}{m}\right)^{m-\alpha}
\left(\frac\alpha m\right)^\alpha
\le\left(\frac\alpha m\right)^\alpha.
\]
The endpoint $m=\alpha$ follows by continuity; $r=0$ is immediate.
Set $Z=(w-1)_+$. The coupling bound and
\eqref{omh:eq:reject-ratio} give
\[
d_m\le\left(\frac\alpha{cm}\right)^\alpha
\E_q[Z(Z+s)^\alpha].
\]
By convexity, $(Z+s)^\alpha\le2^{\alpha-1}(Z^\alpha+s^\alpha)$.
Since $s=\E_qZ$, Jensen gives $s^{\alpha+1}\le\E_qZ^{\alpha+1}$, whence
\[
\E_q[Z(Z+s)^\alpha]
\le2^{\alpha-1}\{\E_qZ^{\alpha+1}+s^{\alpha+1}\}
\le2^\alpha\E_qZ^\ell.
\]
This proves~\eqref{omh:eq:MH-high}. The small-$D$ simplification uses
$c\ge1/2$. For the general overlap bound, Cauchy--Schwarz applied to
\[
1=\E_qw
=\E_q\left[\sqrt{\min(w,1)}\,
\frac{w}{\sqrt{\min(w,1)}}\right]
\]
gives
\[
1\le c\,\E_q\frac{w^2}{\min(w,1)}
=c\,\E_q[w\mathbf1_{\{w\le1\}}+w^2\mathbf1_{\{w>1\}}]
\le c(2+D).
\]
\end{proof}

\begin{remark}[Moment order and update count]\label{omh:rem:m-p}
The restriction $m\ge \ell-1$ belongs to the bound, not to existence of
the moment. At a fixed update count $m$, its largest usable order is
$\ell=m+1$. In particular, a fourth centered moment cannot be inserted with
$m=1$ to claim fourth-order centered accuracy after a single update.
\end{remark}

\subsection{From complete-weight moments to the main theorem}\label{omh:sec:main-proof}
Uniform weight moments supply the baseline. Centered moment bounds then quantify the improvement when the Bellman error term and pool noise are small. Let $\nu_m$ and $\nu_m^T$ denote the physical path and endpoint laws returned by Algorithm~\ref{alg:outer-mh}; $\Pi_n^T$ is the target endpoint marginal.

\begin{proof}[Proof of Theorem~\ref{omh:thm:main-MH}]
Theorem~\ref{omh:thm:proper-global} identifies the augmented target as
$\cP=W\cQ$, with physical marginal $\Pi_n$. Algorithm~\ref{alg:outer-mh}
starts from $\cQ$ and uses independence MH for this target.
For every real $\ell\ge2$, Theorem~\ref{omh:thm:regularity-moments} gives
$\E_{\cQ}W^\ell\le\mathsf M_\ell$, uniformly in $n,K$.
Since $D=\E_{\cQ}W^2-1$, equation~\eqref{omh:eq:overlap-lower} yields
\begin{equation}\label{omh:eq:regularity-overlap}
 c=\E_{\cQ}\min(W,1)
 \ge\frac1{1+\E_{\cQ}W^2}
 \ge\frac1{1+\mathsf M_2}>0.
\end{equation}
Fix a real $p\ge1$ and take $\ell=p+1$ in
Theorem~\ref{omh:thm:MH-high}. For integers $m\ge p$,
\begin{align*}
 \|\cQ\MH^m-\cP\|_{\TV}
 &\le\frac{[2p(1+\mathsf M_2)]^p}{m^p}
            \E_{\cQ}(W-1)_+^{p+1}\\
 &\le\frac{[2p(1+\mathsf M_2)]^p}{m^p}
       \min\{\mathsf M_{p+1},\E_{\cQ}|W-1|^{p+1}\}.
\end{align*}
The last step uses $(W-1)_+^{p+1}\le W^{p+1}$ and
$(W-1)_+^{p+1}\le|W-1|^{p+1}$.
As $\mathsf M_{p+1}\ge1$, a common constant for both branches is
\begin{equation}\label{omh:eq:regularity-Cp}
 C_p^{\rm reg}=[2p(1+\mathsf M_2)]^p\mathsf M_{p+1}.
\end{equation}
Projection cannot increase TV, so in particular
\begin{equation}\label{omh:eq:regularity-MH}
 \|\nu_m-\Pi_n\|_{\TV}
 \le\min\!\left\{1,\frac{C_p^{\rm reg}}{m^p}
             \min\{1,\E_{\cQ}|W-1|^{p+1}\}\right\}.
\end{equation}
Thus the baseline $C_p^{\rm reg}m^{-p}$ holds with no additional assumption on the accumulated Bellman error.

Set $x=Th/(K-1)$ and $\ell=p+1$. Corollary~\ref{omh:cor:weight-scaling}, proved from the local moment and ideal-density lemmas independently of this MH argument, gives
\[
 \E_{\cQ}|W-1|^{p+1}
 \le D_{p+1}^{p+1}(\mathcal A_{2p+2}+\sqrt{x})^{p+1}
 \le2^{(p+1)/2}D_{p+1}^{p+1}
       (\mathcal A_{2p+2}^2+x)^{(p+1)/2}.
\]
For $a,t\ge0$, $\min\{1,at\}\le\max\{1,a\}\min\{1,t\}$. Therefore
\begin{equation}\label{omh:eq:combined-Cp}
 C_p=C_p^{\rm reg}\max\{1,2^{(p+1)/2}D_{p+1}^{p+1}\}
\end{equation}
in~\eqref{omh:eq:regularity-MH} gives~\eqref{omh:eq:main-MH-high}.
Its constants depend only on $p,T,d,c_0,C_0,L_2,\Sigma$ and not on $n,K,m$ or the centered norms of the accumulated Bellman error. Lemma~\ref{omh:lem:ideal-density-moments} supplies the required bounds on these norms and on density moments without an additional hypothesis. Projection to endpoints again cannot increase TV.
One initial complete proposal and $m$ fresh proposals use exactly
$Kn(m+1)$ candidate values and $n(m+1)$ shared-parent coefficient
evaluations, including rejected proposals.
\end{proof}
Equation~\eqref{omh:eq:regularity-MH} also holds with the common constant $C_p$ and remains useful when centered moment bounds are available directly. More general proposals can instead combine the moment bounds in Theorem~\ref{omh:thm:global-moments} with the generic MH bounds above, checking each required moment order.

\subsection{Accumulated Bellman error and fixed correction budgets}
Outer MH reduces the contribution of the Bellman error term and pool noise to output error while leaving the value functions and proposal mechanism fixed.
\begin{corollary}[Near-Doob correction]\label{omh:cor:near-MH}
Under the conditions of Theorem~\ref{omh:thm:main-MH}, for every integer $m\ge1$,
\begin{equation}\label{omh:eq:near-MH}
 \|\nu_m-\Pi_n\|_{\TV}
 =O\!\left(\frac{\mathcal A_4^2+Th/(K-1)}m\right).
\end{equation}
For any fixed real $p\ge1$ and integers $m\ge p$,
\[
 \|\nu_m-\Pi_n\|_{\TV}
 =O_p\!\left(\frac{(\mathcal A_{2p+2}^2+Th/(K-1))^{(p+1)/2}}{m^p}\right).
\]
The constants are independent of $n,K,m$. The local Bellman error bound~\eqref{omh:eq:local-action-cert} with $r=4$ and $\epsilon_{4,k}\le\epsilon_4$ replaces $\mathcal A_4^2$ by at most $T^2\epsilon_4^2$ in the first numerator.
\end{corollary}
\begin{proof}
Use Theorem~\ref{omh:thm:main-MH} at $p=1$ and at the specified order, respectively, and Proposition~\ref{prop:bellman} for the bound on the centered error norm from local Bellman errors. Equivalently, the first conclusion follows from~\eqref{omh:eq:D-near} and Theorem~\ref{omh:thm:MH-second}.
\end{proof}

\begin{corollary}[Fixed outer budget]
\label{omh:cor:exact-MH}
Under the conditions of Theorem~\ref{omh:thm:main-MH}, each fixed integer $m\ge1$ satisfies
\begin{equation}\label{omh:eq:fixed-m}
 \|\nu_m-\Pi_n\|_{\TV}
 =O_m\!\left(\left[\mathcal A_{2m+2}^2+\frac{Th}{K-1}\right]^{(m+1)/2}\right),
\end{equation}
with constants independent of $n,K$. If $B$ is constant, including exact discrete Doob values at the fixed starting state, then every $\mathcal A_r=0$ and the numerator reduces to $(Th/(K-1))^{(m+1)/2}$.
More generally, this constant-$B$ conclusion needs only the conditional moment bound $u_{m+1,k}\le B_{m+1}h/\sqrt{K-1}$ with grid-independent $B_{m+1}$. A variance bound from relative curvature alone gives
\[
 \|\nu_m-\Pi_n\|_{\TV}
 =O\!\left(\frac{\mathcal C_2^2Th}{(K-1)m}\right)
\]
when the variance budget remains bounded.
\end{corollary}
\begin{proof}
Take $p=m$ in Theorem~\ref{omh:thm:main-MH} and absorb the fixed factor $m^{-m}$ into the constant. For the more general constant-$B$ claim,
Theorem~\ref{omh:thm:global-moments} gives $W=M$ and
$\|W-1\|_{m+1}\le\sqrt{e^{\Vbudget_{m+1}}-1}=O_m(\sqrt{Th/(K-1)})$.
Also $\E W^2\le\|W\|_{m+1}^2\le e^{\Vbudget_{m+1}}$, so the bounded moment budget supplies a uniform positive overlap in~\eqref{omh:eq:overlap-lower}.
Theorem~\ref{omh:thm:MH-high} proves the fixed-budget rate.
Theorem~\ref{omh:thm:MH-second} and~\eqref{eq:globalexact} prove the variance-only claim.
\end{proof}
The constants can grow with the order. Fixed-order bounds therefore do not imply a geometric rate in $m$.

\section{A Brownian Resampling Obstruction}
\label{app:proofs}
This appendix proves Theorem~\ref{thm:smc-marginal}. We construct a fixed
Brownian target and value function, then bound the bias caused by full
multinomial resampling. The proof expresses the accumulated population
fluctuations as a sum of Jensen gaps.

\subsection{The fixed target and value function}
Let $P_t$ be the standard Brownian semigroup on $\mathbb R$, and fix
$c\in(0,1)$. Set
\begin{equation}
\begin{aligned}
    \phi_c(x)&=1+c\sin x,
    &\eta_c(dx)&=\phi_c(x)\mathcal N(0,1)(dx),\\
    u_t(x)&=e^{-(1-t)/2}\sin x,
    &P_{t-s}u_t&=u_s\quad(0\le s\le t\le1).
\end{aligned}
\label{smc:eq:target-mode}
\end{equation}
The target is normalized because $\mathbb E\sin B_1=0$ for Brownian
motion started at zero. We use $\phi=\phi_c$ and $\eta=\eta_c$ in
Theorem~\ref{thm:smc-marginal}. Define constants depending only on $c$ by
\begin{equation}
\begin{aligned}
    R_c&=\frac{1+c}{1-c},\qquad q_c=R_c^2,
    &b_c&=\frac12q_c(q_c^2-1),\\
    a_c&=\frac{c(1-c)e^{-1}\sinh(1/8)}{8(1+c)^2},
    &\rho_c&=\min\left\{\frac18,\frac{a_c}{4b_c}\right\}.
\end{aligned}
\label{smc:eq:constants}
\end{equation}
Choose the fixed schedule
\begin{equation}
    \alpha(t)=\frac{(t-1+\rho_c)_+}{\rho_c},\qquad
    H(t,x)=1+c\alpha(t)u_t(x),\qquad 0\le t\le1.
\label{smc:eq:schedule}
\end{equation}
It is Lipschitz in time and smooth in space, satisfies
$1-c\le H\le1+c$, and has $H(0,\cdot)=1$ and
$H(1,\cdot)=\phi_c$. The choice of $c$ fixes the entire schedule
before $n$ and $N$ vary.

Write $h=1/n$, $t_k=k/n$, $H_k=H(t_k,\cdot)$,
$u_k=u_{t_k}$, and $\alpha_k=\alpha(t_k)$. By
\eqref{smc:eq:target-mode},
\begin{equation}
    \frac{P_hH_{k+1}(x)}{H_k(x)}
    =\frac{1+c\alpha_{k+1}u_k(x)}{1+c\alpha_ku_k(x)},
    \qquad 0\le\alpha_{k+1}-\alpha_k\le\frac1{\rho_cn}.
\label{smc:eq:bellman-ratio}
\end{equation}
If $u_k(x)\ge0$, the displayed ratio lies between one and
$1+c(\alpha_{k+1}-\alpha_k)/(1-c)$. If $u_k(x)<0$, its
reciprocal obeys the same bound. Thus
\begin{equation}
    \max\left\{\frac{P_hH_{k+1}(x)}{H_k(x)},
                 \frac{H_k(x)}{P_hH_{k+1}(x)}\right\}
    \le1+\frac{c}{(1-c)\rho_cn}.
\label{smc:eq:local-bound}
\end{equation}
The exact value is $H_k^\star=1+cu_k$. Since both $H_k$ and
$H_k^\star$ lie in $[1-c,1+c]$, either ratio between them is at
most $R_c$.

\subsection{Pair representation and the exact terminal target}
This representation records the two positions needed for each incremental
weight. Let $E_0=\{\star\}$, $E_p=\mathbb R^2$ for
$1\le p\le n+1$, and write $z_p=(x_p,x_{p-1})$ for $p\le n$.
Define
\begin{equation}
\begin{aligned}
    M_1(\star,d(x_1,x'_0))
       &=P_h(0,dx_1)\delta_0(dx'_0),\\
    M_p((x,y),d(z,x'))
       &=P_h(x,dz)\delta_x(dx'),
       \quad 2\le p\le n,\\
    M_{n+1}&=\operatorname{Id},\qquad G_0=1,\\
    G_k(x_k,x_{k-1})&=\frac{H_k(x_k)}{H_{k-1}(x_{k-1})},
       \quad 1\le k\le n.
\end{aligned}
\label{smc:eq:pair-model}
\end{equation}
Every physical mutation is a Brownian increment. The terminal identity
kernel applies the last edge weight without another mutation. Each edge
weight and its reciprocal are bounded by $R_c$.

Set $Q_pf=G_{p-1}M_pf$ and
$Q_{p,q}=Q_{p+1}\cdots Q_q$, with $Q_{q,q}=\operatorname{Id}$.
Their normalized maps are
\begin{equation}
    \Phi_p(\mu)(f)=\frac{\mu(Q_pf)}{\mu(Q_p1)},\qquad
    \Phi_{p,q}(\mu)(f)=\frac{\mu(Q_{p,q}f)}{\mu(Q_{p,q}1)}.
\label{smc:eq:normalized-maps}
\end{equation}
Start from $\eta_0^N=\delta_\star$. Given the particle history
$\mathcal F_{p-1}$, sample $Z_p^1,\ldots,Z_p^N$ independently from
$\mu_p^N=\Phi_p(\eta_{p-1}^N)$ and let
$\eta_p^N=N^{-1}\sum_i\delta_{Z_p^i}$. For $2\le p\le n$,
this means multinomial resampling with $G_{p-1}$ followed by Brownian
mutation; $p=n+1$ is terminal resampling with $G_n$. Equivalently,
each of the $n$ Brownian mutations is followed by its edge-weight
resampling. Keeping the same ancestors and increments couples this pair
system with that implementation exactly.

Along each compatible trajectory, the weights telescope:
\begin{equation}
    \prod_{k=1}^nG_k(x_k,x_{k-1})
    =\frac{H_n(x_n)}{H_0(0)}=\phi_c(x_n).
\label{smc:eq:target-telescope}
\end{equation}
Consequently, the exact terminal pair law is
\begin{equation}
    \eta_{n+1}(dx_n,dx_{n-1})
    =\phi_c(x_n)\mathcal N(0,t_{n-1})(dx_{n-1})
       P_h(x_{n-1},dx_n),
\label{smc:eq:pair-target}
\end{equation}
whose first-coordinate marginal is $\eta_c$. With
$\operatorname{pr}_1(x,y)=x$, the actual output law is
\begin{equation}
    \nu_{n,N}^{\rm SMC}
    =\mathbb E[\eta_{n+1}^N\circ\operatorname{pr}_1^{-1}].
\label{smc:eq:output-law}
\end{equation}
Exchangeability makes this the law of any single terminal particle.

The remaining weights from stage $p$ onward telescope to
$\phi_c(x_n)/H_{p-1}(x_{p-1})$. Future Brownian increments depend
only on $x_p$, so, for $1\le p\le n$,
\begin{equation}
    Q_{p,n+1}(f\circ\operatorname{pr}_1)(x_p,x_{p-1})
    =\frac{P_{1-t_p}(\phi_cf)(x_p)}{H_{p-1}(x_{p-1})}.
\label{smc:eq:future-map}
\end{equation}
In particular,
\begin{equation}
    Q_{p,n+1}1(x_p,x_{p-1})
       =\frac{1+cu_p(x_p)}{H_{p-1}(x_{p-1})}
       \in[R_c^{-1},R_c].
\label{smc:eq:future-weight}
\end{equation}

\subsection{A one-step bound for normalized empirical sampling}
The following estimate controls both the upper bound and the possible
cancellation during the final part of the schedule. Write
$\operatorname{osc}(f)=\sup f-\inf f$.

The bound below specializes the Feynman--Kac bias analysis of \citet[Appendix~E]{zhu2026power} to conditionally i.i.d.\ particle sampling.

\begin{lemma}[A local Feynman--Kac bias bound]
\label{smc:lem:local-bias}
Consider positive operators $Q_1,\ldots,Q_\ell$, the normalized maps
in \eqref{smc:eq:normalized-maps}, and conditionally independent
empirical sampling $\eta_p^N$ from
$\mu_p^N=\Phi_p(\eta_{p-1}^N)$, with deterministic initial law
$\eta_0^N=\eta_0$. Put
\[
    q_{p,\ell}=\sup_{x,y}\frac{Q_{p,\ell}1(x)}{Q_{p,\ell}1(y)},
    \qquad
    P_{p,\ell}f=\frac{Q_{p,\ell}f}{Q_{p,\ell}1},
    \qquad
    \beta(P)=\sup_{\operatorname{osc}(f)\le1}\operatorname{osc}(Pf).
\]
If every $q_{p,\ell}<\infty$, then for $\operatorname{osc}(f)\le1$,
\begin{equation}
\left|\mathbb E\left[
    \Phi_{p,\ell}(\eta_p^N)(f)
    -\Phi_{p,\ell}(\mu_p^N)(f)\mid\mathcal F_{p-1}
\right]\right|
\le\frac{q_{p,\ell}(q_{p,\ell}^2-1)}{2N}\beta(P_{p,\ell}).
\label{smc:eq:local-bias-bound}
\end{equation}
Writing $\eta_\ell=\Phi_{0,\ell}(\eta_0)$ gives
\begin{equation}
    \|\mathbb E\eta_\ell^N-\eta_\ell\|_{\TV}
    \le\frac1{2N}\sum_{p=1}^\ell
       q_{p,\ell}(q_{p,\ell}^2-1)\beta(P_{p,\ell}).
\label{smc:eq:global-bias-bound}
\end{equation}
\end{lemma}

\begin{proof}
Condition on $\mathcal F_{p-1}$ and abbreviate
$\mu=\eta_p^N$, $\nu=\mu_p^N$, $w=Q_{p,\ell}1$,
$g=w/\nu(w)$, $P=P_{p,\ell}$, and $q=q_{p,\ell}$.
Let $a=\Phi_{p,\ell}(\nu)(f)$ and $D=g(Pf-a)$.
Then $\nu(g)=1$, $\nu(D)=0$, and direct expansion gives
\begin{equation}
    \Phi_{p,\ell}(\mu)(f)-\Phi_{p,\ell}(\nu)(f)
    =(\mu-\nu)(D)
       -\frac{(\mu-\nu)(g)(\mu-\nu)(D)}{\mu(g)}.
\label{smc:eq:ratio-expansion}
\end{equation}
The first term has conditional mean zero. The range of $w$ gives
$q^{-1}\le g\le q$, $1/\mu(g)\le q$, and
$\operatorname{osc}(g)\le q-q^{-1}$. For any bounded $v,w_0$,
conditional independence, Cauchy--Schwarz, and
$\operatorname{Var}_\nu(v)\le\operatorname{osc}(v)^2/4$ imply
\[
    \mathbb E\bigl[|(\mu-\nu)(v)(\mu-\nu)(w_0)|
        \mid\mathcal F_{p-1}\bigr]
    \le\frac{\operatorname{osc}(v)\operatorname{osc}(w_0)}{4N}.
\]
Thus the absolute conditional mean in
\eqref{smc:eq:ratio-expansion} is at most
$(q^2-1)\operatorname{osc}(D)/(4N)$. Since $a$ is an average
of values of $Pf$, $|Pf-a|\le\operatorname{osc}(Pf)\le\beta(P)$.
It follows that $\operatorname{osc}(D)\le2q\beta(P)$, proving
\eqref{smc:eq:local-bias-bound}.

The semigroup identity
$\Phi_{p,\ell}\circ\Phi_p=\Phi_{p-1,\ell}$ gives the pathwise
telescoping formula
\begin{equation}
    \eta_\ell^N(f)-\eta_\ell(f)
    =\sum_{p=1}^\ell\left[
       \Phi_{p,\ell}(\eta_p^N)(f)
       -\Phi_{p,\ell}(\mu_p^N)(f)\right].
\label{smc:eq:random-telescope}
\end{equation}
Taking expectations, applying \eqref{smc:eq:local-bias-bound}, and
taking the supremum over functions of oscillation at most one proves
\eqref{smc:eq:global-bias-bound}.
\end{proof}

\subsection{Neutral fluctuations and their Jensen gaps}
Define
\begin{equation}
    p_0=\lfloor(1-\rho_c)n\rfloor+1,\qquad
    f_c(x)=\frac{\sin x}{1+c\sin x},\qquad
    \widetilde f_c=\frac{1-c^2}{2}f_c.
\label{smc:eq:test-function}
\end{equation}
Here $\eta_c(f_c)=0$ and
$\operatorname{osc}(f_c)=2/(1-c^2)$, so
$\operatorname{osc}(\widetilde f_c)=1$. For $p\le p_0$,
$H_{p-1}=1$. The first coordinate therefore follows equal-weight
resampling and Brownian mutation throughout this block. Also,
\eqref{smc:eq:future-map} gives the scalar identity
\begin{equation}
    \Phi_{p,n+1}(\mu)(f_c\circ\operatorname{pr}_1)
    =F_c\bigl(\mu(u_p\circ\operatorname{pr}_1)\bigr),
    \qquad F_c(z)=\frac{z}{1+cz},\quad p\le p_0.
\label{smc:eq:scalar-map}
\end{equation}
Indeed, the numerator is $\mu(u_p\circ\operatorname{pr}_1)$ and
the denominator is $1+c\mu(u_p\circ\operatorname{pr}_1)$.
On $[-1,1]$,
\begin{equation}
    -F_c''(z)=\frac{2c}{(1+cz)^3}\ge\frac{2c}{(1+c)^3}.
\label{smc:eq:curvature}
\end{equation}

\begin{lemma}[Variance in the neutral block]
\label{smc:lem:neutral-variance}
For $1\le p\le p_0$, write $X_p^i=\operatorname{pr}_1(Z_p^i)$ and set
\[
    s_p=e^{-1}\sinh(t_p),\quad
    V_p=\mathbb E\left[\eta_p^N(u_p^2\circ\operatorname{pr}_1)
                   -\eta_p^N(u_p\circ\operatorname{pr}_1)^2\right],
    \quad r=1-\frac1N,
\]
with $s_0=V_0=0$. Then
\begin{equation}
\begin{aligned}
    \mathbb E\operatorname{Var}_{\mu_p^N}(u_p\circ\operatorname{pr}_1)
       &=V_{p-1}+s_p-s_{p-1},\\
    V_p&=r(V_{p-1}+s_p-s_{p-1})
       =\sum_{j=1}^p r^{p-j+1}(s_j-s_{j-1})\ge r^ps_p.
\end{aligned}
\label{smc:eq:variance-recursion}
\end{equation}
\end{lemma}

\begin{proof}
Conditional empirical sampling is unbiased. Since all weights used in
this block are one, induction yields
$\mathbb E\eta_p^N(g\circ\operatorname{pr}_1)=\mathbb E g(B_{t_p})$.
In particular,
\[
    \mathbb E\eta_p^N(u_p^2\circ\operatorname{pr}_1)
    =e^{-(1-t_p)}\frac{1-e^{-2t_p}}2=s_p.
\]
By $P_hu_p=u_{p-1}$, the conditional mean of $u_p(X_p^i)$ is
$\eta_{p-1}^N(u_{p-1}\circ\operatorname{pr}_1)$ (zero at $p=1$).
Therefore the expected conditional variance is
$s_p-(s_{p-1}-V_{p-1})$, proving the first equality in
\eqref{smc:eq:variance-recursion}. For $N$ independent samples from
a law $\nu$, the expected empirical variance is
$(1-1/N)\operatorname{Var}_\nu$. Applying this conditionally gives
the recursion for $V_p$. Its iteration gives the displayed sum.
Finally, $s_j-s_{j-1}\ge0$ and $r^{p-j+1}\ge r^p$, so the sum
is at least $r^ps_p$.
\end{proof}

For $1\le p\le n+1$, define the signed local contributions
\begin{equation}
    L_p=\mathbb E\left[
       \Phi_{p,n+1}(\mu_p^N)(\widetilde f_c\circ\operatorname{pr}_1)
       -\Phi_{p,n+1}(\eta_p^N)(\widetilde f_c\circ\operatorname{pr}_1)
       \right].
\label{smc:eq:local-contributions}
\end{equation}
Equation~\eqref{smc:eq:random-telescope} implies
\begin{equation}
    \eta_c(\widetilde f_c)-\nu_{n,N}^{\rm SMC}(\widetilde f_c)
    =\sum_{p=1}^{n+1}L_p.
\label{smc:eq:signed-telescope}
\end{equation}
For $p\le p_0$, the function
$F_c(z)+cz^2/(1+c)^3$ is concave by
\eqref{smc:eq:curvature}. Conditional Jensen's inequality and the
variance of an empirical mean give
\begin{equation}
    L_p\ge
    \frac{(1-c^2)c}{2(1+c)^3N}
       \mathbb E\operatorname{Var}_{\mu_p^N}(u_p\circ\operatorname{pr}_1)
    \ge0.
\label{smc:eq:jensen-lower}
\end{equation}

\paragraph{Connection to shared ancestry.}
The same calculation makes the role of particle dependence explicit.
Let $U_p=N^{-1}\sum_i u_p(X_p^i)$ and $U_0=0$ in the neutral
block. Brownian harmonicity gives
$\mathbb E[U_p\mid\mathcal F_{p-1}]=U_{p-1}$, hence
\begin{equation}
\begin{aligned}
    \mathbb E U_q^2
    &=\frac1N\sum_{p=1}^q
       \mathbb E\operatorname{Var}_{\mu_p^N}(u_p\circ\operatorname{pr}_1)\\
    &=\frac{s_q}{N}
       +\frac{N-1}{N}\operatorname{Cov}(u_q(X_q^1),u_q(X_q^2)),
       \qquad q\le p_0.
\end{aligned}
\label{smc:eq:ancestry-variance}
\end{equation}
The first equality sums the conditional variances of the martingale
increments. The second follows by expanding the square and using
exchangeability and $\mathbb E U_q=0$. Equal-weight ancestor draws
merge with probability $1/N$ per round, allowing distinct particles
to share Brownian increments. Their covariance contributes to the
variance in \eqref{smc:eq:ancestry-variance}.

More precisely, \eqref{smc:eq:scalar-map} and the martingale property
give $L_p=(1-c^2)\mathbb E[F_c(U_{p-1})-F_c(U_p)]/2$ for
$p\le p_0$. Using $F_c(z)=z-cz^2/(1+cz)$ yields the exact identity
\begin{equation}
    \sum_{p=1}^qL_p
    =\frac{c(1-c^2)}2
       \mathbb E\left[\frac{U_q^2}{1+cU_q}\right],
    \qquad q\le p_0.
\label{smc:eq:ancestry-jensen}
\end{equation}
Since $|U_q|\le1$, this sum is between
$c(1-c^2)\mathbb E U_q^2/[2(1+c)]$ and
$c(1-c^2)\mathbb E U_q^2/[2(1-c)]$. Thus the accumulated Jensen
gap measures the same empirical-mean fluctuations as the covariance
formula. Individual marginals in the neutral block remain Brownian;
these gaps measure their effect after the future normalized weighting.

\subsection{Proof of the terminal TV bounds}
The neutral block supplies a positive contribution at order $n/N$.
The fixed width $\rho_c$ makes the remaining stages too few to
cancel it.

\begin{proof}[Proof of Theorem~\ref{thm:smc-marginal}]
The construction \eqref{smc:eq:schedule}, the target identity
\eqref{smc:eq:target-telescope}, and
\eqref{smc:eq:local-bound} establish the value-function and Bellman-ratio
claims, with $C=c/[(1-c)\rho_c]$.

For the lower bound, define
\[
    J_{n,c}=\left\{\left\lceil\frac n8\right\rceil+1,
                        \ldots,p_0\right\}.
\]
Since $\rho_c\le1/8$, for $n\ge4$,
\[
    |J_{n,c}|\ge
    \left\lfloor\frac{7n}8\right\rfloor
       -\left\lceil\frac n8\right\rceil+1
    \ge\frac{3n}4-1\ge\frac n2.
\]
For $n=2,3$, the same inequality follows by direct substitution.
If $p\in J_{n,c}$, then $t_{p-1}\ge1/8$ and $p-1\le n\le N$.
Also $(1-1/N)^N\ge1/4$ for $N\ge2$. Thus
Lemma~\ref{smc:lem:neutral-variance} gives
\[
    \mathbb E\operatorname{Var}_{\mu_p^N}(u_p\circ\operatorname{pr}_1)
    \ge V_{p-1}\ge r^{p-1}s_{p-1}
    \ge\frac14e^{-1}\sinh(1/8).
\]
Substitution into \eqref{smc:eq:jensen-lower} gives
$L_p\ge a_c/N$ for every $p\in J_{n,c}$.

To bound the remaining contributions, note from
\eqref{smc:eq:future-weight} that the ratio $q_{p,n+1}$ in
Lemma~\ref{smc:lem:local-bias} is at most $R_c^2=q_c$ for
$1\le p\le n$. Its normalized operator is a Markov kernel, so
$\beta(P_{p,n+1})\le1$. The local bound therefore gives
\begin{equation}
    |L_p|\le\frac{b_c}{N}\quad(1\le p\le n),
    \qquad L_{n+1}=0.
\label{smc:eq:late-bound}
\end{equation}
The last equality holds because the future map is the identity at
$p=n+1$ and conditional empirical sampling is unbiased. There are
at most $\rho_cn$ stages with $p_0<p\le n$, since
\[
    n-p_0=n-\lfloor(1-\rho_c)n\rfloor-1\le\rho_cn.
\]
All terms with $p\le p_0$ are nonnegative by
\eqref{smc:eq:jensen-lower}. Equations~\eqref{smc:eq:signed-telescope}
and \eqref{smc:eq:late-bound} now yield
\begin{equation}
\begin{aligned}
    \eta_c(\widetilde f_c)-\nu_{n,N}^{\rm SMC}(\widetilde f_c)
    &\ge\frac{a_c}{N}|J_{n,c}|-\frac{b_c\rho_cn}{N}\\
    &\ge\left(\frac{a_c}{2}-b_c\rho_c\right)\frac nN
     \ge\frac{a_c}{4}\frac nN.
\end{aligned}
\label{smc:eq:terminal-lower}
\end{equation}
Since $\operatorname{osc}(\widetilde f_c)=1$, this proves the TV
lower bound with $c_-=a_c/4>0$.

For the upper bound, apply \eqref{smc:eq:global-bias-bound} to the
pair system with $\ell=n+1$. Each of the first $n$ terms is at
most $b_c/N$, and the last has $q_{n+1,n+1}=1$ and contributes
zero. Projection onto the first coordinate cannot increase TV, so
\begin{equation}
    \|\nu_{n,N}^{\rm SMC}-\eta_c\|_{\TV}
    \le\|\mathbb E\eta_{n+1}^N-\eta_{n+1}\|_{\TV}
    \le b_c\frac nN.
\label{smc:eq:terminal-upper}
\end{equation}
Taking $c_+=b_c$ completes the proof. All constants and the
schedule depend only on the fixed reward amplitude $c$.
\end{proof}

\section{Terminal Importance Resampling}
\label{app:outer_sir_complexity}

This appendix proves a TV bound for one output selected by terminal importance resampling and applies it to complete Stein proposals.

\subsection{Terminal SIR controls an unconditional output law}
One final weight-proportional selection produces a probability law whose TV error can be bounded directly.
For a generic pair $\pi=wq$, suppose $0<w<\infty$ $q$-almost surely
and $\E_qw=1$. Draw $\Omega_1,\ldots,\Omega_N$ independently from $q$
and write $W_i=w(\Omega_i)$. The self-normalized importance measure is
\begin{equation}\label{omh:eq:SNIS-measure}
 \widehat\pi_N=
 \frac{\sum_{i=1}^N W_i\delta_{\Omega_i}}{\sum_{i=1}^N W_i}.
\end{equation}
Terminal sampling--importance--resampling (SIR) additionally selects an
index $I$ with conditional probability $W_i/\sum_jW_j$ and returns
$\Omega_I$. Its unconditional output law is
\begin{equation}\label{omh:eq:SIR-mean-law}
 \overline\pi_N=\E[\widehat\pi_N].
\end{equation}
This is one terminal selection from independent proposals, with no
intermediate population resampling. For every bounded $f$,
$\E\widehat\pi_N(f)=\overline\pi_N(f)$, and hence
\begin{equation}\label{omh:eq:bias-TV-identity}
 \sup_{\osc(f)\le1}
 \abs{\E\widehat\pi_N(f)-\pi(f)}
 =\norm{\overline\pi_N-\pi}_{\TV}.
\end{equation}
The supremum is outside the expectation. In contrast, for a nonatomic
target, the random finitely supported measure $\widehat\pi_N$ has TV
distance one from the target almost surely. Estimator MSE and the
unconditional sampling bias in~\eqref{omh:eq:bias-TV-identity} are
therefore different quantities.

\begin{theorem}[SIR TV and SNIS mean bias]\label{omh:thm:SIR}
Suppose $D=\E_q(w-1)^2<\infty$. For every integer $N\ge1$,
\begin{align}
 \norm{\overline\pi_N-\pi}_{\TV}
 &\le \E_q\frac{w(w-1)_+}{w+N-1}
 \label{omh:eq:SIR-sharp}\\
 &\le \min\left\{s,\frac{D_++s}{N}\right\}
 \le \min\left\{s,\frac{D+\tfrac12\sqrt D}{N}\right\},
 \label{omh:eq:SIR-simple}
\end{align}
where $s=\E_q(w-1)_+$ and $D_+=\E_q(w-1)_+^2$.
The same bound controls the supremum of the SNIS mean bias
in~\eqref{omh:eq:bias-TV-identity}. No uniform positive lower bound or
negative moment of $w$ is required.
\end{theorem}
\begin{proof}
Exchangeability gives, for nonnegative bounded $f$,
\[
 \overline\pi_N(f)
 =N\E\left[\frac{W_1f(\Omega_1)}{W_1+\cdots+W_N}\right].
\]
Conditionally on $\Omega_1=\omega$, the remaining weights are independent
copies of $w$ under $q$. Therefore the density relative to $\pi$ is
\begin{equation}\label{omh:eq:SIR-density}
 g_N(w(\omega))
 :=\frac{\dd\overline\pi_N}{\dd\pi}(\omega)
 =N\E\left[\frac1{w(\omega)+\sum_{i=2}^N W_i}\right].
\end{equation}
For every $a>0$, this expectation is finite and bounded by $N/a$.
Jensen's inequality gives
\[
 g_N(a)\ge\frac{N}{a+N-1}.
\]
Both $g_N\pi$ and $\pi$ have mass one, so
\[
 \norm{\overline\pi_N-\pi}_{\TV}
 =\E_\pi(1-g_N(w))_+
 \le\E_\pi\frac{(w-1)_+}{w+N-1}
 =\E_q\frac{w(w-1)_+}{w+N-1}.
\]
The denominator is at least $w$, and is at least $N$ on $\{w>1\}$.
Thus the last expression is at most $s$ and at most
$N^{-1}\E_q[w(w-1)_+]=(D_++s)/N$. Finally,
$D_+\le D$ and $2s=\E_q|w-1|\le\sqrt D$.
\end{proof}

\begin{corollary}[Terminal importance resampling]\label{omh:cor:Stein-SIR}
Let $\bar\nu_N$ be the physical path law obtained by one weight-proportional selection from $N$ independent complete Stein proposals. Under the finite-moment conditions of Theorem~\ref{omh:thm:global-moments} at order two, write $\dbound=\deltab_2^2$.
\begin{equation}\label{omh:eq:Stein-SIR}
\norm{\bar\nu_N-\Pi_n}_{\TV}
\le\min\left\{1,\frac{\dbound+\tfrac12\sqrt{\dbound}}N\right\}.
\end{equation}
The same bound holds for the endpoint, at cost $KnN$ value queries.
\end{corollary}
\begin{proof}
Set
$q=\cQ$, $\pi=\cP$, and $w=R/\E_{\cQ}R$. Normalizing the weights does
not change the selection probabilities. Theorem~\ref{omh:thm:global-moments}
gives $D\le\dbound$. If $\overline{\cP}_N$ denotes the resulting
augmented output law, Theorem~\ref{omh:thm:SIR} yields
\begin{equation}\label{omh:eq:Stein-SIR-augmented}
 \norm{\overline{\cP}_N-\cP}_{\TV}
 \le\frac{\dbound+\tfrac12\sqrt{\dbound}}{N}.
\end{equation}
For any measurable projection $T$, the preimage of an event is admissible
in the definition of TV; hence
$\norm{T_\#\mu-T_\#\nu}_{\TV}\le\norm{\mu-\nu}_{\TV}$.
Apply this to the physical path and then to its endpoint. Under
\eqref{eq:theory-regularity}, the centered moment bound gives
$\deltab_2=O(\mathcal A_4+\sqrt{Th/(K-1)})$, uniformly over the grid and
pool size. The same assumptions bound this expression, so
$\dbound+\tfrac12\sqrt{\dbound}=O(\mathcal A_4+\sqrt{Th/(K-1)})$.
Consequently,
\begin{equation}\label{omh:eq:near-SIR}
 \norm{\overline\pi_N^{\mathrm{end}}
             -\Pi_n^{\mathrm{end}}}_{\TV}
 \lesssim\frac{\mathcal A_4+\sqrt{Th/(K-1)}}{N}.
\end{equation}
Generating the $N$ complete proposals uses $KnN$ candidate value queries
and $nN$ coefficient evaluations.
\end{proof}

\section{Local Refinement and Exact Weighting}
\label{app:local_refinement}
This extension records the exact weight when local pool MH is included in a complete proposal, and verifies the exponential model's pool and path variances for general pool size. Throughout, $r$ counts local pool updates and $m$ counts outer MH updates.

\subsection{The local proposal and its query cost}
Algorithm~\ref{alg:stein-local} defines the local-refinement family.
Its $r=0$ case is Algorithm~\ref{alg:stein}; retaining the
initial pool average also makes every finite local run usable in an
outer correction.

\begin{sqalgorithm}{\method with pool size $K$ and $r$ local MH updates}\label{alg:stein-local}
Initialize $X_0=x_\star$ and $\log R=\log H_0(x_\star)$. For $k=0,\ldots,n-1$:
\begin{enumerate}
\item Draw the initial pool $U^{(0)}$ by~\eqref{m:pool} and query its values. Record $S_0=S(U^{(0)})$ and retain $U=U^{(0)}$.
\item Repeat exactly $r$ times: draw a fresh pool $V$ at the \emph{same} parent and query its values. Replace $U$ by $V$ with probability $\min\{1,S(V)/S(U)\}$; otherwise retain $U$.
\item Select one candidate from the retained pool proportionally to its value and advance to $X_{k+1}$. Set $\log R\gets\log R+\log S_0-\log H_k(X_k)$.
\end{enumerate}
Return $(X_{0:n},R)$. Launch paths independently for SNIS, or propose fresh complete paths in outer MH with acceptance $\min\{1,R'/R\}$.
\end{sqalgorithm}

All $r+1$ pools can be queried in one batch before the sequential MH decisions. The run uses $K(r+1)n$ candidate values; each retained child value is cached for the next parent. At $K=2,r=0$, the weight telescopes to the two-query statistic in~\eqref{m:weight}.

\paragraph{The invariant pool law.}
At a fixed parent $X_k=x$, let $q_{k,x,K}$ be the centered candidate pool law in
\eqref{eq:construction}. Its pool average is $S(U)$, with expectation
$G_k(x)$, and its normalized pool average is $w(U)=S(U)/G_k(x)$.
The pool target $\pi_{k,x,K}=wq_{k,x,K}$ followed by selection with probabilities
proportional to $H_{k+1}(Y_i)$ gives the exact local tilted kernel $K_k^H$ by
Lemma~\ref{lem:pool}.

Algorithm~\ref{alg:stein-local} approximates this tilted pool law with
exactly $r$ independence MH updates. Each update accepts a fresh pool $V$
from the same parent with probability
\begin{equation}\label{eq:mhaccept}
\alpha(U,V)=\min\{1,S(V)/S(U)\}.
\end{equation}

The unknown $G_k(x)$ cancels from the acceptance ratio. Candidates within a
pool are correlated, whereas the $r+1$ proposed pools are independent at the
same fixed parent. Their value queries can be evaluated in one batch before
the sequential acceptance decisions. The algorithm executes a prescribed
number of updates and retains rejected states. Stopping at the first
acceptance changes its law. Likewise, updating only an index within one
fixed pool does not correct the distribution of that pool.

Write $M=M_{k,x,K}$ for this independence-MH pool kernel.

\subsection{Exact outer correction after local MH}\label{sec:hybrid}

Local MH can also be retained inside the complete-path proposal.  At time
$k$, initialize $U_k^{(0)}\sim q_{k,X_k}$, apply $r$ pool MH updates, and
select the next physical state from $U_k^{(r)}$.  Retain the initial pool
average and use
\begin{equation}\label{eq:hybridweight}
R^{\mathrm{hyb}}
=H_0(x_\star)\prod_{k=0}^{n-1}
\frac{S(U_k^{(0)})}{H_k(X_k)}.
\end{equation}
The initial average is essential: weighting the initial pool changes its
law from $q$ to the invariant pool target $\pi$, after which any finite number
of invariant MH updates preserves the weighted target.

\begin{proposition}[Proper weighting for any finite local run]\label{prop:hybrid}\label{m:hybrid}
Let $\widetilde Q^{\mathrm{hyb}}$ retain the local pool histories and the
physical path.  For every fixed finite $r\ge0$ and bounded measurable
physical path function $F$,
\begin{equation}\label{eq:properweight}
\E_{\widetilde Q^{\mathrm{hyb}}}
\big[R^{\mathrm{hyb}}F(X_{0:n})\big]
=\E_{P_n}\big[\phi(X_n)F(X_{0:n})\big].
\end{equation}
Therefore
$\widetilde\Pi^{\mathrm{hyb}}
=(R^{\mathrm{hyb}}/Z_n)\widetilde Q^{\mathrm{hyb}}$
is a probability law whose physical path marginal is exactly $\Pi_n$.
The case $r=0,K=2$ recovers the path weight of Stein ($K=2$).
\end{proposition}

\begin{proof}
Fix time $k$ and parent $x$, suppress these indices on $q,\pi,M$, and retain
$C_x$ for the terminal child selection kernel.  The identities
$S(U)q(\dd U)=G_k(x)\pi(\dd U)$ and $\pi M^r=\pi$ give
\begin{align}\label{eq:hybridrow}
\E\left[\frac{S(U^{(0)})}{H_k(x)}f(X_{k+1})\right]
&=\frac{G_k(x)}{H_k(x)}\,\pi M^rC_xf\notag\\
&=\frac{G_k(x)}{H_k(x)}\,\pi C_xf\notag\\
&=\frac{P_k(H_{k+1}f)(x)}{H_k(x)},
\end{align}
where the last step uses \eqref{eq:globalpoolidentity}.
For a cylinder function $F(x_{0:n})=\prod_{j=0}^nf_j(x_j)$, iterate this
weighted transition identity from the final time back to the initial
time.  Every intermediate factor $H_k$ cancels.  The prefactor
$H_0(x_\star)$ in \eqref{eq:hybridweight} cancels the initial denominator,
leaving $H_n(X_n)=\phi(X_n)$ and the base transition product $P_n$.
This proves \eqref{eq:properweight} for cylinder functions and hence for
bounded measurable path functions by the monotone class theorem.
In particular $\E R^{\mathrm{hyb}}=Z_n$, and the normalized tilted
augmented law has the stated physical marginal.

For completeness, the standard MH acceptance rule is valid even when the
augmented proposal has no convenient Lebesgue density: with respect to
the product proposal measure, its accepted probability flow is
\[
\frac1{Z_n}\min\{R^{\mathrm{hyb}}(u),R^{\mathrm{hyb}}(v)\}\,
\widetilde Q^{\mathrm{hyb}}(\dd u)
\widetilde Q^{\mathrm{hyb}}(\dd v),
\]
which is symmetric in $u,v$.  The rejection term is supported on the
diagonal.  Detailed balance therefore establishes the claimed invariant
law, without asserting finite update stationarity.  Under an envelope
$R^{\mathrm{hyb}}\le M_\star$, independent proposal acceptance with
probability $R^{\mathrm{hyb}}/M_\star$ has conditional accepted law
$(R^{\mathrm{hyb}}/Z_n)\widetilde Q^{\mathrm{hyb}}$, establishing the
rejection sampling claim.

Finally, $w_k^{(0)}=S(U_k^{(0)})/G_k(X_k)$ has conditional mean one
before pool $k$ is generated, regardless of the subsequent local MH run.
Any conditional pool-moment bounds valid at every reachable hybrid parent
therefore give the product bounds of Theorem~\ref{omh:thm:product} for
$M_{\mathrm{in}}=\prod_kw_k^{(0)}$.
The query count is multiplied by $r+1$ because each parent evaluates one
initial pool and $r$ proposed pools.
\end{proof}

Outer MH on the augmented state consequently uses the
acceptance probability
\[
\alpha=\min\left\{1,
\frac{(R^{\mathrm{hyb}})'}{R^{\mathrm{hyb}}}\right\}.
\]
On rejection it retains the current augmented state and weight.  This
gives the exact invariant physical marginal at any fixed grid.  A finite
number of outer MH updates from a generic initialization still incurs
mixing error.  If a known finite envelope $R^{\mathrm{hyb}}\le M_\star$
is available, independent rejection sampling with acceptance probability
$R^{\mathrm{hyb}}/M_\star$ instead produces exact independent samples at
that grid.  If $P_n$ is an Euler law, neither construction removes the
Euler error relative to a continuous diffusion target.

Replacing $S(U_k^{(0)})$ by $S(U_k^{(r)})$ in
\eqref{eq:hybridweight} is generally invalid. Retaining the initial pool
average in the correction weight does not transfer the higher-order marginal
accuracy of local MH to that weight.

The conditional distribution of the initial pool average is unchanged by subsequent local updates at the same parent.

\subsection{Exact variances for an exponential Brownian tilt}\label{sec:examples}\label{app:brownian}
The following formulas describe the initial normalized pool average and its complete-path product at every pool size.

Consider the one dimensional Brownian reference transition
$P_h(x,\cdot)=N(x,h)$, with $X_0=0$, $T=nh$, and terminal reward
$\phi(x)=e^{\beta x}$, where $\beta\ne0$.  The exact discrete Doob value is
\begin{equation}\label{eq:brownH}
 H_t(x)=\exp\!\left(\beta x+\frac{\beta^2(T-t)}2\right).
\end{equation}
For an antithetic pool, its normalized average is independent of the parent:
\begin{equation}\label{eq:xiBrown}
 \xi_h(Z)=e^{-\beta^2h/2}\cosh(\beta\sqrt h Z),\qquad Z\sim N(0,1).
\end{equation}
Consequently, the raw path proposal has normalized complete weight
$W_h=\prod_{k=1}^n\xi_h(Z_k)$ for independent $Z_k$.
Although $H$ is unbounded, $\xi_h\ge e^{-\beta^2h/2}$ and
$\|\xi_h-1\|_{L^p}=O(h)$ for every fixed $p<\infty$.
Thus the direct weight assumptions for local MH hold uniformly for
$0<h\le h_0$, without assuming a bounded value function.

This example also admits an exact variance formula for every centered pool
size $K\ge2$.  Writing $u=\beta^2h$, the normalized pool average and its
variance are
\begin{equation}\label{eq:exp-exact-K}
 \xi_{h,K}=\frac{e^{-u/2}}K\sum_{i=1}^K e^{\beta\sqrt h Z_i},
 \qquad
 v_K(h):=\Var(\xi_{h,K})
 =\frac{e^u+(K-1)e^{-u/(K-1)}}K-1.
\end{equation}
By symmetry, the sign of $\beta$ does not affect this formula.  With exact
$H$, the initial pool weights are independent across physical steps even
when local MH is used.  Consequently the auxiliary proposal and its
exactly weighted target satisfy
\begin{equation}\label{eq:exp-path-exact-K}
 \chi^2(\widetilde\Pi_n\|\widetilde Q)
   =(1+v_K(h))^n-1.
\end{equation}
Both identities hold at every $h>0$ and follow directly from Gaussian
moment generating functions.  For $K>2$ the second identity is on the
auxiliary space; projection gives a physical path upper bound.

The distribution of this normalized complete weight is exactly independent of
the number of local MH updates. By~\eqref{omh:eq:stationary-rejection}, the
stationary rejection probability of the outer MH chain is therefore also
independent of that number in this model. This statement concerns outer
acceptance; the physical proposal law does change with local correction.

\paragraph{Verification of the exact moment formulas.}
For the centered Gaussian construction,
$\mathbb E Z_iZ_j=-1/(K-1)$ whenever $i\ne j$.  The Gaussian moment
generating function gives $\mathbb E\xi_{h,K}=1$, and expansion of the
square in~\eqref{eq:exp-exact-K} yields
\[
 \mathbb E\xi_{h,K}^2
 =\frac1{K^2}\left(Ke^u+K(K-1)e^{-u/(K-1)}\right).
\]
This proves~\eqref{eq:exp-exact-K}.  Exactness of $H$ implies that each
incremental factor in the normalized complete weight is its initial normalized pool average.
In the exponential example this average is independent of the parent
state and depends only on the fresh Gaussian variables of that pool.
The normalized complete weight therefore has second moment
$(\mathbb E\xi_{h,K}^2)^n$, both for raw paths and for paths with local
pool MH.  The exact auxiliary weighting identity then proves
\eqref{eq:exp-path-exact-K}.

\subsection{Conditions behind the method comparison}\label{app:guidance_comparison}
Table~\ref{tab:guidance-comparison} concerns correction with approximate values, unconditional output TV as numerical depth changes, and parallel candidate queries. For Stein guidance, the primary correction is proposal-initialized outer independence MH. The depth-uniform guarantee uses the fixed start, terminal agreement and uniform regularity conditions of Theorem~\ref{omh:thm:main-MH}; $K=2$ is included. The theorem separates finite-moment Bellman error from candidate-pool noise. Each output uses $Kn(m+1)$ value queries. Our Gaussian theory does not cover discrete state diffusion.

At finite candidate count, SVDD gives approximate local sampling even with exact values and has no outer exact correction. The SMC obstruction in Theorem~\ref{thm:smc-marginal} uses Brownian proposals and multinomial resampling at every step. Its single-output TV error is $\Theta(n/N)$ for $2\le n\le N$, with a fixed target and local Bellman error $O(1/n)$.

\paragraph{Discussion on the DOIT bound.}
For the Monte Carlo estimator in \citet[Lemma~5.2 and Theorem~5.4]{zhu2026doit}, the local mean square error contains $\sigma_{t_\ell}^{-2}$, and its accumulated contribution depends on
\begin{equation}
\kappa_\sigma=\frac{T}{L}\sum_{\ell=1}^{L}\sigma_{t_\ell}^{-2}.
\label{eq:doit-kappa-comparison}
\end{equation}
Symbols retain that paper's conventions. Under its stated $O(\log(T/t_1))$ schedule scaling, $t_1\asymp T/L$ gives $O(\log L)$. This is an upper bound dependence, not observed or necessary variance growth. Increasing $M$ or keeping a positive time cutoff changes the comparison. Algorithms~1 and~2 distinguish full rollouts from the practical surrogate with correction only for $1<\ell\leq\ell^\star$; the latter incurs additional approximation error.

\section{Experimental details and selected results}
\label{app:supporting-results}
This section collects the essential experimental settings, uncertainty estimates,
the learned-policy correction-length ablation, and Gaussian diagnostics.
\subsection{Synthetic centering protocol}
\label{app:centering-essential-settings}
For scalar Brownian motion with $X_0=0$, horizon one, and $t_j=j/n$,
the terminal potential $g(x)=e^{1.5x}$ gives target $\mathcal N(1.5,1)$.
The step size is $h=1/n$. We compare the exact value
$H^*(t,x)=\exp\{1.5x+1.5^2(1-t)/2\}$ with the perturbed value
$H_\delta(t,x)=H^*(t,x)e^{\delta(1-t)\sin x}$, with $\delta=0.5$.
Each step draws $K$ children from the Brownian transition, with opposite
innovations for centered $K=2$ and independent innovations for IID pools.
Selection probabilities are proportional to the child values.
Paths remain independent, with no path resampling or MH.
Writing $X_{j+1}$ for the selected child, each path retains weight
\[
 R=e^{1.5X_n}\prod_{j=0}^{n-1}
 \frac{K^{-1}\sum_{k=1}^K H_\delta(t_{j+1},Y_{j,k})}
 {H_\delta(t_{j+1},X_{j+1})}.
\]
Each independent system averages 16 path weights to obtain $\widehat Z$.
We report adjusted NESS,
\[
 \mathrm{NESS}_{\rm adj}
 =\left(1+16\frac{s^2(\widehat Z)}{\overline{\widehat Z}^{\,2}}\right)^{-1},
\]
where the sample variance $s^2$ and mean $\overline{\widehat Z}$ are computed
across independent systems.
The $K=2$ study uses 4096 systems at each $n=12,24,48,96,192$.
The larger IID pools, $K=4,8,16,32,64,128,256,512,1024$, use 1024 systems
at $n=24,192$.
The 95\% intervals in Figure~\ref{fig:synthetic-centering} use 600 bootstrap
draws of complete systems.

\subsection{RL settings, reference, and uncertainty}
\label{app:rl-experiment-details}
The critic-tilted target favors higher-scoring actions while penalizing KL
divergence from the original policy~\citep{kim2025das}.
The penalty limits policy shifts that could exploit critic errors.
MPO~\citep{abdolmaleki2018mpo} uses the same form of target to train an actor.
The eight fixed Hopper observations have source indices $0,2,3,4,6,12,14,15$.
Every method uses 15 stochastic DDIM steps with $\eta=0.6$ and the frozen
\texttt{behavior\_ckpt600.pth} policy and \texttt{critic\_ckpt100.pth} critic.
The score is $Q=\min(Q_1,Q_2)$, evaluated before environment action clipping.
The main comparison uses $K=2,8$ and $L=m+1=32$ complete proposals per output
for Stein: one initial draw followed by 31 whole-path MH updates.
FK and DAS use $N=32$ particles; both DOIT variants and SVDD use $M=32$
candidates.
Native DOIT scales lookahead scores by their sample standard deviation;
the fixed-temperature variant uses the declared $\beta_s$.
SVDD uses the posterior-mean approximation.
FK resamples at every nonterminal step; DAS resamples when ESS falls below
half its particle count, both multinomially.
TFG uses eight recurrences and one clean-prediction update; TTS-local uses
four recurrences and four clean-prediction updates.

\paragraph{Reference.}
A separate pilot of $4\times65536$ actions per state fixes the highest-decile
threshold $c_s$ and the three inverse temperatures.
After calibration, four independent blocks of 262144 base-policy actions
per state are reserved for validation and reference consistency checks.
The remaining twelve blocks, containing 3,145,728 actions per state, define
the evaluation reference through weights $w_i\propto e^{\beta_sQ(s,a_i)}$.
The weighted CDF uses all these actions directly.
Across states and target strengths, reference ESS ranges from 49674 to
1851193, and the largest KS difference between the validation and evaluation
references is 0.01584.

\paragraph{Proposal values and selection.}
Let $D_j(x)$ denote the clean prediction at node $j$.
For $j<15$, our sampler uses
$H_j^{(c)}(x)=\exp\{c\beta_s[Q(s,D_j(x))-c_s]\}$;
at the endpoint it uses $H_{15}(a)=\exp\{\beta_s[Q(s,a)-c_s]\}$.
Thus $c$ changes the intermediate guidance while preserving the target.
With $S_j=K^{-1}\sum_kH_{j+1}(Y_{j,k})$, the complete-path weight satisfies
$\log R=\beta_sQ(s,X_{15})+\sum_{j=0}^{14}[\log S_j-\log H_{j+1}(X_{j+1})]$.
Whole-path MH accepts with probability $\min\{1,R_{\rm new}/R_{\rm old}\}$.
For each $K=2,8,32$, four validation blocks of 16 chains select
$c\in\{0.25,0.5,1\}$ by mean KS at $L=32$.
In order of increasing target strength, the selected values are $(0.5,1,1)$
for $K=2$ and $(0.5,0.5,0.5)$ for $K=8$.
The follow-up $K=32$ ablation selects $(1,0.25,1)$ using the same rule
and fresh validation draws.
Tuned baseline settings use four validation blocks of 64 outputs per state.
All selections precede the fresh evaluation draws: 16 independent blocks of
64 outputs, totaling 1024 outputs per state and method.

\paragraph{Distribution curves.}
Figure~\ref{fig:rl-distribution-curves} uses the same main-table settings
and fresh evaluation outputs.
For each state and target, let $\mu_s,\sigma_s$ be the weighted mean and
standard deviation of the numerical target reference.
Every method uses $z=(Q-\mu_s)/\sigma_s$, and the displayed CDF equally mixes
the eight states.
The horizontal axis is linear near zero and logarithmic in the tails.
Black denotes the independent numerical SIS target reference.

\paragraph{Statistical comparisons.}
For each of 10000 bootstrap draws, method blocks are resampled independently
and the twelve reference blocks are resampled jointly across methods.
The states remain fixed.
For method $m$, the centered CDF-process norm is
\[
 R_m^*=\frac18\sum_s\left\|
 (F_{m,s}^*-\widehat F_{m,s})-
 (F_{{\rm ref},s}^*-\widehat F_{{\rm ref},s})\right\|_\infty.
\]
Its $1-0.05/27$ bootstrap quantile gives a simultaneous radius for each of
the 27 main method--target rows.
Subtracting the resulting intervals gives the 42 comparisons of $K=2,8$
against seven baselines at three targets.
The calculation uses upper CDF-grid envelopes at all method values and
4097 weighted reference quantiles.
Table~\ref{tab:rl-fixed32-intervals} reports the corresponding pointwise intervals.
\begin{table}[!htbp]
\centering\scriptsize
\setlength{\tabcolsep}{5pt}
\begin{tabular}{lccc}
\toprule
Method & $30\%$ & $60\%$ & $80\%$ \\
\midrule
Stein $K2/L32$ & 0.0306 [0.0000, 0.0656] & 0.0697 [0.0362, 0.1032] & 0.1128 [0.0778, 0.1477] \\
Stein $K8/L32$ & 0.0426 [0.0086, 0.0766] & 0.0942 [0.0574, 0.1310] & 0.1511 [0.1152, 0.1870] \\
DOIT native & 0.2418 [0.2073, 0.2763] & 0.4778 [0.4419, 0.5138] & 0.6179 [0.5828, 0.6531] \\
DOIT fixed $\beta$ & 0.2780 [0.2436, 0.3123] & 0.4774 [0.4414, 0.5134] & 0.5903 [0.5545, 0.6262] \\
SVDD $M32$ & 0.1218 [0.0853, 0.1583] & 0.2276 [0.1929, 0.2623] & 0.2979 [0.2604, 0.3354] \\
FK $N32$ & 0.0504 [0.0138, 0.0869] & 0.1093 [0.0746, 0.1439] & 0.1816 [0.1439, 0.2193] \\
DAS $N32$ & 0.0307 [0.0000, 0.0662] & 0.0769 [0.0401, 0.1136] & 0.1355 [0.0974, 0.1737] \\
TFG & 0.1788 [0.1437, 0.2139] & 0.2808 [0.2456, 0.3160] & 0.3586 [0.3199, 0.3972] \\
TTS-local & 0.1781 [0.1404, 0.2158] & 0.2767 [0.2416, 0.3118] & 0.3752 [0.3381, 0.4123] \\
\bottomrule
\end{tabular}
\caption{Mean KS with pointwise 95\% centered-CDF bootstrap bands.}
\label{tab:rl-fixed32-intervals}
\end{table}

\subsection{Whole-path correction with learned values}
The configurations in Table~\ref{tab:rl-fixed32-ablation} use the same proposals
at each reported correction length. Thus the $L$ comparisons isolate whole-path
MH. The intermediate strength is selected separately for each $K$ and target.
\begin{table}[!htbp]
\centering\small
\setlength{\tabcolsep}{5pt}
\begin{tabular}{rrrrrrrr}
\toprule
Target & $K$ & $c$ & $L=1$ & $L=8$ & $L=32$ & Accept. & Unmoved \\
\midrule
30\% & 2 & 0.5 & 0.2174 & 0.0463 & 0.0306 & 51.73\% & 0.20\% \\
30\% & 8 & 0.5 & 0.2136 & 0.0525 & 0.0426 & 46.43\% & 0.26\% \\
30\% & 32 & 1 & 0.1180 & 0.0544 & 0.0469 & 35.85\% & 0.50\% \\
60\% & 2 & 1.0 & 0.2804 & 0.1032 & 0.0697 & 34.74\% & 0.57\% \\
60\% & 8 & 0.5 & 0.3617 & 0.1376 & 0.0942 & 29.79\% & 0.66\% \\
60\% & 32 & 0.25 & 0.4553 & 0.1602 & 0.0907 & 30.24\% & 0.56\% \\
80\% & 2 & 1.0 & 0.4041 & 0.1804 & 0.1128 & 29.08\% & 0.63\% \\
80\% & 8 & 0.5 & 0.4713 & 0.2333 & 0.1511 & 24.37\% & 0.92\% \\
80\% & 32 & 1 & 0.2933 & 0.1958 & 0.1436 & 19.25\% & 1.40\% \\
\bottomrule
\end{tabular}
\caption{Fresh-sample ablations. The intermediate strength $c$ is selected separately for each $K$ and target by validation KS at $L=32$. Acceptance and unmoved fractions refer to the 31 MH decisions.}
\label{tab:rl-fixed32-ablation}
\end{table}

\subsection{Gaussian guidance and proper-weight diagnostics}
\begin{figure}[t]
\centering\includegraphics[width=\linewidth]{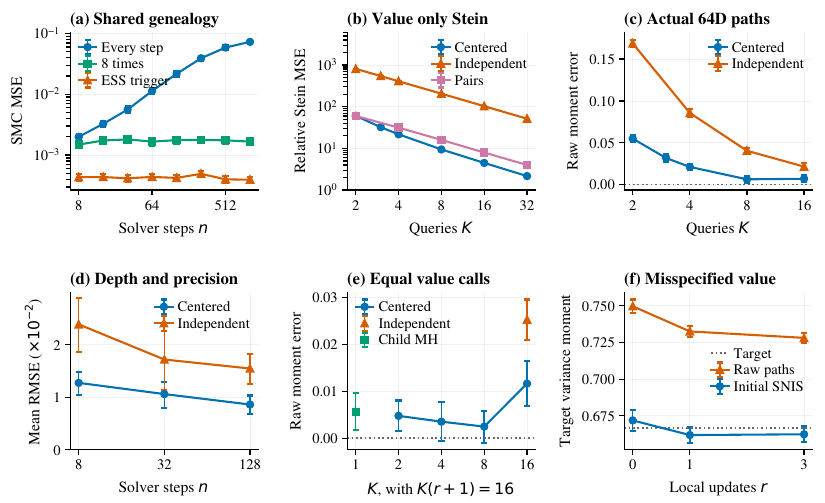}
\caption{Executed experiments. (a) Neutral SMC, 500 populations of $N=256$. (b) Relative MSE of $\shat$, $d=64$, $h=1/64$, 8192 pools per point. (c) Raw variance moment error, $d=64$, $n=32$. (d) SNIS mean RMSE, $K=4$. (e) Actual equal cost runs, $n=16$, $K(r+1)=16$. (f) Misspecified values, $\alpha=4$, $K=4$, $n=32$. Panels (c--f) use 32 groups of 128 complete paths. Bars are pointwise 95\% intervals; (d) uses a population bootstrap. The dotted lines mark exact targets.}
\label{fig:stein_experiments}
\end{figure}

Figure~\ref{fig:stein_experiments} separates estimator accuracy, raw endpoint
moments, and weighted estimation. These runs use local pool refinement and
self-normalized importance sampling.
Here $r$ counts the local pool MH updates at each transition.
For Brownian motion started at zero with horizon one, $\tau^2=2$, and
$y=\mathbf1/2$, the terminal
potential and approximate values are
\begin{equation}\label{eq:gaussian-diagnostic-model}
 \phi(x)=e^{-\|x-y\|^2/(2\tau^2)},\qquad
 H_t^\alpha(x)=\left(\frac{\tau^2}{v_\alpha(t)}\right)^{d/2}
 e^{-\|x-y\|^2/(2v_\alpha(t))},\quad
 v_\alpha(t)=\tau^2+\alpha(1-t).
\end{equation}
The value is exact at $\alpha=1$; changing $\alpha$ preserves the terminal potential.
The target is $\mathcal N(\mathbf1/6,(2/3)I_d)$.
The observables $f(x)=d^{-1}\sum_jx_j$ and
$q(x)=d^{-1}\|x-\mathbf1/6\|^2$ have target means $1/6$ and $2/3$.
Each complete-path setting uses 32 independent groups of 128 paths.
Mean intervals use $t_{31}$ across group summaries; RMSE intervals use
10000 bootstrap draws of whole groups.
In the equal-query experiment, path methods use $n=16$ and $K(r+1)=16$,
giving 32768 candidate-value evaluations per group.
The SMC comparator uses 2048 particles and 16 mutations at the same query cost,
with multinomial resampling when ESS falls below half the particle count.
For local refinement, all path weights retain the pool averages recorded
before the local MH updates.

\end{document}